\documentclass[twocolumn]{autart}
\newcommand{\E}{{\mathbb E}}

\newcommand{\R}{{\mathbb R}}

\newcommand{\trnsp}{\mathsf{T}}

\newcommand{\argmax}[2]{\underset{#1}{\mathrm{arg}\,\mathrm{max}}\,\,#2}

\RequirePackage{amssymb, graphicx, url,mathtools,
	xspace,subfigure,braket,epstopdf}
\usepackage{amsmath,amsfonts,framed,enumitem}
\usepackage{comment} 
\usepackage[makeroom]{cancel}

\usepackage{algorithm,algpseudocode}

\DeclareMathOperator*\argmin{arg\,min}

\usepackage{graphicx,subfigure}% for pdf, bitmapped graphics files
\usepackage{epsfig} % for postscript graphics files
\usepackage{mathptmx} % assumes new font selection scheme installed
\usepackage{times} % assumes new font selection scheme installed
\usepackage{amsmath} % assumes amsmath package installed
\usepackage{amssymb}  % assumes amsmath package installed
\usepackage{amsfonts}
\usepackage{euscript}
\usepackage{color}
\usepackage[dvipsnames]{xcolor}
\usepackage{soul}
\usepackage[hidelinks]{hyperref}
\edef\endfrontmatter{%
	\unexpanded\expandafter{\endfrontmatter}% the current code
	\noexpand\endNoHyper % add \endNoHyper at the end to match \NoHyper
}

\begin{document}
	\begin{frontmatter}
		
		\thanks[footnoteinfo]{This research has been partially supported by the
			Progetto di Ateneo CPDA147754/14-New statistical learning approach
			for multi-agents adaptive estimation and coverage control, 
			the project \emph{Deep probabilistic regression -- new models and learning algorithms} (contract number: 2021-04301), funded by the Swedish Research Council,
			the \emph{Wallenberg AI, Autonomous Systems and Software Program (WASP)} funded by Knut and Alice Wallenberg Foundation and 
			by \emph{Kjell och M{\"a}rta Beijer Foundation}.
			This paper was not presented at any IFAC meeting.
			Corresponding author Gianluigi Pillonetto Ph. +390498277607.}
		
		\title{Deep networks for system identification: a survey}
		
		\author[padova]{Gianluigi Pillonetto}
		\author[washington]{Aleksandr Aravkin}
		\author[uppsala]{Daniel Gedon}
		\author[linkoping]{Lennart Ljung}
		\author[uppsala]{Ant\^onio H. Ribeiro}
		\author[uppsala]{Thomas B. Sch\"{o}n}
		\address[padova]{Department of Information  Engineering, University of Padova, Padova, Italy (e-mail: giapi@dei.unipd.it)}
		\address[washington]{Department of Applied Mathematics, University of Washington, Seattle, USA (e-mail: saravkin@uw.edu)}
		\address[uppsala]{Department of Information Technology, Uppsala University, Sweden (e-mails: \\ \{daniel.gedon, antonio.horta.ribeiro, thomas.schon\}@it.uu.se)}
		\address[linkoping]{Department of Electrical Engineering, Link\"{o}ping University, Link\"{o}ping, Sweden (e-mail: lennart.ljung@liu.se)}

		\begin{abstract}
			Deep learning is a topic of considerable current interest. The availability of massive data collections and powerful software resources has led to an impressive amount of results in many application areas that reveal essential but hidden properties of the observations. System identification %which learns dynamic models from inpu-output data, 
			learns mathematical descriptions of dynamic systems from input-output data and can thus 
			benefit from the advances of deep neural networks to enrich the possible range of models to choose from. For this reason, we provide a survey of deep learning from a system identification perspective. We cover a wide spectrum of topics  
			to enable researchers to understand the methods, providing rigorous practical and theoretical insights into the benefits and challenges of using them. % for system identification. 
			The main aim of the identified model is to predict new data from previous observations. This can be achieved with different deep learning based modelling 
			techniques and we discuss %. We discuss deep %neural network 
			architectures commonly adopted in the literature, like feedforward, convolutional, and recurrent networks. Their parameters have to be estimated from past data trying to optimize the prediction performance. %Dealing with nonlinear deep models, 
			For this purpose, we discuss a 
			specific set of first-order optimization tools that is emerged as efficient. % and is discussed. % in the survey. %\sa{I suggest using present tense consistently throughout the survey; e.g.  `will be discussed' should be `is discussed'. } 
			The survey then draws connections to the well-studied area of kernel-based methods. They control the data fit by  regularization terms that penalize models not in line with prior assumptions. We illustrate how to cast them in deep architectures to obtain deep kernel-based methods. The success of deep learning also resulted in surprising empirical observations, like the counter-intuitive behaviour of models with many parameters. We discuss the role of overparameterized models, including their connection to kernels, as well as implicit regularization mechanisms which affect generalization, specifically the interesting phenomena of benign overfitting and double-descent. Finally, we highlight numerical, computational and software aspects in the area with the help of applied examples.
		\end{abstract}
		\maketitle
		
	\end{frontmatter}

	\section{Introduction}

	The term \emph{system identification} was coined in the late 1950s by Zadeh in \cite{Zadeh1956}.
	His seminal work %underlined the importance of developing 
	set up a framework  
	where mathematical laws governing dynamic systems %(like input-output relationships)
	are built starting from input-output measurements \cite{Astrom71}. 
	Since then, system identification has enjoyed many theoretical developments, becoming a key subfield of Automatic Control with several applications in science and engineering \cite{Ljung:99}. Despite its long history,  research in data-based dynamical modeling is still extremely active.  
	Many complex physical processes, arising e.g. in biology and industry, require  an increasingly deep knowledge for improved prediction and control.
	This often requires handling high-dimensional data, posing new challenges. In addition, recently there has been a lot of interesting cross-fertilization 
	between system identification and other areas, including  machine learning and Bayesian statistics, with developments rooted in the theory of regularization \cite{SpringerRegBook2022}. 
	One emerging link, central in this survey, concerns the use of hierarchical structures known as \emph{deep neural networks} for learning dynamic models from data. 
	This link motivates a close look at seemingly well-established facts about the classical bias/variance trade-off:
	while classical principles suggest that the model should describe the experimental data without being too complex to be fooled by noise, there are complex architectures that only emphasize fitting data and still perform well on validation data. 
	To properly contextualize deep networks, we first give a brief overview on state-of-the-art identification techniques. 
	Our focus is on a black-box view of the problem. The aim is to estimate input-output relationships 
	assuming that there is no structural/physical knowledge available about the system we are interested in. %, with the goal of estimating the input-output relationships associated with such system. %functions which maps system inputs into outputs.

\begin{figure}
		\begin{center}
			
			{ \includegraphics[width=\columnwidth]{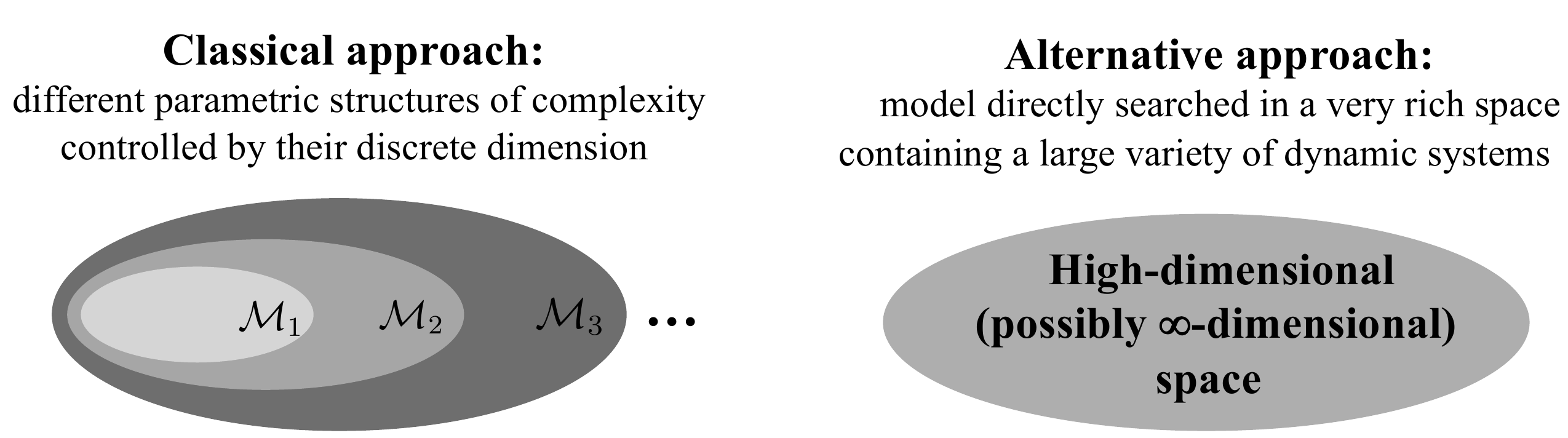}}   
			\caption{\emph{Classical approach to system identification (left).} Parametric (and often nested) structures $\mathcal{M}_i$  are postulated, each depending on an unknown parameter vector. Complexity is related to discrete model-order (dimension of the parameter vector) and controlled using e.g. AIC/BIC criteria or cross validation. \emph{Alternative approach to system identification (right).} Estimates of a dynamic system are searched for in a high-dimensional space (dimension could be also infinite) with complexity tuned by regularization. Examples % spaces 
				are obtained by introducing a large set of basis functions or by implicitly encoding them in a kernel (hence defining a reproducing kernel Hilbert space). Another approach to define the high-dimensional space uses deep neural networks (compositional functions) to model dynamic systems. This will be the central theme of this survey.
			} \label{Fig1}
		\end{center}
	\end{figure}

	\paragraph*{Classical approach}
	%As done in Below, we focus on a black-box view of the problem, assuming that there is no structural/physical knowledge available about the system we are interested in. 
	Classical textbooks treating the subject are \cite{Ljung:99,Soderstrom} where techniques relying on 
	paradigms from mathematical statistics are described.
	They strongly rely on 
	prediction error methods (PEM) \cite{Astrom1965} and the concept of discrete model order. 
	Different structures are postulated which depend on
	an unknown parameter vector %$\theta$ 
	whose (finite) dimension determines complexity. 
	Examples are FIR and ARX models in the linear setting, and their nonlinear counterparts 
	NFIR and NARX. 
	%In these stuctures, the dimension of $\theta$ determines the system memory, i.e. the number of
	%past input-output couples which influence the system output. 
	%If the true system has the postulated model structure, and the noise is Gaussian, this procedure is asymptotically optimal: it cannot be outperformed by any other unbiased estimator as the number of measurements grows to infinity \cite{CaseBerg:01}. However, a crucial point here is the 
	Selection of the most adequate model structure is a key aspect 
	of the identification procedure and it is typically performed 
	by following the Occam's razor principle.
	The dimension of the unknown parameter vector % $\theta$  
	is tuned in order to reach 
	a good trade-off between bias and variance.
	Complexity measures that implement such principle include 
	AIC \cite{Akaike:74} and BIC \cite{Rissanen:78,Schwarz:78}, 
	where data fitting is complemented with a penalty term that
	depends on the model (discrete) dimension. 
	Another popular approach to select the structure is cross validation (CV) \cite{Hastie01,StoneCV:77}. 
	Fig. \ref{Fig1} (left) graphically describes the classical route to system identification.
	
	\paragraph*{Regularized approach}
	%Alternative approaches to system identification have emerged in recent years.
	As illustrated in the right part of Fig.~\ref{Fig1},
	%a common thread that binds them is that, 
	rather than introducing finite-dimensional models of increasing complexity, 
	an alternative to the classical approach is to search for the dynamic system directly 
	in a high-dimensional (possibly infinite-dimensional) space. 
	It could e.g. contain all the impulse responses
	of stable systems in the linear setting or all the continuous input-output maps in the nonlinear one. 
	This typically makes system identification an ill-posed problem in the sense of Hadamard \cite{Had22}.
	The challenge is to control model complexity without decreasing the model dimension and
	regularization theory becomes an important tool here.  
	A common thread that binds the multitude of available regularized approaches
	is the introduction of a ranking of potential solutions in the search space. 
	Among different models that fit the data in a similar way, the one that reflects our expectations the most 
	is selected \cite{Bertero1}. Regularizers, like norms of solutions, can in this way implicitly
	reduce the hypothesis space to a subset of
	manageable complexity. % \cite{SpringerReg2022}.\\
	
	Two important examples following this paradigm %have been adopted and 
	have been well documented in the system identification literature.
	In the first one, the dynamic system is seen as the sum of a large number of basis functions 
	%which,  e.g. in the nonlinear setting they 
	that can account for different inputs-outputs lags and nonlinear interactions \cite{SchoukensLjung:2019}. %, hence reflecting 
	%our poor knowledge on system dynamics. 
	Then, sparsity promoting penalties are introduced
	to control the complexity by selecting only those functions relevant to describe the system dynamics. 
	For instance, the use of the $\ell_1$-norm 
	%as regularizer on $\theta$  
	leads to the LASSO \cite{Tibshirani:96}
	while more recent variants are described in \cite{Bai2019,Birpoutsoukis2017,Bai2018,Rosasco2013,Smith2014,Stoddard2017}.
	%This allows to jointly perform estimation and variable selection, 
	%trying to automatically set to zero groups of variables in the regression vector. 
	%The use of the $\ell_1$-norm 
	%as regularizer on $\theta$  leads to the LASSO \cite{Tibshirani:96}
	%while more recent variants include \cite{Rosasco2013,Bai2018,Bai2019}.
	Other penalties for system identification  
	relying on atomic and Hankel nuclear norms are reported in
	\cite{Chand2012,liu2009,Mohan2010,Smith2014}.
	Regularization methods to control (truncated) Volterra series \cite{Boyd1985,Cheng2017} 
	are described in \cite{Birpoutsoukis2017,Stoddard2017}.
	
	A second class of regularized approaches that has become popular in recent years is the so called
	kernel approach to system identification \cite{PillonettoDCNL:14}. 
	The class of models in the right part of Fig.~\ref{Fig1}
	is set to a reproducing kernel Hilbert space (RKHS), completely defined
	by a positive-definite kernel  \cite{Aronszajn50}. 
	Systems inherit kernel properties,
	e.g. continuous kernels induce spaces of continuous functions \cite{Cucker01}
	while absolutely integrable kernels induce stable RKHSs  
	that contain only absolutely integrable impulse responses 
	\cite{MathFoundStable2020}. 
	%In addition,  the so called universal kernels can approximate any continuous
	%function over any compact set \cite{Micchelli06}.
	%Regularization then relies on penalty terms induced by the kernel \cite{Scholkopf01b} 
	Important estimators that exploit RKHSs are regularization networks (kernel ridge regression) and support vector regression/classification \cite{PoggioMIT}. 
	They have been widely employed in system identification, see e.g. 
	\cite{CALCP14,COL12a,DallaLib2021,PartLin1,Frigola2014,Frigola2013a,Frigola2013b,Lindsten2013,Pillonetto:10a,PCCDL16}.
	%, see also \cite{PillonettoDCNL:14} for a survey. PCP17
	The popularity of these methods derives also from their connection with Gaussian regression \cite{AravkinNN2015,Kimeldorf70,Rasmussen} and from the famous representer theorem \cite{ArgyriouD2014,Scholkopf01,Wahba:90}. It shows that  all the kernel-based estimators commonly adopted 
	are sums of a finite number (equal to the dataset size) of basis functions, %immediately
	computable from the kernel. Hence, the system estimate has the structure of a 
	one-hidden layer network, also called shallow network in the literature.
	The aim of this survey is to illustrate a third way to define the high-dimensional search space
	in the right part of Fig.~\ref{Fig1} where shallow are replaced by deep neural networks.\\
	
	\paragraph*{Deep neural networks}
	Spaces of dynamic systems (possibly) even more complex than those 
	described in the two examples above can be obtained by 
	neural networks which are composed of sequences of linear transformations and static nonlinearities. 
	This concept is further expanded by adopting deep neural networks  \cite{Haykin:99}.
	They represent hierarchical structures with many layers
	and processing units, interpretable as composition of functions. % and can be understood as the composition of functions.
	These models can include thousands of parameters
	to describe weights in the connections and different types of
	nonlinearities, with the capability of approximating a large class of systems.
	
	Neural networks have been around for more than 70 years \cite{McCulloch1943}
	and multilayer networks %have been around 
	at least 
	since 1970s, notable examples being the Necognitron used to 
	determine characters \cite{Fukushima1980}
	and the convolutional networks \cite{LeCun1998}.
	A great advancement was obtained in 2006
	thanks to \cite{Hinton2006} which introduced a
	fast learning scheme (based on a greedy strategy)
	for deep belief networks. The procedure extracts features from the input data,
	suitably reducing their dimension, hence mitigating the presence
	of local minima in the training process. 
	Since then, the popularity of deep networks has further increased 
	thanks also to massively increased datasets, availability of larger-scale computational resources, improved software tools, 
	investment from industry, and other architectural and algorithmic innovations.
	Deep networks have thus found an impressive number of applications, in particular in unsupervised
	learning and pattern recognition where a large amount of unlabeled data have to be analyzed.
	Examples include for example computer vision, motion estimation, speech recognition and detection, natural language processing, see
	\cite{DeepNature} and the recent survey  \cite[Section 6]{DeepSurveyNC2017}
	for a detailed list of references.
	
	\paragraph*{Deep networks for system identification}
	Neural networks have enjoyed a long and fruitful history also within the system identification community, where they have been a popular choice for modeling nonlinear dynamical systems for many years \cite{Chen1990,narendra_identification_1990}. Over the past decade, renewed interest in neural networks has been motivated by the many success stories in science and engineering mentioned in the previous paragraph. There has been quite a few interesting and highly relevant technical developments in deep learning that are also interesting for the system identification community. The aim of this survey is to reinforce the bridge between the system identification and deep learning communities 
	by reviewing the (still surprisingly vague) connection between these two areas, stimulating research at their intersection.
	While compositionality was a main motivation
	for hierarchical models of the visual cortex \cite{PoggioCortex1999}, 
	our goal is to describe its relevance for the design of new identification procedures.
	For this purpose, available models and methods of deep learning will be introduced in a pedagogical way, tailored for researchers with a system identification background. The aim is to make deep learning a natural tool for system identification focusing on its use for regression problems.
	
	Among the hierarchical structures relevant for system identification, we will survey \emph{multilayer feedforward neural networks} that lead 
	to deep NFIR and NARX models \cite{DynoNetPiga2021,LjungDeep2020} and can include dynamic systems features, e.g.  
	fading memory concepts as recently proposed in \cite{SchoukensNFIR2021,zancato2021novel}.
	Here, deep learning can also bring extra flexibility for probabilistic approaches using e.g.
	\emph{energy-based models} \cite{Hendriks2021_ebm}.
	%For robust control purposes, uncertainty bounds for the estimated networks are also highly desirable. 
	Other structures, like \emph{temporal convolutional networks}, find connections
	with the important area of nonlinear modeling via Volterra series and block-oriented models, like Wiener and Hammerstein structures \cite{andersson_deep_2019,Wray1994Volterra}.
	Another route for system identification has been recently traced 
	by combining \emph{kernel-based methods and deep networks} 
	%A first approach to further enhance kernels flexibility consists of using 
	%a neural network to map inputs to points in an intermediate feature space that forms the domain
	%of functions belonging to a reproducing kernel Hilbert space.
	%This then leads to multilayer feedforward neural networks
	%with last layer defined by a kernel, see \cite{Ober2021}.
	%%which also describes difficulties related to complexity control using 
	%where connections with Gaussian regression can be also found. % and the concept of marginal likelihood.
	%Deep kernel learning is also obtained 
	also proposing structures completely defined
	by compositions of
	kernels 
	%, hence associating vector-valued reproducing kernel Hilbert spaces 
	%\cite{Micchelli:2005} to all the layers 
	\cite{Calandra2016,KernelDEEP2009}.
	%This leads to new spaces of compositional functions,
	%with connections also with multiple kernel learning when two layers are adopted
	%\cite{Dinuzzo2010}.
	An extension of the representer theorem has recently been obtained in \cite{Bohn2019}, the so called \emph{concatenated representer theorem}. 
	%that reduces a nonlinear infinite-dimensional optimization problem
	%to a nonlinear finite-dimensional one.
	%A third approach to survey is related to the use of deep kernel learning
	Deep learning to enhance long-term prediction capability of a model %of an identification procedure
	is also described in \cite{DallaL:2021b}. 
	%This is an important point related to a distinctive feature of system identification
	%where the target is not only the reconstruction of static maps, but  
	%the simulation of complex, possibly nonlinear, systems \cite{SchoukensLjung:2019}.
	%This makes concepts like 
	In this setting, concepts like input-to-state stability \cite{Jiang2001,Sontag1989} 
	in deep hierarchies become relevant, % to survey, 
	see e.g. \cite{BonassiStability2021,Miller2018,Sanchez1999Stability}.
	%The main problem is that simulation is much more difficult than one-step-ahead prediction since regressors have to
	%be defined using previously estimated outputs, leading to a particular form of  
	%compositionality. % used to define predictors over different horizons.
	%Hierarchical structures containing layers, each associated to a different prediction horizon
	%and to a kernel, 
	%have been recently proposed in \cite{DallaL:2021b}.\\ 
	%to %promote stability and 
	%improve long-term predictions and model stability, the so called \emph{deep prediction networks} \cite{DallaL:2021b}.\\
	\emph{Deep state-space models} are also gaining in popularity. They require structures even more expressive 
	than feedforward neural networks with recurrent components. % have also to be included. 
	In this framework, \emph{recurrent
		neural networks} and \emph{long short term-memory networks} have become
	important \cite{Gedon2021,Karl2017,LjungDeep2020,Watter2015}.
	In recent years, many approaches also exploit hierarchical structures to map
	past input-output couples into a state vector by combining
	\emph{Koopman operator theory} \cite{Koopman1931,Koopman1932} and \emph{deep autoencoders} \cite{Baldi1989,Hinton2006}.
	Autoencoders are relevant in dynamic systems since their scope is ideally to map information vectors into
	themselves, obtaining a lower-dimensional
	data representation in a hidden layer that can be interpreted as the system state in a latent space.
	%with connections with principal component analysis in the linear setting \cite{Baldi1989}. 
	Recent work that exploit this idea % to estimate state space models 
	using linear or nonlinear embeddings can be found in %, respectively, in 
	\cite{MastiDeepSS2021,Iacob2021,Lusch2018,MastiDeepSS2018}. 
	%Recent works derive linear embeddings, i.e. nonlinear models are estimated 
	%by learning a latent space where the dynamics are linearized and propagated via the Koopman operator \cite{Iacob2021,Lusch2018}.
	%Examples are \cite{Lusch2018} for the autonomous case (including deep learning
	%of dynamic systems with continua spectra) and
	%\cite{Iacob2021} where availability of measurements of the full state is not required. 
	%State space models estimated with
	%Works that do not require linear dynamics in the latent space 
	%are \cite{MastiDeepSS2021,MastiDeepSS2018}. % while 
	Other recent approaches exploit \emph{deep variational autoencoders} %  \cite{Kingma2019}
	%and \emph{energy-based models} \cite{Hendriks2021_ebm}
	to estimate a probability distribution over the latent space \cite{Kingma2019}.

	\paragraph*{Optimization and regularization} 
	Once a deep structure is selected, its parameters have to estimated from
	the available input-output data. This usually requires the solution of
	a high-dimensional non-convex optimization problem. Some form
	of regularization is also often needed to control complexity. 
	There are some differences between parameter estimation through optimization in system identification and deep learning
	that will be discussed in this survey.
	Because of prohibitive costs, second-order methods are rarely used in deep networks, while 
	\emph{first-order stochastic methods} are dominant. We present different approaches for large models and datasets, including.
	steepest Gradient, Prox Gradient, and Stochastic Gradient %coming e.g. from %like e.g. described in 
	\cite{Bottou2018,Bousquet2008,lydia2019adagrad,reddi2016stochastic,zhang2018improved,zou2019sufficient}.
	We also describe related problems, like vanishing and exploding gradients \cite{hanin2018neural,pascanu_difficulty_2013}, and benefits coming from techniques like variance reduction, adaptive scaling 
	and recent powerful software packages for automatic differentiation \cite{abadi2016tensorflow,gulli2017deep,paszke2019pytorch}.
	Computing parameters in large structures may also call for the use of special techniques
	that can be interpreted as regularization measures. % and have been recently applied also in system identification.
	Examples are weight regularization, weight and batch normalization, dropout or dilution \cite{Labach:19,PoernomoK:18}
	and early stopping \cite{sjobergL:1995}.
	
	\paragraph*{Theoretical properties of deep networks} 
	%Beyond the many proposed structures and tricks for optimization/regularization,
	%there have been some recent theoretical developments regarding deep networks.
	Another aim of the survey is to review 
	recent theoretical developments on deep networks 
	%discussing their relevance 
	under a system identification perspective.
	It has been shown that shallow networks, like those induced by 
	many kernel-based methods, are \emph{universal approximators}~\cite{Barron93,Mhaskar96},
	able to reproduce almost perfectly most systems of interest. 
	However, the number of required trainable parameters 
	might scale exponentially with the required accuracy.
	Deep networks can instead exploit knowledge on the compositional nature of the problem %can help with the curse of dimensionality 
	by reducing the complexity from exponential to linear \cite{PoggioPNAS2020,IJAC2017}.
	%Indeed for compositional functions, a deep network can
	%achieve the same resolution with much less parameters \cite{PoggioPNAS2020,IJAC2017}.
	%Adoption of  deep networks for regression is also supported by \cite{Daubechies2021,LjungDeep2020,Telgarsky2015} 
	Advantages for specific classes, like power and Lipschitz functions, are also
	discussed in \cite{Daubechies2021,LjungDeep2020,Telgarsky2015,YarotskyA2017,YarotskyB2017}.
	%Such approximation capabilities could be one of the factors explaining why 
	%hierarchical learning systems show superior performance in some engineering applications.
	%Such approximation capabilities are relevant  for us %for control community also 
	This explains the potential for a better bias/variance trade-off also for system identification, by facilitating the search for flexible black-box models with as few free parameters as possible.
	But there is another phenomenon that complicates this scene regarding \emph{overparameterized models}. % that complicates the scene.
	Some neural networks (as well as some kernel machines/shallow networks) can almost perfectly fit the data while still performing well on validation data~\cite{ZhangRethink2021}. The survey will include recent insights on this, which leads to \emph{double-descent curves}: beyond a certain threshold, prediction capability may improve as the model complexity increases, apparently contradicting Occam's razor principle \cite{bartlett_benign_2020,bartlett_deep_2021,Belkin2018,hastie2022surprises,Liang2020,PoggioPNAS2020,Ribeiro2021}.
	%It has been in part explained by considering that overtraining obtained e.g. using (stochastic) gradient descent
	%may lead to minimum-norm solutions %(pseudo-inverses in Hilbert spaces when using kernel-based methods) 
	%\cite{Yao2007,ZhangRethink2021}.  
	%All of these arguments are also connected with 
	These studies are also connected with the so called implicit regularization in neural networks, the equivalence between infinitely wide neural networks and Gaussian process \cite{rahimi_random_2008}, the neural tangent kernel regime \cite{jacot_neural_2018}, the loss landscape and 
	the delicate link between number of parameters and model flexibility/effective dimension/capacity
	\cite{Abbas2021b,LjungDeep2020,SchoukensLjung:2019,ZhangRethink2021}. 
	
	{\bf{Outline}} We start by an overview of how dynamical systems are modelled. This is done in Section~\ref{sec:modeling-dynamical-sys} by describing the three essential decisions underlying system identification. Concepts like bias-variance trade-off and model generalization capability are also discussed. 
	Section~\ref{sec:DNN architectures} reports important deep networks architectures, like feedforward, convolutional, and recurrent networks.
	Optimization and regularization issues are discussed in Section
	\ref{SectionDeepOpt}. 
	Kernel-based methods and their inclusion in deep architectures, leading to deep kernel-based learning, are covered in Section~\ref{SectionDeepKernel}. Certain theoretical insights that shed some light on the success of deep models are illustrated in Section~\ref{sec:theoretical development}.
	Applications are reported in Section \ref{SectionDeepApp}.
	Conclusions then end the survey.
	%The final part of the survey will touch on some  open problems at the intersection of system identification and deep learning, like for example 
	%the use of sequential models with longer memories for applications where the exponential decay assumption is not applicable.
	
	\section{Modeling of dynamical systems}
	\label{sec:modeling-dynamical-sys}
	%LL text begin
	\subsection{Three main players}
\begin{figure}
		%\begin{center}
		%	\begin{tabular}{c}
		%\hspace{.1in}
		{ \includegraphics[scale=0.54]{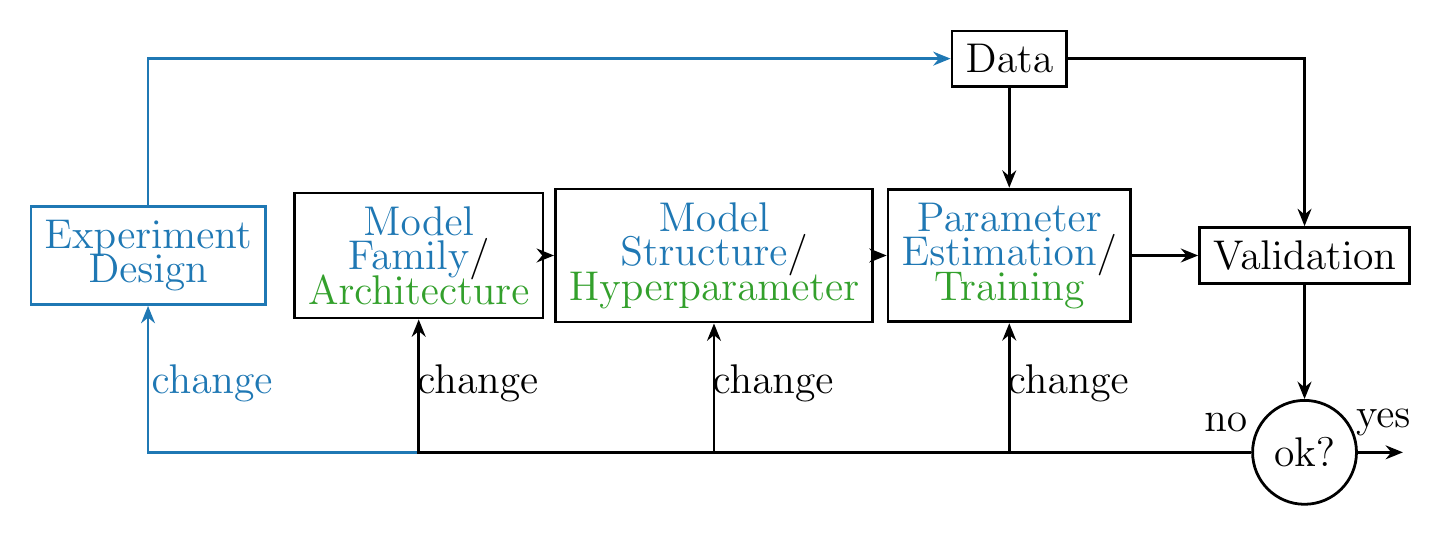}}   
		%	\end{tabular}
		\caption{{\emph{System Identification and Deep Learning
					pipelines.} Both represent methodologies for obtaining data-driven
				models and contain many similarities. In the classical framework, \emph{System
					identification} (in blue) usually starts from a family of models containing
				(often nested) parametric structures, as seen in the left part of Fig.~\ref{Fig1}.
				Some method is used to select one of these structures and the parameters are then estimated by an optimization procedure to adjust the structure to the data.  \emph{Deep learning} (in green) starts from a choice of a neural network architecture. Usually, there are many design choices and configurations that can be adjusted using the so called hyperparameters. The hyperparameters are selected and the model weights of the neural network are usually optimized to fit the dataset (called training) often using
				a first-order stochastic optimization method. System identification has extensive studies on how to design experiments to yield good models. In deep learning the datasets are instead often pre-collected and curated while experiment design is usually not
				an integral part of the modelling procedure.}}
		\label{fig_SI}
	\end{figure}

	Classical system identification, as depicted in the left panel of Fig. \ref{Fig1}, is characterized by three essential decisions:
	\begin{enumerate}
		\item A \textbf{family of parameterized models}. Each model is defined as a mapping $g(\cdot)$ from past observed input-output data available at time $t$: $Z(t)$ to a prediction of the next output $y(t+1)$: $\hat y(t+1|\theta,Z(t))=g(\theta,Z(t))$. As the parameter $\theta$ ranges over a set $D_\theta$ the models define a \emph{model structure} or \emph{model family}.
		\item A \textbf{parameter estimation method}, by which the parameter is adjusted to the observed data. This can be done in two steps by first using hyperparameters to fine tune the model set. But the archetypical method is to determine the parameter to be the one that minimizes the error between the observed outputs and the ones predicted by the model: 
		\begin{subequations}
			\begin{align} \nonumber 
			&  \text{Fit to a dataset}\;Z: \\ \label{FitZ}
			& V_N(\theta,Z) =\frac{1}{N} \sum_{t=1}^N\|y(t)-\hat y(t|\theta)\|^2 \\ \nonumber
			&\text{Parameter estimate for an estimation dataset $Z_e$:}\\  \label{Estimate} %Z_e(N) \text{with $N$ data}\\
			&   \hat \theta_N = \arg \min_{\theta \in D_\theta} V_N(\theta,Z_e).%\sum_{t=1}^N |y(t)-\hat y(t|\theta)\|^2
			\end{align}
		\end{subequations}
		The value % value %of the fit 
		\begin{equation}\label{EmpiricalFit}
		V_{\text{emp}}=V_N(\hat \theta_N,Z_e)
		\end{equation}
		is called the \emph{empirical fit}. % is then given by 
		
		%$V_N(\hat \theta_N,Z_e)$ is called the \emph{empirical fit}. 
		\item A \textbf{validation process} by which the model  $\hat y(t|\hat \theta_N)$ is validated or falsified.
	\end{enumerate}
	%Some comments of model validation will be given later \dg{specify where? Right in the next subsection or even later as well?}. 
	Often the estimated model will not be accepted and then earlier decisions, in particular regarding the model structure, will have to be revised. Hence, system identification will typically be ``a loop", as depicted in Fig.~\ref{fig_SI}.
	\subsection{Number of parameters -- a blessing and a curse (bias and variance)}
	Denote the number of parameters in the model structure by
	\begin{align}
	m = \dim \theta.
	\end{align}
	Typically, the model structures considered are nested in increasing complexity (flexibility) that have more and more parameters (think of linear models of increasing orders). It is trivial that the empirical fit is monotonically non-increasing as a function of $m$ (we are minimizing over larger and larger parameters sets). Eventually we will reach a perfect fit to the estimation data, %$V_N(\hat \theta_N,Z_e)=0$ 
	$V_{\text{emp}}=0$ (typically when $m=N$). What happens is that more parameters give more model flexibility and allow  a better fit between model and data. They thus give less \emph{bias}. But more parameters also allow adjustments to the particular disturbance features in the estimation data $Z_e$. The improved fit due to such adjustments (\emph{overfit}) has no value - it is actually harmful when the model is tested on a new dataset $Z_v$. The effect of the particular disturbance features in the estimation data $Z_e$ is captured by the statistical concept of \emph{variance} of the model.%The improved fit due to such adjustments (\emph{overfit}
	
	%There is a remarkable general result that says that the %variance is proportional to the number of estimated %parameters $d$, symbolically written
	
	\paragraph*{Model Validation}\label{par_mv}
	There are many techniques to validate and falsify models.
	One is \emph{residual analysis}. The prediction errors for the chosen model $\varepsilon(t)=y(t)-\hat y(t|\hat \theta_N)$ (the \emph{residuals}) are computed and tested if they are independent of each other and of the input. Any significant correlation will falsify the model.
	Another method is \emph{cross validation} where one collects an independent validation dataset $Z_{\text{v}}$ and computes the fit 
	\begin{equation}\label{ValidationFit}
	V_{\text{valid}}=V_N(\hat \theta_N,Z_{\text{v}}).
	\end{equation}
	% (recall that $V_{emp}=V_N(\hat \theta_N,Z_e)$ is instead the empirical fit).
	%$V_M(\hat y(t|\hat \theta_N,Z_v(M))$. 
	This should give an acceptable fit. If you have several candidate models, your selected model could be the one that has the best fit to validation data.
	
	\paragraph*{Choice of Model Order and Bias-Variance Trade-off} As you compute models of increasing orders, you will find that the model error initially decreases since more parameters will first improve the fit due to smaller bias. However, with more parameters, the variance effect %(\ref{eq:variance}) 
	will eventually dominate and the model error starts to increase. A curve showing the model error as a function of the number of parameters will thus
	typically be U-Shaped with the minimum at the best bias-variance trade-off. An important step in system identification is to find the sweet-spot in this trade-off.
	
	\paragraph*{Variance and Number of Parameters - Traditional Truths}
	The number of estimated parameters
	%\begin{align}
	$m = \dim \theta$
	%\end{align}
	and the number observations in the estimation data
	%\begin{align}
	$N =  \dim Z_{\text{e}}$
	%\end{align}
	are key concepts when studying the model quality, and has been a central subject in system identification, see e.g. Section 16.4  in \cite{Ljung:99}. Such analysis is normally based on asymptotic statistical results as $ N\to \infty$ and typically assuming that $m << N$.\\
	The observed data are subject to stochastic disturbances, which means that the estimate $\hat \theta_N$ 
	in \eqref{Estimate} will be a random %(column) 
	vector with covariance matrix denoted by $P_\theta$.
	% Its mean and variance are as usual defined by
	%\begin{align}
	%    \theta_m = \mathcal{E} \hat \theta_N, \qquad P_\theta = \mathcal{E}(\hat \theta_N-\theta_m)(\hat \theta_N-\theta_m)^\top.
	%\end{align}
	Note that $P_\theta$ is of interest only if the minimum $\hat \theta_N$ is uniquely defined (``the parameter is identifiable"). This requires  that the $N|m$-dimensional gradient $\frac{d}{d\theta}\hat y(t | \theta)$, with $t=1,\dots,N$,  has full rank at
	the minimum (which certainly requires $m\le N$). We can also find an important asymptotic expression for such covariance.
	Let %ting 
	$\bar{V}$ denote the expected loss function, defined for any $\theta$ by
	\begin{equation}\label{barV}
	\bar{V}(\theta) = \mathcal{E}  V_N(\theta,Z_{\text{e}}).
	\end{equation}
	%and $\nabla^2_{\theta}(\bar{V})$ its Hessian.
	Then, assuming that our structure contains the ``true" system described by $\theta^*$, 
	the minimum of \eqref{barV} is $\bar{V}(\theta^*)$ and 
	equals the noise variance $\sigma^2$ in the estimation data. Furthermore,  
	%that has generated the data
	%(that also is the minimizer of $\bar{V}$)
	typically %for large $N$ one has
	\begin{align}\label{eq:pvar}
	%P_\theta \sim 2 \bar{V}(\theta^*) \frac{[\nabla^2_{\theta} \bar{V}(\theta^*)]^{-1}}{N}.
	P_\theta \sim 2 \sigma^2 \frac{[\nabla^2_{\theta} \bar{V}(\theta^*)]^{-1}}{N}.
	\end{align}
	%where $\sigma^2$ is the mean squared prediction errore using $\theta^*$ ().
	Let us ask ``the generalization question",  i.e. how the model reproduces new (validation) data from the same system according to 
	\eqref{ValidationFit}. If $Z_{\text{e}}$ and $Z_{\text{v}}$ follow the same (second-order) statistics,
	the prediction capability can be assessed by inserting the model $\hat{\theta}_N$ into the expected loss function
	%$\bar{V}$ in 
	\eqref{barV}. 
	From \eqref{ValidationFit} one thus has $\mathcal{E} \bar{V}(\hat \theta_N)=\mathcal{E} V_{\text{valid}}$
	and we find as in (16.30a) in \cite{Ljung:99},
	\begin{align}\label{eq:630} \nonumber
	\mathcal{E} V_{\text{valid}}  & \approx \bar V(\theta^*) + \frac{1}{2} \rm{trace} [\nabla^2_{\theta} \bar{V}(\theta^*)P_\theta] \\
	& = \sigma^2 \big(1 +  \frac{m}{N} \big). %\bar{V}(\theta^*) \frac{d}{N}.
	\end{align}
	Here we directly see how the model parameter variance affects the model's generalization properties 
	(ability to reproduce the output in new validation sets) by increasing the noise in the estimation data 
	through the additional term $m/N$. This phenomenon can also be related to the empirical fit \eqref{EmpiricalFit}, resulting in 
	\begin{align}\label{eq:625} %\nonumber
	% \mathcal{E} \bar{V}(\hat \theta_N) \approx \mathcal{E} V_N(\hat \theta_N,Z_e) +  2\sigma^2 \frac{d}{N}. %\bar{V}(\theta^*) \frac{d}{N}.
	\mathcal{E} V_{\text{valid}} \approx \mathcal{E} V_{\text{emp}} +  2\sigma^2 \frac{m}{N}. 
	\end{align}
	All of these formulations are general features in traditional analysis: 
	\emph{The model variance effects will be proportional to $m/N$}. It is essentially such results that
	are behind the celebrated Akaike criterion for model structure
	selection (AIC).\\

	%A remarkable
	%result in estimation theory, see, e.g.~Section 16.4 in
	%\cite{Ljung:99}, tells us that if the variance is evaluated for data of
	%the same character as the data used for estimation, then % the error can be computed as, symbolically written
	%\begin{align}
	% \label{eq:variance}
	%E\|  \hat{\mathfrak{m}} - \mathfrak{m}^*\|^2 = \frac{d %\sigma^2}{N}
	%\end{align}
	%w%here $N$ is the number of data points in the dataset %f%or estimation,
	%$\sigma^2$ is the variance of the noise component in the %data, and $d$
	%is the number of estimated parameters in the model %structure,
	%\emph{regardless of its character}. It is essentially this result that
	%is behind the celebrated Akaike criterion for model structure
	%selection (AIC).
	
	%It is the variance of the model that affects its \emph{generalization} properties, i.e. how well the model can reproduce the output in new (validation) sets.
	
	To summarize, the important implication of (\ref{eq:630}) is that many parameters are bad for the model's generalization properties -- no matter what. This has lead to the common basic principles in estimation and system identification: Occam's razor (cut away unnecessary parts of a model), the Principle of Parsimony (choose the most parsimonious model structure), Popper's thesis (accept the simplest unfalsified model).
	
	But a word of caution: recall that this expression is a result in conventional \textit{asymptotic} statistical analysis, as $N \to \infty$ including assumptions that $m < N$. %The derivation, cf (16.31) in \cite{Ljung:99} requires that the $N|d$ gradient of $V_N$ be full rank (implying $d < N$). 
	So for very big parameterizations with $m>N$ the results do not hold. This is what makes ``the double-descent" phenomenon discussed in Section~\ref{sec:double-descent} possible.
	%\begin{align}
	%   E\|\hat 
	%\end{align}
	\subsection{Model flexibility and number of parameters}
	It follows from the above that ideal model structures achieves high and relevant flexibility with a small number of parameters. This, as such, points to gray-box model structures where parameters with physical significance are used in models based on physical modeling. 
	For black-box models it is  a matter of finding very flexible model structures with cleverly chosen parameterizations. This can in fact be seen as the main drive for deep learning.  Consider the simple task of finding a good structure for a simple curve fit. If we work with piece-wise linear models, the flexibility can be measured as the number of breakpoints in the model curve. A simple one-layer structure will require $m$ parameters to achieve $m$ breakpoints. We will see later on in the survey in Section~\ref{sec:theoretical development} how multi-layer structures can be much more powerful in creating (many) more breakpoints with few parameters, see also e.g.~\cite{Daubechies2021} and \cite{LjungDeep2020}. 
	%%% End LL Text	 

	\section{Deep neural networks architectures}
	\label{sec:DNN architectures}

	% ----------- Start AHR text ----------- %

	Artificial neural networks are  a class of nonlinear functions. They map the input $x(t)$ to a predicted output $\hat{y}(t)$ and are defined in a way that is loosely inspired by the way biological neurons work.  Neural networks can have multiple layers and can be seen as a composition of functions. A neural network with $L$~layers is defined by the following compositional map:
	\begin{equation}\label{DeepStart}
	f = f^L \circ \ldots \circ f^2 \circ f^1,
	\end{equation}
	where $f^L: \R^{n_L} \mapsto \R^{n_{\text{out}}}$ is called the output layer and $f^{L-1}, \cdots, f^1$ are called intermediate layers, $f^{\ell}: \R^{n_{\ell}} \mapsto \R^{n_{\ell+1}}$. The name ``hidden layer" is sometimes also used to mean that changes in these layers affect the output only indirectly: by modifying the input to the subsequent layers. We will denote the dimension of the $\ell^{\text{th}}$ hidden layer---the number of nodes or neurons in that layer---by  $n_\ell$ and the dimension of the first layer by $n_1 = d$. The word deep is here commonly used to refer to neural networks with a large numbers of layers.

	% WIP = Work in progress
	% Still need to make  the notation uniform with the others
	\subsection{Fully connected neural networks}
	\label{sec:fully_connected}
	
\begin{figure}[htb]
		\centering
		\includegraphics[width=0.75\columnwidth]{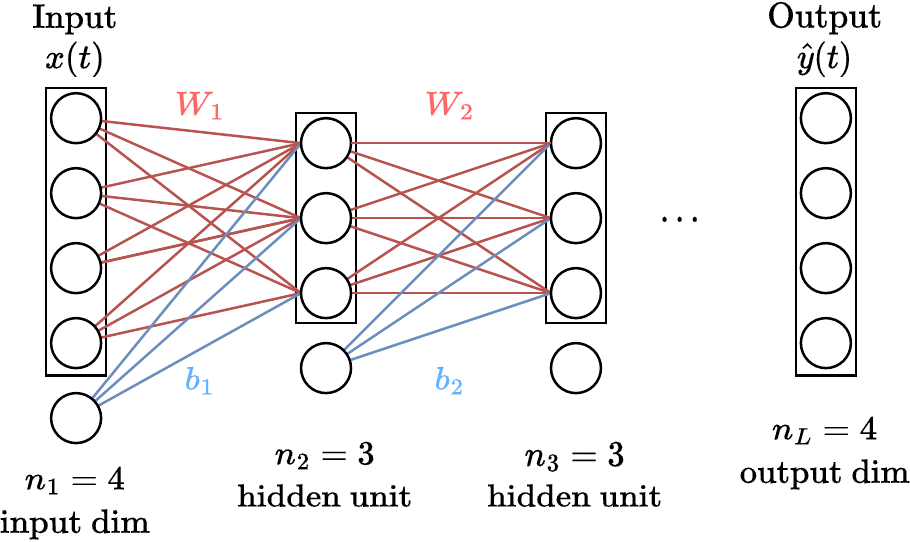}  
		\caption{Fully-connected neural network with hidden layers and dimensions.}
		\label{fig:full-connected-network}
	\end{figure}

	Fully connected neural networks (sometimes also called multilayer perceptron, MLP, or feedforward neural networks) were among the first architectures to appear. We present them here both to introduce the notation and to discuss some notion that appear when applying these models in system identification. 
	% \dg{GP: is it maybe useful for the reader to add a figure of a fully connected neural network with the caption that explains what is a layer, what is the width of a layer, notation like  n sub ell that will be used to indicate the dimension of the layer... }
	Fig.~\ref{fig:full-connected-network} depicts a fully connected neural network with input dimension $n_1=4$ of the input $x(t)$ and $L-1$ hidden layers. We show the weights $W_\ell$ and biases $b_\ell$ distinctly.
	
	\paragraph*{Fully connected neural networks}
	A fully connected neural network with one hidden layer is defined by
	\begin{equation}
	\label{eq:fully_connected}
	\hat{y}(t) = W_2  \sigma(W_1 x(t) + b_1)  + b_2,
	\end{equation}
	and could be decomposed as $f = f^2 \circ f^1$ which are called layers of the neural network: $f^1(z) = \sigma(W_1 z + b_1)$ and $f^2(z) = W_2 z + b_2$. In \eqref{eq:fully_connected} we are considering a dynamic setting and $t$ denotes time.

\begin{figure}[htb]
		\centering
		\includegraphics[width=0.9\columnwidth]{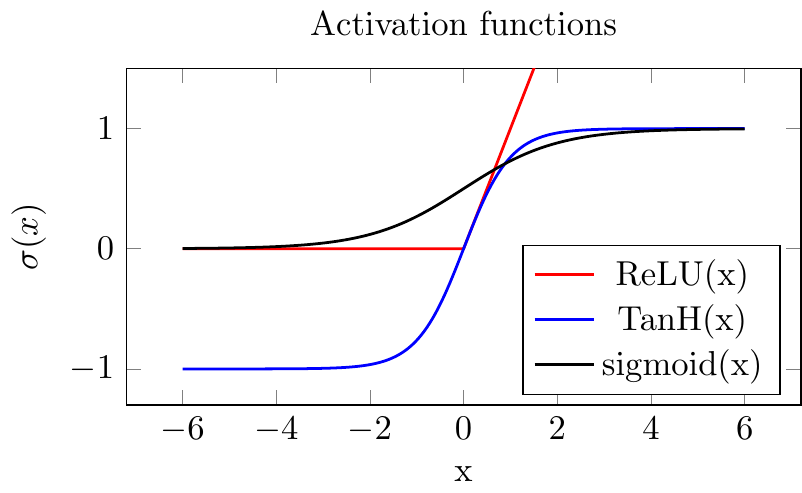}  
		\caption{\emph{Activation functions.} Visualization of three popular activation functions.}
		\label{fig:activation}
	\end{figure}

	In fully connected networks, layers are affine transformations followed by nonlinear transformations: $f^{\ell}(z) = \sigma(W_{\ell} z + b_{\ell})$, where $W_{\ell} \in \R^{n_{\ell+1} \times n_{\ell} }$ are weight matrices, $b_{\ell} \in \R^{n_{\ell + 1}}$ are bias vectors and $\sigma$ is an \textit{activation function}, i.e., a nonlinear function that act element-wise on its inputs. Examples of commonly used activation functions are the so  called ReLU activation function $\sigma(x) = \max(x, 0)$, the hyperbolic tangent $\sigma(x) = \tanh(x)$ and the sigmoid function $\sigma(x) = 1 / ( 1+\exp(-x))$. The unknown parameters of the neural network are the weight matrices $W_i$ and the biases $b_i$, for $i = 1, 2,\dots, L$. Hence, we denote $\theta = (W_1,  b_1, W_2,  b_2, \cdots)$. Fig.~\ref{fig:activation} shows the three described activation functions.
	
	\paragraph*{Neural network input}
	We use the word \emph{input} both when referring to the input~$x(t)$ to the neural network and when referring to the input~$u(t)$ to the dynamical system being estimated. They might be different since the model might benefit from not only using the input of the dynamical system as an input to the model, there might very well be benefits in using additional signals as inputs to the model. Indeed, how to define $x(t)$ can give rise to different architectures. One natural option is to define $x(t)$ as a concatenation of past inputs up to a time $m$,
	\begin{equation}
	\label{eq:nnfir}
	x(t)=[u(t),\ldots,u(t-m)]^\top,
	\end{equation}
	and this brings about the so-called ``Finite Impulse Response" (FIR) neural network.  Another alternative is to use ``AutoRegressive with eXogenous input" (ARX) structure. In the latter case, the input to the neural network would also include the past measured outputs, i.e.
	\begin{equation}
	\label{eq:nnarx}
	x(t)=[ y(t-1),\ldots,y(t-m), u(t),\ldots,u(t-m-d)]^\top,
	\end{equation}
	where $m$ still denotes system memory and $d$ can be used to consider a possible delay in the action of the input to the output.

	\paragraph*{Long-term predictions}
	We notice that using an ARX neural network to make long-term predictions requires additional considerations. Past output values need to be fed into the neural network to make the prediction and these will not be available during multiple-steps-ahead predictions. 
	
\begin{figure}
		\includegraphics[width=0.45\columnwidth]{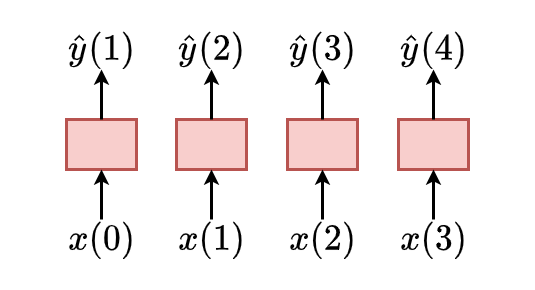}  
		\includegraphics[width=0.45\columnwidth]{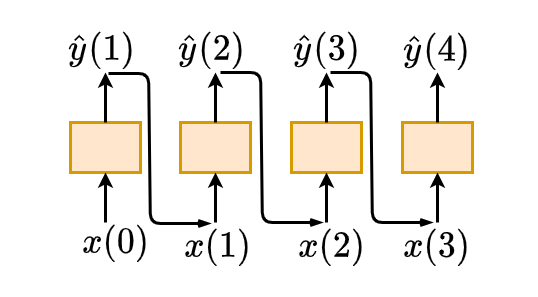}   
		\caption{{\emph{Left:} configuration used during training of an ARX neural network. \emph{Right:} configuration for long-term predictions.}}
		\label{fig:narx-configurations}
	\end{figure}

	Indeed, long-term prediction is
	more difficult
	since regressors need to be defined using previously estimated outputs \cite{SchoukensLjung:2019}. For concreteness, let $m=2$,
	If $f$ is our estimated ARX neural network,
	the following recursions are typically adopted
	\vspace{-15pt}
	\begin{small}
		\begin{eqnarray}\label{LongTerm} 
		\hat{y}(t+1 |t) &=& f(y(t),y(t-1), u(t+1), u(t)), \\ \nonumber
		\hat{y}(t+2|t) &=& f(\hat{y}(t+1| t),y(t),u(t+2),u(t+1)), \\ \nonumber
		&\vdots&\\ \nonumber
		\hat{y}(t+k+1|t) &=& f(\hat{y}(t+k|t),\hat{y}(t+k-1|t),u(t+k+1),u(t+k)),
		\end{eqnarray}
	\end{small}
	where we use the notation $\hat{y}(t+k+1|t)$ to denote the prediction of the output at time $t+k+1$ using measured outputs up to time t. Hence, for any $k>m$ the input to the neural network in our example would be $$x(t+k+1)=[\hat{y}(t+k|t),\hat{y}(t+k-1|t),u(t+k+1),u(t+k)]^\top,$$ where the measured values $y$ are all replaced by the predicted values $\hat{y}$ that were recursively computed. This is illustrated in Fig.~\ref{fig:narx-configurations}.
	
	When the ultimate goal is to make long-term predictions, one natural idea that appears is to use the feedback connection illustrated to the right of Fig.~\ref{fig:narx-configurations} also during training. This gives rise to output error (OE) neural networks. The name originates from the fact that this model can be derived by considering the output error (also called measurement error).

	Fully connected neural networks have been used in system identification from early on~\cite{Chen1990,narendra_identification_1990}. Regularization strategies for FIR neural networks are discussed in~\cite{SchoukensNFIR2021}. The advantages and disadvantages of OE and ARX neural networks are discussed in the pioneering work of~\cite{narendra_identification_1990} and, more recently, in~\cite{aguirre_prediction_2010,ribeiro_parallel_2018}. On the one hand, OE neural networks may be more suitable to deal with output error noise (hence the name)~\cite{ribeiro_parallel_2018}, however, they might also yield more challenging optimization problems~\cite{narendra_identification_1990,ribeiro_smoothness_2020}. OE neural networks can be understood as a type of recurrent neural network, which will be the focus of Section~\ref{sec:recurrent-nn}.

	\paragraph*{Stability}
	
	Numerical problems can arise (prediction can diverge)
	and stability of the one-step ahead predictor is not sufficient to prevent them. 
	In the linear setting, stable models can be enforced \cite{DNM-RAC:13,Romeres2015} using linear matrix inequalities (LMIs) 
	\cite{CM-GP:96} while other stabilizing optimization methods 
	can be found in \cite{Pillonetto2022SSS,Manch2018}.
	In the nonlinear setting, a first approach to favour system stability
	is to design estimators that include fading memory concepts. This means that
	one should expect the inputs to have less and less influence on future outputs as the lag increases \cite{Birpoutsoukis2017,Pillonetto:11nonlin}.
	Neural networks for nonlinear FIR and ARX models can be found in \cite{SchoukensNFIR2021,zancato2021novel} where it is also shown that
	fading memory information may act as an important regularizer to avoid overfitting. A second approach is to obtain structural conditions on the architecture 
	that ensure some form of stability \cite{Barabanov2002,Lisang2012}, then possibly including them in the estimation process. 
	This problem is far from trivial since stability in nonlinear systems is multifaceted.
	For input-output models, the seminal work \cite{Zames1966} 
	requires bounded inputs to lead to bounded outputs 
	and also that small input variations do not produce large changes in the output. 
	These properties are not implied by the stability of a particular equilibrium
	\cite{Sontag2008} but are instead guaranteed 
	by contraction \cite{Lo1998}  and incremental stability (IS) \cite{Angeli2002,Desoer1975} 
	(neither of them depend on the input or equilibrium). 
	In particular, IS ensures that two system trajectories, defined by the same input,
	will converge to each other starting from any initial condition.
	Recall also that stability analysis via input-output or state-space approaches
	are not equivalent in nonlinear systems.
	Sontag's concept of Input-to-state stability (ISS) \cite{Sontag2008} 
	merges the two different views also including
	Lyapunov-like stability notions \cite{Jiang2001,Sontag1989}.
	The so called Incremental-ISS ($\delta$ISS) is also important 
	in control since it can ensure that the distance between any
	two system trajectories can be bounded by the distance of
	disturbances and initial states.
	
	ISS and $\delta$ISS are important requirements in applications like 
	autonomous driving \cite{Zyner2017}, model predictive control \cite{Bayer2013,Bonassi2021}, 
	safety verification \cite{Bonassi2020LSTM,Tarraf2019} and moving horizon estimation \cite{Alessandri2008}.
	Model stability can also influence the training of neural networks, due to the exploding and vanishing gradients problem discussed in Section \ref{sec:recurrent-nn}, see e.g.
	\cite{pascanu_difficulty_2013,Revay2020}. It also has connections with generalization properties e.g. in classification tasks \cite{Zhang2018}. 
	ISS and $\delta$ISS are hard to include as priors in a system identification procedure.
	Quite recent results on ISS and $\delta$ISS for recurrent networks 
	can be found in  \cite{Bonassi2021Gated,Bugliari2019,Miller2018,Stipanovich2021,Terzi2021},
	see also \cite{Revay2020} for convex parametrizations of stable recurrent models through contraction analysis. 
	Deep ARX nonlinear network models have been studied in continuous- and discrete-time, respectively, see \cite{Sanchez1999Stability} and \cite{BonassiStability2021}.
	In the latter, a feedforward neural network is reformulated as normal state-space canonical form. The authors then obtain a sufficient condition for ISS in terms of an inequality that depends on networks weights and Lipschitz constants of the activation functions, see  \cite[Theorem 4]{BonassiStability2021}. 
	%\subsection{Designing stability for long-term predictions}%deep prediction networks}
	%\label{sec:stability-deepkernel}

	\subsection{Skip and direct connections}\label{sec:cascade}

	\begin{figure}[htb]
		\begin{center}
			\includegraphics[width=0.35\columnwidth]{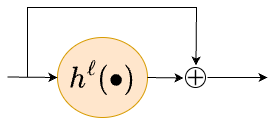}  
			\includegraphics[width=0.35\columnwidth]{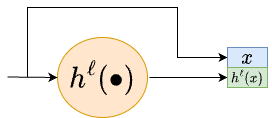}
		\end{center}
		\caption{{\emph{Left:} Skip connection~\eqref{eq:residual-cnn}. \emph{Right:} Direct connection~\eqref{eq:direct-connection}.}}
		\label{fig:connections}
	\end{figure}
	
	Additional connections between the layers can facilitate the training, two popular approaches are illustrated in Fig.~\ref{fig:connections}. Perhaps the most popular strategy is the use of skip connections, which constitute an important part of the popular ResNet architecture~\cite{he_deep_2016}. Let~$x$ be an intermediate signal and let~$h^\ell$ be a transformation (for instance a fully connected layer discussed in the previous section). A layer with \emph{skip connection} then has the form 
	\begin{equation}
	f^{\ell}(x) = h^{\ell}(x) + x.
	\label{eq:residual-cnn}
	\end{equation}
	Here, the output of the layer is the sum of the output of the nonlinear transformation and the input. The purpose of this extra ``connection" is to let the layers learn deviations from the identity map rather than the entire transformation. This property is beneficial, especially in deep networks where we want the information to gradually change from the input to the output as we proceed through the layers.
	
	With a similar purpose,  a layer can propagate the concatenation of the transformed value with the un-transformed input. We say that it adds a \emph{direct connection} from its input to the output. This connection extend the dimension of the layer,
	\begin{equation}
	\label{eq:direct-connection}
	f^{\ell}(x) = \begin{bmatrix}
	x\\
	h^{\ell}(x) 
	\end{bmatrix}.
	\end{equation}
	Multiple layers with this design yield a densely connected network, where for each layer, the feature-maps of all preceding layers are used as inputs, and its own feature-maps are used as inputs into all subsequent layers.
	An example of this design choice is found in Cascade Networks~\cite{fahlman1989cascade}.
	This choice also appears in other contexts, for instance in the DenseNet~\cite{huang_densely_2016} architecture commonly used for image classification.
	
	Skip connections give an interpretation of another recently emerging idea.
	A way to see the skip connection $x^{\ell+1}= h(x^\ell) + x^\ell$ is as a discretized
	version of the following ordinary differential equation (ODE): $\frac{dx}{d\ell} = h(x).$
	Neural ordinary differential equation (NODE) \cite{chen2018neural}  use this view to implement deep neural networks. 
	The network will output the solution to this ODE parametrized as neural network. Any numerical ODE solver can be used here by considering the solver as black box and utilizing the adjoint sensitivity method \cite{pontryagin1987mathematical}. The formulation of neural networks as ODEs are particularly interesting for systems which can naturally be described with ODEs such as dynamic systems. There are initial attempts to use NODEs for system identification \cite{quaglino2019snode} with a comparison of different approaches in \cite{rahman2022neural}. Software developments to simplify the use of NODEs are available \cite{poli2020torchdyn}.

	\subsection{Convolutional neural networks}
	
	Convolutional neural networks (CNNs)~\cite{lecun_gradientbased_1998} had a very strong impact on computer vision, achieving state-of the-art-results in tasks such as image classification~\cite{krizhevsky_imagenet_2012}, segmentation~\cite{long_fully_2015} and object detection~\cite{redmon_you_2016}. Interestingly, it has recently been shown that the architecture is also highly useful for tasks with sequences as input~\cite{bai_empirical_2018}. More specifically, recent results show that the convolutional architecture can match recurrent architectures in language and music modelling~\cite{bai_empirical_2018,dauphin_language_2017,oord_wavenet_2016}, text-to-speech conversion~\cite{oord_wavenet_2016}, machine translation~\cite{gehring_convolutional_2017,kalchbrenner_neural_2016} and other sequential tasks~\cite{bai_empirical_2018}. This architecture have also been studied for system identification~\cite{andersson_deep_2019}. In this section, we focus on uni-dimensional convolutional neural networks for sequential tasks, trying to build the basic convolutional neural network blocks from signal processing and dynamical system theory.

	\paragraph*{The uni-dimensional convolutional layer}
	
	Convolutional neural networks have as building blocks some basic operations that are well known in system identification. Here, we present a uni-dimensional version of the convolutional layer. For images, two dimensional implementations would be used instead. The \textit{convolutional layer} outputs a filtered version of the output by applying the convolution operation
	\begin{equation}
	w(t) * x(t)  =  \sum_{j=0}^{k-1} w(j)^T x(t-j),
	\end{equation}
	where $w(j), j= 0, \dots, k-1$ is a vector of weights.
	Most often, a convolutional layer consist not only of a single filter, but rather a bank of filters. If we denote the $b$ different filters by $W = \{w^{1}, \dots, w^{b}\}$, the $i^{\text{th}}$ output of the convolutional layer is the concatenation of $z^i(t) =  w^i(t) * x(t)$ for $i = 1, \dots, b$. Hence, a convolutional layer can be seen as a bank of filters with adjustable coefficients. If the input $x(t)$ is a vector of  dimension~$d$, the total number of learnable parameters within this bank of filters is $d \times b \times k$. The usual nomenclature in convolutional neural networks literature is to refer to~$k$ as the \textit{kernel size}, $b$ and $d$ as the \textit{number of channels} of the input and output, respectively.
	
	A convolutional neural network commonly consists of the concatenation of layers that usually include convolutional layers, element-wise nonlinear transformations and downsampling operations. We note that if only convolutional layers and activations were present, the use of CNNs for time series is actually equivalent to the use of ARX and FIR fully connected networks, which were described in  Section~\ref{sec:fully_connected}. For instance:
	\begin{equation}
	\label{eq:conv_2layers}
	W_2(t) * \sigma(W_1(t) * x(t) + b_1)  + b_2
	\end{equation}
	is equivalent to~\eqref{eq:fully_connected}. If for example $x(t) = u(t)$, then the above is actually equivalent to the FIR fully connected neural network we described before. Let us define a matrix containing all the filter coefficients
	\[
	M_i = \begin{bmatrix}
	w^1_i(0)^T & & w^1_i(k-1)^T\\
	\vdots & \ddots&   \vdots \\
	w^b_i(0)^T & & w^b_i(k-1)^T
	\end{bmatrix}
	\]
	for $i = 1, 2$. Then~\eqref{eq:conv_2layers} is actually equivalent to:
	\begin{equation}
	M_1\sigma\left(M_2
	\begin{bmatrix}
	u(1) \\
	\vdots \\
	u(k)
	\end{bmatrix}
	+ b_1 \right)  + b_2.
	\end{equation}
	
	Similarly if $x(t) = [u(t), y(t-1)]$ the convolutional neural network is actually equivalent to the 2-layer fully connected neural network ARX model described in Section~\ref{sec:fully_connected}. Such similarities are pointed out in~\cite{andersson_deep_2019}. 
	
	\paragraph*{CNN architectures}
	CNNs were first conceived for image classification problems~\cite{lecun_handwritten_1990} and many famous architectures were designed with such problems in mind. Examples include the ResNet~\cite{he_deep_2016}, Inception~\cite{szegedy_going_2015}, DenseNet~\cite{huang_densely_2016}, EfficientNet~\cite{tan_efficientnet_2019}.  
	In the case of dynamical system models, as we pointed in the previous section, convolutional layers just present a different view on the standard way neural networks are applied to system identification: A rolling window of inputs (and, possibly, measured outputs) is fed into the neural network. A key difference, however, is that practical convolutional neural network architectures also include intermediary down-sampling operations. These allow them to encode longer dependencies.
	
	Downsampling (or decimation) is a common pre-processing operation in signal processing. Decimating a given sequence $x(t), t = 1, \dots, N$ by a factor of~$q$ implies that we only keep every $q^{\text{th}}$ sample and discard the others. Mathematically, this would correspond to a new sequence defined as ${\tilde x(t) = x(qt)}$ for $t = 1, \dots, \left\lfloor N / q\right\rfloor$. If the original signal was sampled with frequency $\omega$, the new signal after downsampling will now be sampled with a lower frequency $\omega/q$. 
	
	In neural networks downsampling appears not only as a pre-processing stage (as it is standard practice in system identification), but also as intermediate layers. Here, we mention three types of operations that implement downsampling in  a CNN: \textit{strides}, \textit{dilations} and  \textit{pooling}. Most available implementations include \textit{strides} and \textit{dilations} as parameters that are passed to the convolutional layer; \textit{pooling} on the other hand is usually implemented as a separate block. A convolutional layer with \textit{strides}=$q$ downsamples the output signal by a factor $q$ after the convolution operation. Both operations are often jointly implemented. A \textit{max pooling layer} takes the maximum among every $q$ samples, i.e. ${\tilde x(t) =\max(x(q t), x(q t - 1), \dots, x((q-1)t)}$, whereas the \textit{average pooling layer} take the average. 
	
	Dilations and pooling are used in many of the most popular convolutional neural networks in computer vision~\cite{he_deep_2016,huang_densely_2016,szegedy_going_2015,tan_efficientnet_2019}. The purpose is that as we progress through the neural network, the input is progressively downsampled and larger portions of the input are actually encoded by features of the neural network and deeper layers deal with increasingly complex features. For instance, in the image classification problem, the first layers would implement edge or texture detectors, intermediate layers would look for contours and final layers would look at larger regions of the image and try to match for specific parts of the  objects~\cite{goodfellow_deep_2016,olah2017feature}. 
	
	Many of these architectures were first conceived for cases where there was a single output. For example in image classification applications: the probabilities of  the image being one of the classes under consideration.  This is not the case when dealing with dynamical systems, where we are interested in predicting an output sequence from the input sequence, with the output sampled at the same sample frequency as the input. 
	Dilations present an alternative that allow the convolutional transformations to be applied in this case (sometimes called sequence-to-sequence)
	where the input and output are sequences with the same length and sampling frequency~\cite{andersson_deep_2019,bai_empirical_2018,dauphin_language_2017,oord_wavenet_2016}.
	
	Dilation is another type of operation that implement downsampling in CNN.  The operation has its roots in the multirate signal processing literature. \textit{Polyphase} filters were introduced as an effective way to combine upsampling/downsampling operations with filtering and they are at the center of this line of study~\cite{oppenheim_discretetime_1999}. A signal $x(t)$ can be decomposed in its $q$ \textit{polyphase}  components, 
	\begin{equation}
	\label{eq:polyphase_decomp}
	\tilde{x}^{\kappa}(t) =  x(tq +i), \qquad t = 1, \dots, \left\lfloor\frac{N+i}{q}\right\rfloor,
	\end{equation}
	where each component $\tilde{x}^{\kappa}$ is the signal decimated after a shift of $\kappa = 0, 1, \dots, q-1$ samples. We can always decompose a signal into $q$~ \textit{polyphase}  components and these components can, in turn, be used to reconstruct the original signal. 
	
	Convolutional layers with dilations can be seen as polyphase implementations of a bank of filters with learnable parameters. Using the nomenclature we introduced for polyphase decomposition, a convolutional layer with dilations applies the following sequences of operations: (1) Decompose the signal $x(t)$ into $q$ different components using polyphase decomposition; (2) Apply the  convolutions to all polyphase components $\tilde{x}^{\kappa}(t)$ with the same convolutional weights. The result should be $q$ polyphase components $z^{\kappa}(t)$; (3) Reconstruct the output signal $z(t)$ from its $q$ polyphase components.

	\paragraph*{The aliasing effect}
	Aliasing is a well-known side-effect of downsampling that may take place: it causes high-frequency components of the original signal to become indistinguishable from its low-frequency components.
	While, in principle, CNNs are capable of implementing anti-aliasing filters, it has been observed that this does not happen in practice~\cite{ribeiro_how_2021}. However, redundancy in the intermediary layers can still succeed in distinguishing oscillations in the input~\cite{ribeiro_how_2021}.
	The rather natural idea of introducing anti-aliasing filters before the downsampling operations in CNNs has been explored recently, see e.g.~\cite{azulay_why_2019,vasconcelos_effective_2020,zhang_making_2019,zou_delving_2020}. This helps in making them invariant to translations~\cite{zhang_making_2019} and other transformations~\cite{azulay_why_2019}.
	
	\subsection{Recurrent neural networks}
	\label{sec:recurrent-nn}
	
	The recurrent neural network (RNN) is a neural network architecture that models dependence from past predictions that need to be computed recursively during training. 
	The OE neural network in Section~\ref{sec:fully_connected} is one instance of an RNN.
	Another example is the Elman RNN. For this model, the input $x(t)$ is related to the predicted output $\hat{y}(t)$ by means of two equations: a state-propagation equation
	\begin{equation}
	\label{eq:elman-state}
	h(t+1) = \sigma(W_{hh} h(t) + W_{hx} x(t) + b_h),
	\end{equation}
	and an output equation
	\begin{equation}
	\label{eq:elman-output}
	\hat{y}(t) = W_{yh} h(t) + W_{yx} x(t) + b_y.
	\end{equation}
	Here, the parameter vector that needs to be estimated is the concatenation of weight matrices and bias vectors, i.e.,~$\theta = (W_{hh}, W_{hx}, W_{yh}, W_{yx}, b_h, b_y)$.

	Another natural idea is to stack recurrent layers to obtain deep recurrent architectures. For instance, stacking two Elman recurrent layers yields
	\begin{subequations}
		\begin{align}
		h^{(1)}(t+1) &= \sigma(W_{hh}^{(1)} h^{(1)}(t) + W_{hx}^{(1)} x(t) + b_h), \\
		h^{(2)}(t+1) &= \sigma(W_{hh}^{(1)} h^{(2)}(t) + W_{hx}^{(1)} h^{(1)}(t) + b_h), 
		\end{align}
	\end{subequations}
	and the output can then be computed as in~\eqref{eq:elman-output}. The same idea applies to LSTM and GRU blocks (both of which will be presented next). In all cases, both deep and shallow, RNNs are nonlinear and deterministic state-space representations with learnable parameters, which can be represented using a state-propagation and an output reading equation.
	
	How to define the input can again result in models with different types of behavior. On the one hand, $x(t)$ can consist of past inputs, as in~\eqref{eq:nnfir}, or of past inputs and outputs, as in~\eqref{eq:nnarx}. Including auto-regressive terms can allow for complex dynamics to be learned while solving smoother optimization problems~\cite{ribeiro_exploding_2020} and is often used in practice. Most of the considerations discussed for fully connected neural networks apply here as autoregressive terms require different configurations during training and testing (similar to Fig.~\ref{fig:narx-configurations}).
	
	\paragraph*{Exploding and vanishing gradients} Neural network parameters are usually estimated by minimizing the fit to the dataset, i.e.  $V_N$ in~\eqref{FitZ}. The gradient plays a central role in this optimization. For RNNs, however, this gradient is often either blowing up or vanishing during training, which makes these models more challenging to find.

	The gradient  of each of the terms that compose the sum in~\eqref{FitZ} is just $\frac{1}{2} J(t) (y(t) - \hat{y}(t))$ where $J(t) = \partial \hat{y}(t)/ \partial \theta$ is the Jacobian matrix containing the derivatives of $\hat{y}(t)$ with respect to the parameter vector $\theta$. Let us denote $D(t) =  \partial h(t)/ \partial \theta$, implying the following \textit{sensitivity equations}:
	\begin{subequations}
		\begin{align}
		D(t+1) &= \frac{\partial h(t+1)}{\partial h(t)} D(t) + \frac{\partial h(t+1)}{\partial x(t)}, \\
		J(t) &= \frac{\partial \hat y(t)}{\partial h(t)} D(t) + \frac{\partial \hat y(t)}{\partial x(t)}. 
		\end{align}
	\end{subequations}
	In this case, it follows from system analysis that the stability of the system of equations depends on the matrix $A(t) = \frac{\partial h(t+1)}{\partial h(t)}$. If: 
	1) all its eigenvalues are smaller than zero for every $t$, the matrix system is exponentially stable and $J(t)$ will converge to zero as $t\rightarrow \infty$; 2) on the other hand, if one of the eigenvalues of $A(t)$ is always larger then $1$, than the system can become unstable. Both cases are undesirable from the training perspective: vanishing gradients yield slow convergence and make it hard to learn long term dependencies. Exploding gradients make the optimization problem hard to optimize.
	
	The problem of exploding and vanishing gradients was detected early on~\cite{bengio_learning_1994,hochreiter_long_1997}. More modern analysis include~\cite{pascanu_difficulty_2013} and~\cite{ribeiro_exploding_2020}. The problem motivated many modifications of the RNN structure, including the use of gating mechanisms (to be discussed next), gradient clipping~\cite{pascanu_difficulty_2013}, non-saturating activation functions~\cite{chandar_nonsaturating_2019}, the manipulation of the propagation path of gradients~\cite{kanuparthi_hdetach_2019} and the use of orthogonal RNNs where the eigenvalues of the hidden-to-hidden weight matrix are fixed to one using manifold optimization techiniques~\cite{helfrich_orthogonal_2018,lezcano-casado_trivializations_2019,lezcano-casado_cheap_2019,mhammedi_efficient_2017,vorontsov_orthogonality_2017}.
	
	Intuitively, the problem of vanishing gradients can be identified by the shape of commonly used activation functions which are flat for large portions of the input space, recall Fig.~\ref{fig:activation}. Once the inputs are in these regions, the gradients will vanish. For exploding gradients the problem lies intuitively in the optimization landscape and the initialization. If we reach a point of sharp decay in the optimization landscape, we may obtain a gradient larger than one. For RNNs we optimize by unrolling the network in time and therefore multiply this gradient with itself multiple times which can lead to large, exploding gradients.

	\paragraph*{Gating mechanism} Arguably  one of the most popular  RNN architectures is the \textit{long short term memory} (LSTM) network~\cite{hochreiter_long_1997}. The architecture tries to solve the problem of vanishing or exploding gradients during training. The idea of the LSTM is to include gates that control when the internal state is modified by the recurrent block or when it enters as an identity connection. Gates consist of sigmoid functions which output a value in $[0,1]$ which the incoming signal is multiplied with and hereby these gates control the flow of information in an LSTM. Another popular architecture that implements a gating strategy is the \textit{gated recurrent unit} (GRU)~\cite{cho_learning_2014,chung_empirical_2014}. Gating units try to solve the problem of vanishing gradients to make it possible for the model to learn long-term dependencies. The motivation for it is obvious when modeling language and text: the next word depends on the nearby words, but also on the entire context that can only be fully understood if taking into account words that might be far away in the sequence. When modeling dynamical systems, this is not necessarily a desirable property: often the system of interest is known to be exponentially stable or it is Markovian and in this case it is natural that long-term dependencies are not necessarily needed. The use of gated units has been evaluated for nonlinear system identification~\cite{LjungDeep2020,rehmer_using_2019}, and also modifications to make it more suitable have been proposed~\cite{gonzalez_nonlinear_2018}.
	
	\paragraph*{Attractors and long-term dependencies} While learning long-term dependencies are not necessarily desired in system identification, there are scenarios where it is important. These include modeling of complex nonlinear systems with multiple fixed points, oscillatory or chaotic behavior. As these scenarios are not the main focus of this article, we will only highlight some development in this area as pointers for the more interested reader without going into more detail. There are attempts to formalize the idea of long-term memory in terms of the ability of the recurrent model to learn attractors~\cite{bengio_learning_1994,ribeiro_exploding_2020}. Sussilo et al.~\cite{sussillo_opening_2013} show how RNNs can a) implement memory by learning fixed-points and b) implement oscillators by learning oscillatory attractors. Analysing dynamic attractors has been used, for instance for sentiment analysis models~\cite{maheswaranathan_reverse_2019}. Moreover,~\cite{ribeiro_exploding_2020} show how RNNs go through bifurcations during the training procedure. RNNs have also been successfully used to model chaotic systems. For instance, the LSTM has been used to learn chaotic attractors~\cite{sangiorgio_robustness_2020,vlachas_datadriven_2018}.

	\paragraph*{Reservoir computing} The Echo State Network~\cite{jaeger_harnessing_2008} is one type of RNN that has shown very good ability to model chaotic systems~\cite{lu_attractor_2018,lu_reservoir_2017,pathak_using_2017}. 
	This network uses a framework called \textit{reservoir computing}: The model maps the input signal into a higher dimensional space through randomly generated dynamics, called a reservoir. On the one hand, the random dynamics give the system flexibility. On the other hand, training can still be done very efficiently since only the output reading equation is trained.
	In its simplest form the Echo State Network is described by a state-propagation equation
	\begin{equation*}
	h(t+1) = \sigma(\bar{W}_{hh} h(t) + \bar{W}_{hh} x(t) + \bar{b}_h),
	\end{equation*}
	and an output reading equation
	\begin{equation}
	\hat{y}(t) = W_{yh} h(t) + W_{yx} x(t) + b_y.
	\end{equation}
	The difference here compared to a RNN from~\eqref{eq:elman-state}, \eqref{eq:elman-output} is that the weights and bias parameters of the state-propagation equation ($\bar{W}_{hh}$, $\bar{W}_{hh}$, $\bar{b}_h$) are selected at random and fixed during training. The weight matrices are usually set to be sparse (with the sparsity pattern also selected at random). Only the output reading equation are adjusted from the data, i.e, $\theta = (W_{yh},  W_{yx}, b_y)$. In this case, this can be done by solving a least-squares problem.  
	% ----------- End AHR text ----------- %
	
	% ----------- Begin DG text ----------- %
	\paragraph*{Probabilistic recurrent neural network} 
	We can extend the idea of RNNs which are inherently deterministic into a probabilistic version by learning the parameters of a parameterized distribution as output of the network \cite{fortunato2017bayesian}. We can interpret this in a Bayesian way as posterior distribution learning, which is practically applied in \cite{chien2015bayesian,mirikitani2009recursive}. For instance, we could assume that the output is described by a normal distribution $\mathcal{N}(\mu(t), \sigma(t)^2)$. In this case, the output reading equation, see e.g.~\eqref{eq:elman-output} provides the mean and the logarithm of the standard deviation, constituting a complete characterization of this distribution:
	\begin{align}
	\label{eq:prob RNN}
	\begin{bmatrix}
	\mu(t)\\
	\log\sigma(t)^2
	\end{bmatrix} &= W_{yh} h(t) + W_{yx} x(t) + b_y.
	\end{align}
	The logarithm is used to ensure that the variance remains non-negative. The model can still be trained by maximum likelihood.

	\subsection{Latent variable models}
	\label{sec:latent-variable}
	
	The term \textit{latent variable} is widely used in machine learning and describes an unobserved variable which summarizes information about observed variables \cite{jordan1999graphical}. RNNs with their deterministic hidden state, e.g. $h(t)$ in~\eqref{eq:prob RNN}, can be viewed as latent variable models. In this section we concentrate on models where the latent variable is explicitly modelled by using autoencoders.
	
	\paragraph*{Autoencoder}
	% motivation from PCA and PCA reconstruction
	An autoencoder \cite{Baldi1989,hinton2006reducing,hinton1993autoencoders} consists of two parts. 1) An encoder $e(\cdot)$ which maps the inputs $x$ into a latent space variable $z$ which is usually of much lower dimension, i.e. $e:x\mapsto z$ with $\dim(z)\ll\dim(x)$; and 2) a decoder $d(\cdot)$ which tries to reconstruct the input from the latent variable by mapping back to the original space, i.e. $d:z\mapsto \hat{x}$ 
	\begin{align}
	\label{eq:autoencoder}
	z &= e(x), \qquad \hat{x}=d(z).
	\end{align}
	In its simplest form a linear, one-layer autoencoder can be constructed by using \textit{principal components analysis} (PCA) as encoder and its inverse operation as decoder. For more general autoencoders we replace the linear functions in the encoder and decoder network with deep nonlinear networks, e.g. fully connected or convoluational layers can be used. Hence, there exists a direct connection to nonlinear PCA \cite{hinton2006reducing}. In contrast to PCA, local linear embedding (LLE) \cite{roweis2000nonlinear} encodes the data with geometric constraints such as translation, rotation and rescaling. While nonlinear autoencoders do not enforce this by default, studies combined LLE into nonlinear autoencoders for example in \cite{lu2020generalized}. The autoencoder model is used for example for anomaly detection \cite{sakurada2014anomaly}, information retrieval \cite{krizhevsky2011using,salakhutdinov2009semantic} and denoising via its dimension reduction \cite{Vincent2008,vincent2010stacked}. The model is learnt by minimizing the reconstruction error between the original input $x$ and the reconstruction $\hat{x}$ from the decoder. 
	
	The vanilla version of an autoencoder is not suited for modelling dynamics. However, it can be extended to a dynamic autoencoder by including past inputs or past inputs and outputs, see~\eqref{eq:nnfir}, \eqref{eq:nnarx} to the encoder input $x(t)$ \cite{MastiDeepSS2018}. The autoencoder then reconstructs a collection of system outputs $\hat{y}(t)$, i.e., $\hat{y}(t) =d(z(t))$ that have been obtained by encoding the input $z(t) = e(x(t))$.  A natural option to model the temporal dependencies is to include a neural network $f:z(t) \mapsto z(t+1)$ as a temporal propagation network between the latent variables. Similar models are proposed in \cite{beintema2021nonlinear,otto2019linearly}.

	\paragraph*{Variational autoencoder} 
	The variational autoencoder model is a probabilistic extension of the autoencoder. As motivation, we will extend the PCA from the previous section into a probabilistic setting to a probabilistic PCA \cite{bishop_pattern_2006}. In the probabilistic PCA we can shape the latent encodings with the choice of a prior distribution. In this framework, when choosing to work with Normal distributions, we have $p_\theta(x|z)=\mathcal{N}(x;\mu_\theta(z),\sigma_\theta^2(z))$ and we can choose a prior for the latent variable, e.g. $p(z)=\mathcal{N}(z;0,I)$. We can optimize this model now either in a closed form solution (since we work with linear functions $\mu(x)$, $\sigma^2(x)$ and Normal distributions) using the maximum likelihood solution or more generally by iteratively optimizing the log-likelihood which is differentiable w.r.t. the parameters in this setting. This model allows for generating new samples~$\hat{x}$ by drawing from the latent distribution $p(z)$ and using the decoder model. This \textit{generative} behavior was not possible in the deterministic autoencoder since the latent space would only cover the points corresponding to the training data but not the complete space. 
	
	The use of probabilistic PCA is limited due to its inherent linearity. The variational autoencoder (VAE) \cite{kingma_autoencoding_2014,rezende2014stochastic} is a natural extension for a nonlinear, probabilistic setting \cite{roweis1997algorithms,tipping1999probabilistic}. The generative model is now depicted by the joint distribution $p_\theta(x,z)=p_\theta(x|z)p_\theta(z)$. Here, compared to the linear setting from above, the functions for the parameters of the decoding distribution $p(x|z)$, i.e. $\mu_\theta(z)$ and $\sigma_\theta^2(z)$, are deep nonlinear networks. The prior $p(z)$ is often chosen to be a standard normal distribution. For the decoder, the posterior $p(z|x)$ is intractable in general and therefore approximated by a parametric distribution $q_\phi(z|x)$. The latter is often chosen to be normal, where the mean and covariance are learnable through a deep network. In this setup we have to optimize the parameters $\theta$ of the encoder and the parameters $\phi$ of the decoder jointly. Therefore, it is not possible to evaluate the log-likelihood with respect to all parameters. We can however make use of variational inference \cite{blei_variational_2017,wainwright2008graphical} to approximate the log-likelihood. This is achieved by maximizing the Evidence Lower BOund (ELBO), which consists of a reconstruction term and a regularization term. In order to backpropagate through the sampled latent variable, the reparameterization trick \cite{kingma_autoencoding_2014} is used. Tutorials for VAEs are given in \cite{doersch2016tutorial,Kingma2019}. 
	
	Due to the full coverage of the latent space by the distribution $p_\theta(z)$, the VAE can be used as a generative model to generate new data points from a sampled latent vector $z$, used for example for images \cite{razavi2019generating,vahdat2020nvae} or speech \cite{hsu2017learning}. Similarly to the autoencoder, we can use a VAE for anomaly detection \cite{an2015variational}. Furthermore, the encoder is shown to learn disentangled representations \cite{bengio2013representation,higgins2016beta} which are useful for transfer learning.
	
	Similar to the extension of autoencoders to dynamic autoencoder, it is natural to extend the VAE to a dynamic VAE. Here, we do not reconstruct the input $x$ but the output of the dynamical system $y(t)$. The input to the VAE is $x(t)$ which can utilize past inputs as well as past inputs and outputs, see~\eqref{eq:nnfir}, \eqref{eq:nnarx}. We impose a causal ordering of the latent vectors with temporal propagation network for the latent state $f:z(t)\mapsto z(t+1)$. The joint distribution can be factorized as
	\begin{align}
	p(y(1:T)&,z(1:T)|x(1:T)) = 
	\\\prod_{t=1}^T& p(y(t)|y(1:t-1),z(1:t),x(1:t)) \notag
	\\ &p(z(t)|y(1:t-1),z(1:t-1),x(1:t)). \notag
	\end{align}
	To simplify the factorization, different models from literature introduce assumptions for conditional independence between $x(t)$, $z(t)$ and $y(t)$ as well as past variables \cite{alias2017z,bayer2014learning,chung2015recurrent,fabius2014variational,fraccaro2017disentangled,fraccaro2016sequential,goel2014learning,leglaive2020recurrent}. Depending on these assumptions, one can use different deep networks to model e.g. the propagation of the latent state $z(t)$ dependent on past latent states $z(t-1)$ as well as the current input $x(t)$ and output $y(t)$. We can use e.g. a RNN to accumulate all the past information or use a temporal convolutional network for a certain range of past information or use a fully-connected network to model a Markov assumption. To learn the parameters, one requires to extend the procedure from the VAE to maximize the ELBO to a temporal setting. The thesis \cite{fraccaro2018deep} provides a useful introduction to dynamic VAEs and \cite{Gedon2021} applies them to system identification where it becomes clear that dynamic VAEs can be seen as deep probabilistic state-space models.

	We highlight two examples of dynamic VAEs; a comprehensive review is given in \cite{girin2020dynamical}. The Kalman VAE \cite{fraccaro2017disentangled} uses a RNN to process the inputs $x(t)$ with a latent state $h(t)$ and updates the prior with the RNN output such that $p_\theta(z(t))=p_\theta(z(t)|h(t))$. The Variational RNN \cite{chung2015recurrent} uses in the RNN additionally the past latent variable to obtain $p_\theta(h(t))=p_\theta(h(t)|h(t-1),z(t-1),x(t))$.
	% ----------- End DG text ----------- %
	
	% ----------- Begin TS text ----------- %
	\subsection{Formulating and solving regression problems using deep learning}\label{sec:regressionDeep}
	% 1. General introductory paragraph to motivate at a high level
	% 
	Let us take a few steps back and think about the more fundamental problem of formulating and solving regression problems using deep learning. While the classification problem has received massive attention, the same cannot be said about the regression problem. When it comes to system identification we often make use of a regression formulation. Via a classical nonlinear autoregressive model with exogenous inputs (NARX) we will point out the problems with the perhaps most common use of deep neural networks in system identification. After this the available solutions are grouped into four classes and then introduce one way in which the flexibility available in the deep neural networks is better exploited when it comes to formulating and solving regression problems. 
	%
	% 2. Become very specific via the NARX example
	% 
	\paragraph*{Concrete motivational example} Using a NARX model, the current output $y(t)$ is assumed to be described by a nonlinear function of the past outputs  and past inputs. The aim is then to find the predictive distribution $p(y(t)\mid x(t))$, where $x(t)$ is the vector in~\eqref{eq:nnarx} i.e. $x(t) = [y(t-1), \dots, y(t-m), u(t), \dots, u(t-m)]$. This is often done by first assuming that the output is given by a nonlinear functional of $x_t$ distorted by additive Gaussian noise,
	\begin{align}
	\label{eq:EBM1}
	y(t) = f_{\theta}\left(x(t)\right) + e(t), \quad e(t) \sim \mathcal{N}(0, \sigma^2).
	\end{align}
	The unknown parameters $\theta$ are then often solved for using maximum likelihood resulting in the following optimization problem:
	\begin{align}
	\hat{\theta} = \argmax{\theta}{\sum_{t=1}^{N}\| y(t) - f_{\theta}\left(x(t)\right)   \|_2^2}.
	\end{align}
	which is exactly the parameter estimation method described in~\eqref{FitZ}.
	
	By formulating the problem in this way we restrict the flexibility of the deep neural network. This restriction is made implicitly, since when we assume that the noise $e(t)$ in~\eqref{eq:EBM1} is Gaussian we also assume that the solution $p(y(t)\mid x(t))$ is Gaussian and the flexible deep neural network is then used ``only'' to model the mean of this Gaussian. A natural question to ask is if the flexibility can be used to allow for a more general solution $p(y(t)\mid x(t))$? The answer is yes and the solution lies in finding ways of expressing flexible probability density functions using deep neural networks. One concrete way forward is offered by the so-called energy-based model (EBM). However, before introducing that idea, let us briefly look at other approaches that have been considered. 
	
	\vspace{2mm}
	% 3. Related work
	% 
	\noindent \textbf{Approaches to deep probabilistic regression} The most commonly used way of solving regression problems using deep learning is probably when the deep neural network is used to directly predict the output, an approach we refer to as \textit{direct regression}. This approach is already described in the motivational example above. A second class of methods is that of 
	\emph{probabilistic regression} \cite{ChuaCML:2018,KendallG:2017,LakshminarayananPB:2017}, where the networks are used to model one or several parameters of the distribution, see for example the probabilistic RNN section above. In this way the flexibility inherent in the deep model is put to better use, but this solution is still rather rigid in that its functional form is fixed by design. In the third group of algorithms---\textit{confidence-based regression}---the neural network is used to predict a scalar ``confidence value'' for each input-output pair. This confidence value is then maximized w.r.t. the output in order to find the prediction for the corresponding input. The  approach has been developed and used mainly within the computer vision community, see e.g. \cite{JiangLMXJ:2018,ZhouZK:2019}. While these methods have shown impressive performance, they are hard to use since they require several challenging and highly task-dependent design choices to be made. The last approach is that of \textit{regression by classification} where the regression problem is turned into a classification problem by discretizing the output space into $M$ classes. Once this has been done standard classification techniques are applicable. A key problem with this approach is that the neighbourhood structure is lost. This structure can be recovered by using ordinal regression approaches \cite{cao2020rank}. Yet, the predictions always lie in the interval of the $M$ classes which define a lower bound on the prediction performance.
	
	\vspace{2mm}
	% 4. One concrete suggestion via EBM
	%
	\noindent \textbf{Energy-based model} These models have been available for a long time, see e.g. \cite{BengioDVJ:2003,HintonOWWT:2006,LeCunCHRH:2006,TehWOH:2003}. Traditionally the EBM has been used for unsupervised tasks, but more recently also for supervised tasks, like regression, see e.g. \cite{GustafssonDBS:2020}. For nonlinear system identification a conditional EBM can be employed to model the conditional output density according to 
	\begin{align}
	p_{\theta}\left(y(t)\mid x(t)\right) &= \frac{e^{g_{\theta}\left(y(t), x(t)\right)}}{Z_{\theta}\left(x(t)\right)},\notag\\ 
	\text{with}\qquad Z_{\theta}\left(x(t)\right) &= \int e^{g_{\theta}\left(z, x(t)\right)}dz,
	\end{align}
	where $g_{\theta}$ denotes a neural network mapping a pair $(y_t, x_t)$ to a scalar value. By defining $g_{\theta}$ via a neural network the model becomes expressive enough to model a very wide range of distributions, including asymmetric, heavy-tailed and multimodal. Therefore generalizing the implicit assumption of Gaussian distribution which we made earlier. The cost of this flexibility is that the normalization constant $Z_{\theta}(x_t)$ is intractable. This can be handled using approximate inference methods. See \cite{HendriksGRWS:2021} for a first take on how EBMs can be used for nonlinear system identification.
	
	% ----------- End TS text ----------- %

	\section{Optimization and regularization}
	\label{SectionDeepOpt}
	In this section, we frame deep learning as an optimization problem, highlight specific challenges inherent in current practice and highlight relevant techniques. 
	
	\subsection{Optimization Formulation for Deep Learning}
	
	Generalizing the formulations we have presented in the text, the basic formulation for deep learning can be written as 
	
	\begin{equation}
	\label{eq:general}
	\min_{\theta} \mathcal{F}(\theta) := \sum_{t\in \mathcal{T}} \mathcal{L}(y(t), f(x(t),\theta)),
	\end{equation}
	where $\mathcal{T}$ is the set of times where measurements are collected, $\{x(t), y(t)\}$ is the input/output tuple corresponding to time $t$, and $\mathcal{L}$ is a smooth loss function.  
	
	For example, in the case of the autoencoder, the loss is the least squares norm, while the input and output are the same data. We can make the nonlinear map $f$ and its dependence on the parameters $\theta$ more explicit by specifying $L$ layers, with corresponding weights and biases $\theta_1, \dots, \theta_L$, for example
	\begin{equation}
	\label{eq:auto}
	\min_{\theta_1, \dots, \theta_L} \sum_{t\in \mathcal{T}} \mathcal{L}\Big( 
	y(t), f_L(\theta_L, f_{L-1}(\dots (f_1(x_t, \theta_1 ))\Big).
	\end{equation}
	
	Explicit regularization may be considered for any architecture, and we may write the final objective of interest as
	\begin{equation}
	\label{eq:overallOpt}
	\min_\theta \mathcal{F}(\theta) + \mathcal{R}(\theta).
	\end{equation}
	Before we discuss methods for solving~\eqref{eq:overallOpt}, we consider the overall goal, as well as theoretical and computational properties of $\mathcal{F}$ and $\mathcal{R}$,
	where $\mathcal{R}$ may be smooth, such as the ridge or kernel-based regularizers described in Section \ref{SectionDeepKernel}, or nonsmooth, as in many emerging approaches exploiting sparsity-inducing priors \cite{ZhouAuto2022}. Some approaches may also regularize implicitly, e.g. using early stopping or dropout; these techniques are discussed later on in the section.
	
	Before we discuss methods for solving~\eqref{eq:overallOpt}, we consider the overall goal, as well as theoretical and computational properties of $\mathcal{F}$ and $\mathcal{R}$.
	
	\paragraph*{Overaching goal.} The main goal of optimization for deep learning is to reduce generalization error, that is, to create a model that is capable of providing high-quality predictions. Solving the optimization problem~\eqref{eq:overallOpt} is a means to that end. Approaches are often evaluated based on (1)~out of sample performance with regard to a specific learning task and (2)~computational efficiency of the overall approach. 
	
	\paragraph*{Non-convex objective.} The deep learning optimization problem~\eqref{eq:general} or \eqref{eq:overallOpt} is non-convex for any choice of $\mathcal{L}, f$, and $\mathcal{R}$. This is a consequence of the nested structure of nonliearities in the model.  As a result, it is only possible to find stationary points in the general sense, as opposed to local or global minimizers.  
	
	\paragraph*{Global quadratic bounds.} Algorithm design and analysis for some algorithms below rely on presence of global upper bounds. The most common  nonlinearities (ReLU, softmax, sigmoid, and hyperbolic tangent) are globally bounded or grow linearly with their input, and thus the composition $\mathcal{F}$ is globally bounded by a quadratic whenever this is true for the chosen loss $\mathcal{L}$. 
	
	\paragraph*{Smoothness of $\mathcal{F}$.}  Smoothness properties of $\mathcal{F}$ are inherited from those of the nonlinear map $f$. 
	For example,  the sigmoid, hyperbolic tangent, and softmax nonlinearities all have derivatives of all orders. On the other hand, the ReLU function is merely continuous, and not smooth; thus $\mathcal{F}$ is also merely continuous when ReLUs are used in any layer. Existing methods do not differentiate between continuous and smoothly differentiable nonlinearities, using them interchangeably with no apparent loss in performance. 
	
	\paragraph*{Calculus for $\mathcal{F}$,} 
	The function $\mathcal{F}$ has a special structure induced by nested nonlinearities, as seen e.g. in~\eqref{eq:auto}. 
	The chain rule for $\mathcal{F}$ is equivalent to backpropagation. Current software packages, including TensorFlow~\cite{abadi2016tensorflow}, PyTorch~\cite{paszke2017automatic} and JaX~\cite{jax2018github} implement automatic differentiation across multiple architectures and nonlinearities. This means that first and second derivatives of $\mathcal{F}$ can be computed, though second derivatives are rarely used because of scalability concerns. For example, classic Newton's method requires forming and inverting a large matrix, typically requiring \(O(Nm^2 + m^3)\) operations, with $N$ training points and $m$ parameters, a prohibitive cost. Instead of fully relying on these methods, some ideas that incorporate curvature approximations, such as diagonal scalings, are used along with first-order methods.

	\paragraph*{Nonsmooth $\mathcal{R}$.} In contrast to 
	$\mathcal{F}$, nonsmoooth $\mathcal{R}$ are used to induce specific structure in the weights (such as sparsity). In this case treating $\mathcal{R}$ as differentiable leads to substantial loss in optimization performance, and special techniques that exploit the structure of $\mathcal{R}$ should be used to develop far more efficient algorithms.

	\paragraph*{Scalability.} The dataset is a major issue for deep learning approaches. The size of $\theta$  may be quite large, particularly as we increase layers or otherwise complicate the structure. In order to identify the model, we need an even larger dataset, which can have millions or billions of training points. As a result, methods that can make the most progress relative to the number of passes through the data are of great interest. 
	
	\paragraph*{Conditioning.} The issue of vanishing gradients in neural networks points to ill-conditioning, due to compositions of multiple non-linearities. Ill-conditioning refers to the presence of multiple scales in the problem -- while some gradients may be large, others can vanish. Several papers have pointed to the structural changes to deep networks that improve performance and conditioning, most notably 
	batch normalization~\cite{ioffe2015batch}. 
	Batch normalization adds nodes within layers that de-mean and normalize inputs to the nonlinearities, thus normalizing activations. Since these techniques modify the neural network architecture, they can be used with any optimization algorithm described below, and can be thought of as a preconditioner~\cite{santurkar2018does} adapted to the architectures, smoothing out the optimization landscape and stabilizing the gradients used in the optimization process.

	\subsection{Overview} 
	
	While classic optimization methods seek to effectively decrease the necessary total number of update steps or iterations (for example, by developing methods that converge quickly), the key metric for deep learning is the number of {\it epochs}, i.e. times the method passes through the data, over the course of the training.  This lead to an emphasis on {\it stochastic} methods, which make progress using subsampled data volumes, so each iteration need only see a small fraction of the total data. The focus on generalization error has also led to multiple types of regularization, including both explicit (i.e. using $\mathcal{R}$) and implicit methods, such as dropout.

	The rest of the section proceeds as follows. In Section~\ref{sec:opt_basics} we develop a simple view of gradient, prox-gradient, and stochastic gradient methods for~\eqref{eq:overallOpt}. We also identify key challenges, including making fewer passes through the data, allowing for non-smooth regularizers, and present optimization techniques that help to address them. 
	We discuss variance reduction in Section~\ref{sec:variance_reduction}, adaptive scaling in Section~\ref{sec:adaptive_scaling}, and dropout in Section~\ref{sec:additional_techniques}.
	The analysis presented here is complemented
	in Section~\ref{sec:implicit-regularization} where we show that, in the absence of $\mathcal{R}$,   
	gradient and stochastic gradient approaches may implicitly regularize the problem, finding minimum norm solutions which may generalize well on new data.

	\subsection{Steepest Gradient, Prox Gradient, and Stochastic Gradient}
	\label{sec:opt_basics}
	
	Starting at a point $\theta^0$, the aim is to move to a new point $\theta^1$ that will decrease objective~\eqref{eq:overallOpt}. 
	A foundational idea of optimization is to develop and minimize a simple global upper bound, 
	\[
	\mathcal{F}(\theta) \leq \mathcal{M}(\theta) 
	\]
	that satisfies $\mathcal{M}(\theta_0) = \mathcal{F}(\theta^0)$. If \(\mathcal{M}\) is easy to optimize, by taking 
	\[
	\theta^1 = \argmin_\theta(\mathcal{M}(\theta))
	\]
	we obtain
	\[
	\mathcal{F}(\theta^1) \leq \mathcal{M}(\theta^1) \leq \mathcal{M}(\theta^0) = \mathcal{F}(\theta^0),
	\]
	ensuring descent. This is a key idea for both steepest descent and prox-gradient descent.

	\paragraph*{Steepest descent.}
	In the absence of $\mathcal{R}$, 
	suppose we can find \(\alpha\) such that 
	\begin{equation}
	\label{eq:upper}
	\mathcal{M}_0(\theta) = \mathcal{F}(\theta^0) + \nabla\mathcal{F}(\theta^0)^T(\theta-\theta^0) + \frac{\alpha}{2}\|\theta-\theta^0\|^2
	\end{equation}
	is a global upper bound for $\mathcal{F}$. 
	
	Here minimizing $m_0$ to find \(\theta^1\) can be done with simple calculus, and we get 
	\[
	\theta^1 = \theta^0 - \frac{1}{\alpha}\nabla \mathcal{F}(\theta^0),
	\]
	yielding the steepest descent method. Furthermore, completing the square in~\eqref{eq:upper}, we get an explicit estimate of minimal descent:  
	\[
	\mathcal{F}(\theta^1) \leq \mathcal{M}_0(\theta^1) = \mathcal{F}(\theta^0) - \frac{1}{2\alpha}\|\nabla \mathcal{F}(\theta^0)\|^2 
	\]
	where the inequality is immediate from the fact that $\mathcal{M}_0$ is a global upper bound for $\mathcal{F}$. Thus steepest descent guarantees descent proportional to the step $\frac{1}{\alpha}\|\nabla\mathcal{F}\|^2 = \alpha\|\theta^0 - \theta^1\|^2$. Making no further assumptions on $\mathcal{F}$, we can guarantee that the minimum norm of the observed gradients goes to $0$ no slower than $\nu^{-1/2}$, with $\nu$ the number of iterations. Stronger assumptions (such as convexity or strong convexity) lead to faster rates of convergence and stronger statements about solutions. 
	
	There are two problems with steepest descent. First, computing $\nabla \mathcal{F}$ requires seeing all of the data, since $\mathcal{F}$ is defined as a sum across the training set. 
	Second, while $\alpha$ does exist, an estimate is difficult to obtain and any global estimate would be conservative, resulting in unnecessarily small step sizes for the method. We will return to these challenges presently. 
	
	\paragraph*{Prox-gradient descent.}
	When we have a simple term $\mathcal{R}$, e.g. a non-smooth separable regularizer, we can use the bounding idea in the previous section to obtain a simple technique that still guarantees descent and converges as fast as steepest descent. We can simply include  $\mathcal{R}$  into the upper bounding model yielding
	\begin{align}
	\begin{split}
	\label{eq:upperR}
	\mathcal{M}_0(\theta) = \mathcal{F}(\theta^0) &+ \nabla\mathcal{F}(\theta^0)^T(\theta-\theta^0) \\
	&+ \frac{\alpha}{2}\|\theta-\theta^0\|^2 + \mathcal{R}(\theta).
	\end{split}
	\end{align}
	
	Completing the square, we arrive at the following:
	\[
	\theta^1 = \arg\min_\theta \frac{\alpha}{2}\|\theta-(\theta^0-\frac{1}{\alpha} \nabla \mathcal{F}(\theta^0))\|^2 + \mathcal{R}(\theta). 
	\]
	This operation motivates the definition of proximal operator for $\mathcal{R}$: 
	that is,
	\begin{equation}
	\label{eq:prox_def}
	\mbox{prox}_{s \mathcal{R}}(z) := \arg\min_\theta \frac{1}{2s} \|\theta-z\|^2 + \mathcal{R}(\theta)
	\end{equation}
	and using this operator, the iteration simplifies to 
	\[
	\theta^1 = \mbox{prox}_{(1/\alpha)\mathcal{R}} \left(\theta^0 - \frac{1}{\alpha} \nabla \mathcal{F}(\theta^0)\right).
	\]
	where the step $s$ in~\eqref{eq:prox_def} is $\frac{1}{\alpha}$, and the $z$ in~\eqref{eq:prox_def} is given by \(\theta^0 - \frac{1}{\alpha}\mathcal{F}(\theta^0)\).
	
	Just as in the gradient descent case, we are guaranteed 
	descent that is proportional to $\alpha\|\theta^1-\theta^0\|$, 
	now defined with respect to the prox-gradient algorithm. 
	
	The gradient and prox-gradient methods are analyzed at length in multiple sources, see e.g.~\cite{beck2017first}. 
	The main point here is that, when the prox operator for $\mathcal{R}$ is available, the prox-gradient method behaves very similarly to gradient descent. The downsides are also 
	analogous: both require a full pass through the data, and both require a step-size strategy. 
	
	\paragraph*{Stochastic gradient descent.} Considering the smooth case, with $\mathcal{R} = 0$, we again write the objective as an aggregate over the training data:
	\[
	\mathcal{F}(\theta) = \sum_{t\in \mathcal{T}} \mathcal{L}(y(t), f(x(t), \theta)). 
	\]
	The full gradient is given by 
	\[
	\nabla \mathcal{F}(\theta) = \sum_{t\in \mathcal{T}} \nabla \mathcal{L}(y(t), f(x(t), \theta))
	\]
	and requires a full pass through the data. Stochastic gradient instead requires only a selection of the elements of $\mathcal{T}$, called a `batch', to make progress:
	\[
	\theta^{\nu+1} = \theta^\nu - s_\nu \sum_{t \in \mathcal{S}_\nu} \nabla\mathcal{L}(y(t), f(x(t), \theta^\nu)).
	\]
	The selection of the batch $\mathcal{S}_\nu$ is typically  random, and may be as small as a single element. When using a fixed step size, the method does not stay at stationary points -- for example it will move away even when initialized at the global minimum, because every stationarity condition is the result of consensus across all the data, with individual datapoints (and their gradients) pulling in different directions. Thus step sizes $s_\nu$ must decrease to have any hope of convergence, but not too quickly, so the method does not stagnate. The classic condition here is given by 
	\[
	\sum_\nu s_\nu = \infty, \qquad \sum_\nu s_\nu^2 < \nu. 
	\]
	which ensures that the sequence of iterates can reach any point in the space, and allows for a convergence analysis, see e.g.~\cite{Bottou2018}.
	
	\paragraph*{Challenges and extensions.} Stochastic gradient descent removes the need to process the entire dataset $\mathcal{T}$ in order to make progress. 
	However, a number of challenges remain:
	\begin{itemize}
		\item Requiring step sizes that decrease to $0$ can slow down the algorithm - and tuning the step size is typically a difficult problem in its own right. 
		
		\item There is no convergence theory for stochastic prox-gradient descent:
		\[
		\theta^{\nu+1} = \mbox{prox}_{s_\nu \mathcal{R}}\left( \theta^\nu - s_\nu \sum_{t \in \mathcal{S}_\nu} \nabla\mathcal{L}(y(t), f(x(t), \theta^\nu))\right) 
		\]
	\end{itemize}
	
	Many researchers have proposed creative heuristic techniques to handle the first challenge. These include restarts~\cite{loshchilov2016sgdr}, cyclical schedules~\cite{smith2017cyclical}, and batch size schedules~\cite{smith2017don}. Each of of these approaches is fighting against the fundamental limitation of requiring shrinking step sizes -- both restarts and cyclic schedules allow the sizes to increase again after decreasing, while larger batches decrease the stochastic noise in the resulting gradient. Each may help to improve  performance given a fixed computational budget, and all point to the underlying difficulties of relying on SGD as a workhorse in training deep neural networks.
	
	In the remaining sections, we review a number of recent extensions that address these challenges. In particular, we discuss variance reduction ideas, 
	that build on stochastic gradient methods. The resulting algorithms are provably more efficient in their use of information, and they rigorously incorporate 
	proximal operators. We also discuss parameter-specific scaling, which can help to accelerate methods, and highlight recent work that extends these approaches to prox-gradient methods. 
	
	\subsection{Variance Reduction}
	\label{sec:variance_reduction}
	
	Stochastic gradient uses a subset of available training data to compute gradients. In principle it is able to make progress (that is, update weights) after seeing only a single data point, out of billions. However, the fundamental weakness is the noise induced by this process -- each gradient may pull the weights in its own direction, so consensus requires a decreasing step size schedule and may take a while to converge. 
	
	We can express this idea more precisely as follows:
	\[
	\mathbb{E}\left\| \nabla \mathcal{F}(w) - \sum_{t \in \mathcal{S}_\nu} \nabla\mathcal{L}(y(t), f(x(t), \theta^\nu)) \right\|^2 \geq C > 0,
	\]
	for some constant $C$. In other words, for any fixed batch size, even if the stochastic gradient estimate is unbiased, the variance has a floor, and we need to decrease the steps to have any hope for convergence. 
	
	Variance reduction techniques use stochastic information in a smarter way, allowing for faster progress in practice and faster theoretical convergence rates. A key idea is to use a variable that is more correlated with the true gradient, by leveraging historical information. 
	
	Suppose that at iteration $k$, our gradient estimate is called $g^\nu$, 
	and we are about to sample a fresh batch $\mathcal{S}_{\nu+1}$ of gradients. The key idea is that instead of completely forgetting $g^{\nu}$ and estimating the gradient using the new batch, that is, 
	\[
	g^{\nu+1} = \sum_{t \in \mathcal{S}_{\nu+1}} \nabla\mathcal{L}(y(t), f(x(t), \theta^{\nu+1}))
	\]
	we actually {\bf replace} the old gradients in that batch with the fresh evaluations: 
	\begin{equation}
	\label{eq:gradUpdate}
	g^{\nu+1} = g^\nu - \sum_{t \in \mathcal{S}_{\nu+1}} g_t^{\nu} + 
	\sum_{t \in \mathcal{S}_{\nu+1}} \nabla\mathcal{L}(y(t), f(x(t), \theta^{\nu+1}))
	\end{equation}
	This requires either storing extra information, or re-evaluating $g_t^\nu$ on the fly~\cite{defazio2014saga,johnson2013accelerating}. It is nontrivial to show that the replacement strategy gives 
	\[
	\mathbb{E}_\nu \left\| \nabla \mathcal{F}(\theta) - g^\nu \right\|^2  \rightarrow 0
	\]
	as $\nu\rightarrow \infty$, and in particular reducing the step size is not necessary for these algorithms. However, to achieve best performance and convergence rates, full gradient evaluations must be performed a small proportion of the time~\cite{johnson2013accelerating}, which makes the method less practically appealing compared to SGD, which can start training right away with any batch size. However, there are advantages in using variance reduction, both in theoretical performance and generalization to harder problems. 
	
	Variance reduction methods generalize fully for the class~\eqref{eq:overallOpt}, where $\mathcal{F}$ is continuously differentiable while $\mathcal{R}$ is lower semi-continuous and has a computable prox. The algorithm alternates betweeen batch updates of the gradient, as 
	shown in~\eqref{eq:gradUpdate}, and prox evaluations according to
	\begin{equation}
	\label{eq:SVRG}
	\begin{aligned}
	g^{\nu+1} & = \sum_{t \in \mathcal{S}_{\nu+1}} 
	(\nabla \mathcal{L}(y(t), f(x(t), \theta^{\nu}) - g_t^\nu) + g^\nu, \\
	\theta^{\nu+1} & = \mbox{prox}_{\alpha \mathcal{R}}(\theta^\nu - \alpha g^{\nu+1}). \\
	%g^{\nu+1} & = g^{\nu} + \sum_{d \in \mathcal{S}_{\nu+1}}  \nabla \mathcal{L}(d) - g_d^\nu \\
	\end{aligned}
	\end{equation}
	Just as in the classic variance reduction for smooth problems, the entire gradient must be fully evaluated a small portion of the time. 
	Full details, analysis, and convergence rates are given in~\cite{aravkin2020trimmed}. 
	The step size $\alpha$ required for convergence is problem dependent, but 
	need not decrease as the iterations proceed. 
	
	\subsection{Adaptive scaling and momentum}
	\label{sec:adaptive_scaling}
	
	The discussion in the previous sections assumed a uniform step size for each coordinate in the parameter vector $w$. Unfortunately, if the weights in the network operate across different scales,  the requirement of a single step size for all weights guarantees that gradient methods will squash information coming through features on smaller scales in favor of larger scaled features. 
	
	Second-order methods, such as Newton, adapt to curvature across scales by using a quadratic model (instead of the linear model~\eqref{eq:upper}):  
	\begin{equation}
	\label{eq:quadratic}
	\begin{aligned}
	\mathcal{M}_0^Q(\theta) &= \mathcal{F}(\theta^0) + \nabla\mathcal{F}(\theta^0)^T(\theta-\theta^0) \\
	&+ \frac{1}{2}(\theta-\theta^0)^T\nabla^2 \mathcal{F}(\theta^0) (\theta-\theta^0) 
	\end{aligned}
	\end{equation}
	We can use calculus to obtain the global minimizer of~\eqref{eq:quadratic}:
	\begin{equation}
	\label{eq:Newton}
	\theta^1 = \argmin_\theta \; \mathcal{M}_0^Q(\theta) = -(\nabla^2\mathcal{F}(\theta^0))^{-1} \nabla \mathcal{F}(\theta^0)
	\end{equation}
	
	Replacing the Hessian in~\eqref{eq:Newton} with a diagonal approximation yields parameter-specific step-sizes. In the presence of the regularizer~$\mathcal{R}$, a diagonal Hessian approximation yields a slightly more complicated proximal subproblem. For example, 
	the AdaGRAD algorithm~\cite{duchi2011adaptive} uses
	\begin{equation}
	\label{eq:AdaGRAD}
	\theta^{\nu+1} = \argmin_\theta \alpha( (g^\nu)^T\theta +  \mathcal{R}) + \frac{1}{2}\theta^T D^\nu \theta 
	\end{equation}
	$\alpha$ denotes the step size, $g^\nu$ is the observed (sub)gradient, and $D^\nu[i,i] = \delta + \|g_{1:\nu}, i\|_2$; that is, the diagonal term corresponding to each coordinate grows with with 2-norm of the (sub)gradients
	for that coordinate across the iterations seen so far. For any separable $\mathcal{R}$,~\eqref{eq:AdaGRAD} requires the same computational effort as a prox-gradient update.

	A second powerful idea in speeding up gradient and prox-gradient methods 
	is that of `momentum'.
	The intuitive idea is to take a step in a direction spanned by 
	the gradient we observe and the direction we were moving in at the previous iteration:
	\begin{equation}
	\label{eq:momentum}
	\theta^{\nu+1} = \theta^\nu + a_1(-\nabla \mathcal{F}(\theta^\nu)) + a_2 (\theta^\nu - \theta^{\nu-1}).
	\end{equation}
	for a linear combination denoted by $(a_1, a_2)$, chosen either empirically or from theoretical principles. For convex models, Nesterov~\cite{nesterov2003introductory} showed particular $(a_1, a_2)$ values lead to optimal convergence rates  
	for smooth convex functions; this result was extended by Beck and Teboulle~\cite{beck2009fast} to sums of smooth and nonsmooth convex functions. 
	For non-convex functions, no such acceleration is guaranteed, and $(a_1, a_2)$ are typically chosen by the analyst.
	
	The ADAM method~\cite{kingma2014adam} adapts accelerated gradient with momentum, and combines it with adaptive scaling to speed up stochastic gradient descent. The ADAM iteration is given by 
	\[
	\theta^{\nu+1} = \theta^\nu - a \hat D^\nu s^\nu,  
	\]
	with $s^\nu$ as in~\eqref{eq:momentum},
	and $a_1, a_2$ chosen using the method of moments to estimate the mean, while $\hat D^\nu$ is a diagonal matrix consisting of individual variance estimates of each parameter.
	
	While practically powerful, ADAM is a more complex method than SGD, and the fact that neural nets are non-convex makes it impossible to choose settings that guarantee optimal performance. As a result, ADAM can struggle with standard step size schemes, requiring special `warm-up' regimes that start with small step sizes, then increase them and decrease them again~\cite{liu2019variance}.  
	
	The ideas underlying AdaGRAD and ADAM have recently been generalized by~\cite{yun2021adaptive} 
	and these ideas can be applied also in the general case of smooth non-convex loss $\mathcal{F}$, 
	and non-convex regularizers $\mathcal{R}$. The general iteration is given by 
	\begin{equation}
	\label{eq:proxGEN}
	\begin{aligned}
	g^\nu &= \sum_{t \in \mathcal{S}_{\nu}} \nabla\mathcal{L}(y(t), f(x(t), \theta^{\nu})),\\
	s^\nu &= \rho_\nu s^{\nu-1} + (1-\rho_\nu) g^\nu,\\ 
	\theta^{\nu+1} & = \argmin_\theta \; (s^\nu)^T\theta + \lambda \mathcal{R}(\theta) + \\
	&\qquad \frac{1}{2\alpha_\nu} (\theta-\theta^{\nu})^TD^\nu (\theta-\theta^\nu).
	\end{aligned}
	\end{equation}
	
	It is useful to compare the generalized iteration~\eqref{eq:proxGEN}  
	to~\eqref{eq:SVRG}. The former generalizes stochastic gradient to enable using momentum acceleration, scaling, and preconditioning; however, the effect of the SGD base is still apparent; either $\alpha_\nu$ must go to $0$ or the batch size must increase if we have to control the sampling error. The latter uses a clever gradient approximation that controls sampling error, and can use a fixed batch size and step size.

	\subsection{Dropout}
	\label{sec:additional_techniques}

	The previous sections emphasized explicit regularizers $\mathcal{R}(\theta)$ and 
	algorithmic modifications required to accommodate. The point of these developments
	is to impose particular structure on the output weights, with sparsity a typical 
	flag-ship example. 
	
	Implicit regularization through dropout~\cite{srivastava2014dropout} addresses the key problem of overfitting and generalization, but does not promote a particular type of structure. We close this section by briefly reviewing dropout. 
	
	In dropout, any entries in $\theta$ can be randomly removed during any given iteration of any given training process. In the classic scheme~\cite{srivastava2014dropout}, each weight $\theta_i$ is retained with probability $p_i$, and then at prediction time, the value used to predict is $p_i \theta_i$, the expected weight. 
	
	Rather than viewing Dropout as an optimization technique, it may be more helpful to think of it as an in-place ensembler, just as suggested by~\cite{srivastava2014dropout}. That is, rather than training multiple neural networks and then ensembling their results, dropout interchanges the order of operations, continuously training different nets randomly obtained to achieve the result. This also means that it is possible to try dropout in combination with any structure, any other regularizer $\mathcal{R}$, and any training algorithm.  The interpretation of dropout means that analysis of such combined schemes is quite challenging. Nonetheless, dropout (when used with other strategies, particularly max-norm regularization) can decrease test error across numerous applications~\cite{srivastava2014dropout}, 
	and can be used to estimate model uncertainty~\cite{gal2016dropout}.

	\section{Deep kernel-based learning}\label{SectionDeepKernel}
	%\ref{sec:RKHS},\ref{GaussReg},
	%\ref{IntLatSpace}, \ref{Vector-valued} 
	%\ref{DPredN},\ref{CompKernel}
	In \cite{neal_bayesian_1995} Neal considered
	a neural network with weights modeled as  i.i.d. random parameters. Using the central limit theorem, he showed that the input-output map becomes a Gaussian process as the network's width approaches infinity. 
	Thus, the function estimate becomes available in closed-form and simplifies to manipulation of covariances. 
	Covariance functions are also called kernels in the literature. In fact, there is a fundamental connection between Bayesian estimation of Gaussian processes and regularization in the so called
	reproducing kernel Hilbert spaces (RKHSs) introduced in Section~\ref{sec:RKHS}.  
	We review this link in Section~\ref{GaussReg} and discuss models representing dynamic systems. The focus is on regression problems which, as already recalled in Section \ref{sec:regressionDeep}, play a prominent role in system identification.
	Regularized ARX or NARX models arise when kernels process past input-output data. % and have been widely adopted in control applications. 
	Next, we illustrate how covariances can be introduced in deep architectures to enhance their expressive power. One approach described in Section \ref{IntLatSpace} is to use kernels only to define the last layer of a deep structure while the first layers form an intermediate latent space (defined through any of the deep networks previously described). Another approach uses kernels to define each layer of the structure, see Sections \ref{Vector-valued} 
	and \ref{DPredN}.
	A third way, illustrated in Section \ref{CompKernel}, generalizes \cite{neal_bayesian_1995}  by induction and builds new (compositional) kernels as limits of deep neural networks with increasing layer width.

	%Unsupervised: funzione da R^2 a R, ma x_1=x_2^2. Se scopriamo questa relazione usando solo input locations possiamo ridurre il problema a stimare funzione da R in R, abbiamo ridotto il curse of dimensionality descritto in section precedente.
	
	%The study of the properties of neural networks in the infinite width limit is not new. Indeed, as already recalled at the beginning of Section \ref{SectionDeepKernel}, Neal~\cite{neal_bayesian_1995} showed that a single-layer network with untrained weights converges to a Gaussian process (GP) in the limit of infinite width. The notion generalizes to deep neural networks~\cite{lee_deep_2018}, to convolutional neural networks~\cite{novak_bayesian_2018} and to recurrent neural networks~\cite{yang_wide_2019}. 
	
	%\ahr{I would consider adding a connecting paragraph here giving more context and motivation: my first thoughts would be to either mentioning infinitely wide neural networks (Section~\ref{sec:infinitely-width-nn}); or the connections between deep neural networks and deep kernel learning. } 
	%\dg{I agree with Antonio, maybe also a quick overview of what concepts will be introduced and why? Something like: How are kernels used to model dynamics? Where are deep models used in this concept? What different forms of deep kernels exist?}
	%\dg{generally can we connect the use of neural networks in this section to the last section? That is, what kind of networks are used here?}
	
	\subsection{Reproducing kernel Hilbert spaces}\label{sec:RKHS}

	Positive definite kernels $K$, just called kernels in what follows,
	are important tools to define both the space of dynamic systems illustrated in the right part of Fig. \ref{Fig1}
	and the ranking of possible solutions \cite{PillonettoDCNL:14,Scholkopf01b}.   
	Given any non-empty set $\mathcal{X}$, kernels 
	are symmetric functions 
	over $\mathcal{X} \times \mathcal{X}$ such that
	\begin{equation}
	\label{eq:kernel_def}
	\sum_{i=1}^{p}\sum_{j=1}^{p}a_ia_j K(x_i,x_j) \geq 0
	\end{equation}
	for any finite natural number $p$, choice of scalars $a_i$ and $x_i \in \mathcal{X}$. % \in \mathrm{R}$.  
	The Moore-Aronszajn theorem ensures a one-to-one correspondence between 
	$K$ and a reproducing kernel Hilbert space $\mathcal{H}$ \cite{Aronszajn50,Bergman50}. The function space contains % $f \in \mathcal{H}$ 
	any finite sum of kernel sections, i.e. 
	$$
	f(x) = \sum_{i=1}^{p} a_i K_{x_i}(x), \quad K_{x_i}(x):=K(x_i,x),
	$$
	with norm
	$$
	\|f\|_{\mathcal{H}}^2 = \sum_{i=1}^p  \sum_{j=1}^p a_i a_j K(x_i,x_j)
	$$
	and all the infinite combinations given by limits of Cauchy sequences.
	%w.r.t. the norm $ \|f\|_{\mathcal{H}}^2 = \sum_{i=1}^p  \sum_{j=1}^p a_i a_j K(x_i,x_j)$ \ahr{The definition of norm here is not clear to me. The values of $a_i$ and $x_i$ here would be the ones defining $f$? Maybe be more explicit. Because these terms also appear without any conection to $f$ in other places}.
	These facts suggest that functions in $\mathcal{H}$ inherit the properties of
	$K$: by increasing kernel smoothness, the space $\mathcal{H}$ will contain
	more and more regular functions. %For modeling purposes, %This also suggests important features for modeling purposes. 
	Hence, instead of introducing a set of basis functions, we can specify
	a kernel that encodes the expected system properties \cite{Saitoh88}.\\
	%There are also connections between kernels and neural networks with one layer of hidden units,
	%as discussed in \cite{BengioBook2007,NealBook1996,WilliamsNC1998} and in Section~\ref{sec:infinitely-width-nn} of this Survey.\\
	Some example kernels are now introduced. 
	%Assume that the column vectors $x_i \in X=\mathbb{R}^{m}$ contains past input data, as defined in \eqref{eq:nnfir}. 
	%i.e.
	%\begin{equation}\label{InpLocNARX}
	%	x_{i}=[y_{i-1},u_{i-1},\ldots,y_{i-m},u_{i-m}],
	%\end{equation}
	%\begin{equation}\label{InpLocNFIR}
	%	x_{i}=[u_{i-1},\ldots,u_{i-m}]^\top,
	%\end{equation}
	%where $m$ is the system memory. 

	A linear kernel is defined by a positive semidefinite  matrix $P\in \mathbb{R}^{m \times m}$ as follows 
	\begin{equation}\label{LinKernel}
	K(x_i,x_j) = x_i^\top P x_j.
	\end{equation}
	It induces a space $\mathcal{H}$ containing only linear functions: any function $f$ has the representation
	\begin{equation}\label{LinH}
	f(x)=\theta^{\top} x
	\end{equation}
	with norm (assuming $P$ is invertible for simplicity)
	\begin{equation}\label{NormLinH}
	\|f\|^2_{\mathcal{H}} = \theta^{\top} P^{-1} \theta.
	\end{equation}
	Using a time series as input such as %\eqref{eq:nnfir}, i.e.
	\begin{equation}\label{eq:FIRx}
	x(t)=[u(t),\ldots,u(t-m)]^\top
	\end{equation}
	%\begin{equation}
	% \label{eq:nnarx}
	%    x(t)=[ y(t-1),\ldots,y(t-m-1), %u(t),\ldots,u(t-m)]^\top,
	%\end{equation}
	the related space $\mathcal{H}$ will thus contain FIR  models whose coefficients are the components of $\theta$.
	% in view of \eqref{InpLocNFIR}. 
	%A possible choice is the stable spline/TC kernel: 
	Setting the $(i,j)$ entry of $P$ to
	\begin{equation}\label{SSKernel}
	P_{ij} = \alpha^{\max(i,j)}, \quad 0 \leq \alpha < 1,
	\end{equation}
	we obtain the stable spline or TC kernel \cite{COL12a,PillACC2010,Pillonetto:10a}.
	In view of \eqref{LinH} and \eqref{NormLinH}, 
	if the components of $\theta$
	do not decay to zero sufficiently fast, the RKHS norm induced by the stable spline model will be large. In fact, the diagonal elements of $P^{-1}$ increase exponentially so that information on smooth exponential decay of the FIR coefficients will be encoded.
	In \eqref{SSKernel}, the decay rate $\alpha$ is one example of kernel hyperparameter
	which is typically unknown and has to be estimated from data.
	
	RKHSs of nonlinear functions are induced for example by the popular Gaussian kernel \cite{Scovel10}
	\begin{equation}\label{GaussKernel}
	K(x_i,x_j) = \exp\left(-\frac{\| x_i - x_j \|^2}{\rho}\right), \quad \rho>0,
	\end{equation}
	where $\|\cdot\|$ is the Euclidean norm while the hyperparameter $\rho$ is referred to as the kernel width. %, another example of hyper.
	The Gaussian kernel is infinitely differentiable. 
	If we use \eqref{eq:FIRx}, %i.e.
	%$$\label{eq:nnfir}
	%x(t)=[u(t),\ldots,u(t-m)]^\top,
	%$$%$\end{equation}
	%\begin{equation}
	% \label{eq:nnarx}
	%    x(t)=[ y(t-1),\ldots,y(t-m-1), %u(t),\ldots,u(t-m)]^\top,
	%\end{equation}
	this fact implies that the related $\mathcal{H}$ now contains nonlinear FIR (NFIR) models
	associated with smooth input-output relationships. Nonlinear ARX (NARX) models follow by including also past output data in $x(t)$ as in \eqref{eq:nnarx}.
	
	Kernels can also implicitly encode a large number of basis functions.
	In fact, under mild assumptions, we have 
	\begin{equation}\label{Kexpansion}
	K(x,z) = \sum_{i=1}^d \zeta_i \phi_i(x) \phi_j(z), \ \ x,z \in X, \ \ \zeta_i>0,
	\end{equation}
	(possibly with $d=\infty$) and the $\phi_i$ span all the RKHS.
	%If they are independent functions, the dimension of $\mathcal{H}$
	%is $d$ (with possibly $d=\infty$) and one has 
	%$$
	%f(x) = \sum_{i=1}^{d}  a_i \rho_i(x) \implies  \|f\|_{\mathcal{H}}^2 = \sum_{i=1}^{d}  \frac{a^2_i}{\zeta_i}.
	%$$
	%Kernels can face the curse of dimensionality by the implicit encoding of functions
	%described by \eqref{Kexpansion}. 
	An important example that illustrates the implicit encoding is the polynomial kernel  \cite{Poggio75}
	\begin{align}
	\label{eq:GPvolt}
	K(x_i,x_j)= \left(x_i^\top x_j +1 \right)^r, \quad r \in \mathbb{N}, %, \quad  c \geq 0,
	\end{align}
	which embeds all the monomials up to the $r$-th degree.
	In particular, the expansion  \eqref{Kexpansion} of \eqref{eq:GPvolt} contains 
	the possibly huge number $d=\binom{m+r}{r}$ of basis functions.
	%where $\langle \cdot, \cdot \rangle$ is the Euclidean inner product.
	%This also means that, i
	Thus, in the NFIR case \eqref{eq:FIRx}, % where input locations are given by \eqref{InpLocNFIR} 
	the polynomial kernel is  
	connected with the important (truncated) Volterra series \cite{Boyd1985,Cheng2017}.

	\subsection{Kernel ridge and Gaussian regression}\label{GaussReg} 
	%and Deep Prediction Networks}\label{GaussReg}
	
	%Then, we search for the estimate of $f$ in the associated RKHS  $\mathcal{H}$ using the squared norm
	%as regularizer, hence generalizing the penalty $\theta^TP^{-1}\theta$ in \eqref{eq:15}.
	
	An important class of kernel estimators makes use of the RKHS norm to control complexity. % of function estimates.
	%to define the ranking of solutions over the space depicted in the right part of Fig. \ref{Fig1}.
	Given $N$ noisy system outputs $y_i$, one of the most used approaches to reconstruct $f$
	is  kernel ridge regression, also called regularization network \cite{PoggioMIT}.
	The estimate is 
	%solution has to balance a quadratic data fit term and the squared RKHS norm, i.e.  
	\begin{equation}
	\label{KernelRidge}
	\hat f=\arg\min_{f \in \mathcal{H}} \sum_{i=1}^N (y_i - f(x_i))^2 + \gamma \| f \|_{\mathcal{H}}^2, 
	\end{equation}
	where the regularization parameter $\gamma$ controls the trades-off between the
	quadratic data fit term and the regularizer.
	This would seem an intricate variational problem possibly defined over an infinite-dimensional space 
	$\mathcal{H}$. Actually, a fundamental result known as the \emph{representer theorem}
	ensures that the (unique) solution has the structure of a neural network with only one layer (also called a shallow network). It is sum of the $N$ kernel sections centred around $x_i$ 
	with coefficients solving a linear set of equations \cite{ArgyriouD2014,Scholkopf01,Wahba:90}. %\dg{Here, we mention the first time a (shallow) network, which is one page into the section. Can we work out the connection to deep learning and therefore the other parts of the survey earlier? We may need to introduce these non-deep learning concepts before but give some motivational aspects before?}
	Specifically, we have 
	\begin{subequations}\label{RepTh}
		\begin{align}
		\label{RNNonLin}
		\hat{f}(x) = \sum_{t=1}^N \ \hat{c}_t K_{x_t}(x) \ \ \forall x,\\
		\label{RNcoeffNonLin}
		\hat{c} = \left(\mathbf{K}+\gamma I_{N}\right)^{-1}Y,
		\end{align}
	\end{subequations}
	where $Y=[y_1, \ldots, y_N]^{\trnsp}$, % while  
	$\mathbf{K} \in \mathbb{R}^{N \times N}$ is the so called
	kernel matrix with $(i,j)$-entry $\mathbf{K}_{ij} = K(x_i,x_j)$
	and $I_{N}$ is the $N \times N$ identity matrix. Hence, the substance of the representer theorem is that if one equates $f(x)$ with observations for various $x$, this will define a linear regression model (with the kernel sections being the regression vectors). This explains why the estimate is similar to that returned by a regularized least squares problem.
	
	If the linear kernel \eqref{LinKernel} is adopted, $f(x)=\theta^{\top}x$ and the estimate of $\theta$ is
	\begin{equation}\label{RegFIR}
	\hat{\theta} = \left(\Phi^{\top} \Phi + \gamma P^{-1}\right)^{-1}\Phi^{\top}Y
	\end{equation}
	with the $t$-th row of $\Phi$ given by
	$$
	\Phi(t,:)=x^{\top}(t), \quad t=1,\ldots,N.
	$$
	If \eqref{eq:FIRx} is used, $\hat{\theta}$ contains the FIR coefficient estimates.
	%\dg{Should we mention here explicitly the difference to the convolutional kernel and its filter coefficients as in~\ref{eq:conv_2layers}?}

	The structure of  \eqref{RepTh} also suggests an important connection between regularization in RKHS and Bayesian estimation \cite{Kimeldorf70,Rasmussen}.
	In this deterministic setting the kernel $K$ represents a similarity function 
	between input locations in $X$.
	%(in some sense) 
	%information between values that functions $f \in \mathcal{H}$ 
	%assume over different parts of the domain $X$. 
	But the kernel can be also seen as a  
	covariance function (they indeed share the same properties). %This leads to Gaussian regression:
	Then, in a Bayesian setting our expected function/system properties can be specified by interpreting $f$
	as a zero-mean Gaussian random field of covariance $\lambda K$ with $\lambda$ a positive scalar.
	Assuming white Gaussian noise of variance $\sigma^2$, estimation theory of jointly Gaussian vectors
	\cite{Anderson:1979} leads to
	$$
	\mathcal{E}(f(x)|Y) = \text{Cov}(f(x),Y)\big(\text{Var}(Y)\big)^{-1} Y = \hat{f}(x), 
	$$
	where $\hat{f}$ indeed corresponds to \eqref{RepTh} setting $\gamma=\sigma^2/\lambda$,
	see also \cite[Section 12]{PillonettoDCNL:14}.\\
	
\begin{figure}
		\begin{center}
			\begin{tabular}{c}
				%\hspace{.1in}
				{ \includegraphics[scale=0.23]{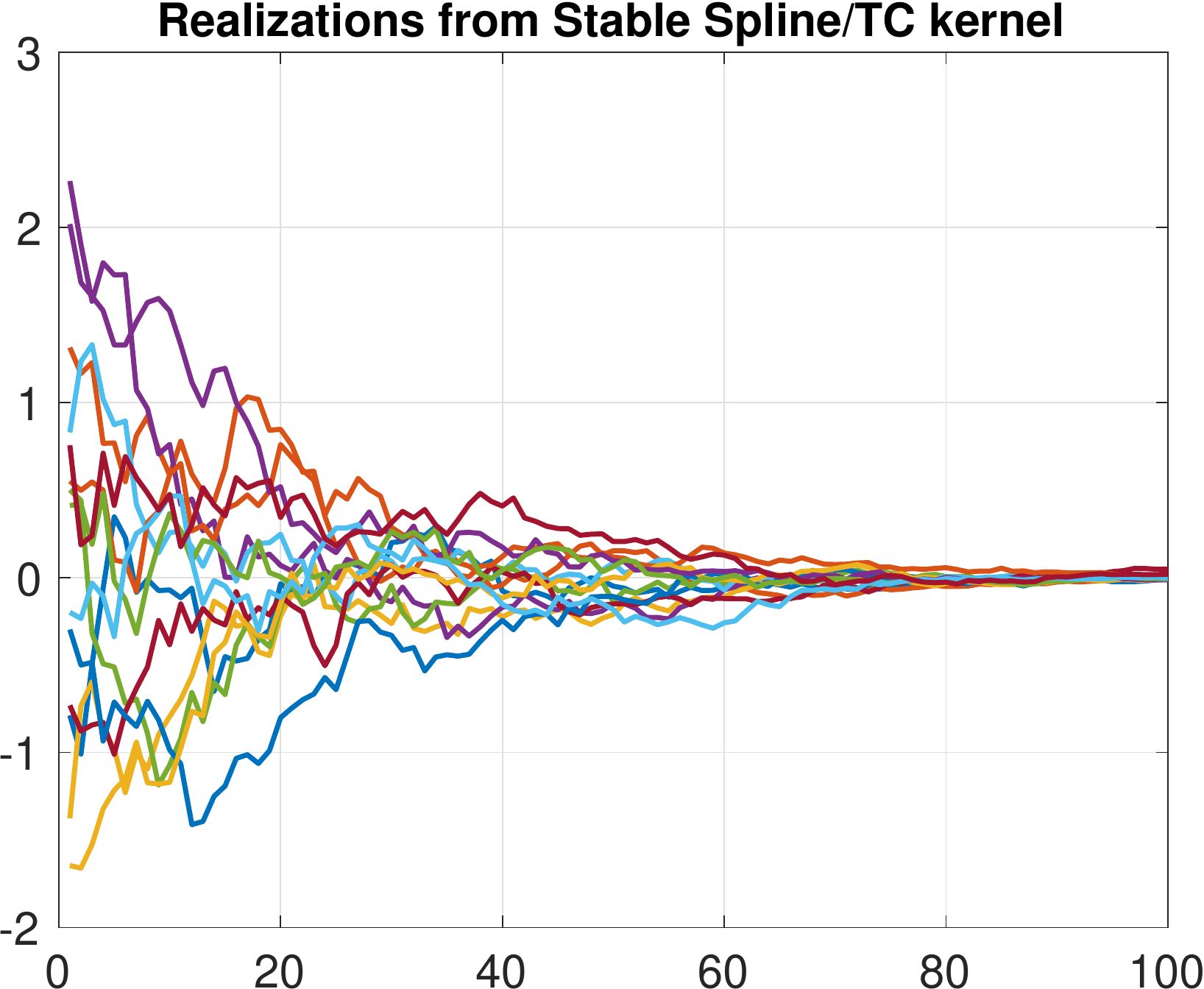}}     
				{ \includegraphics[scale=0.23]
					{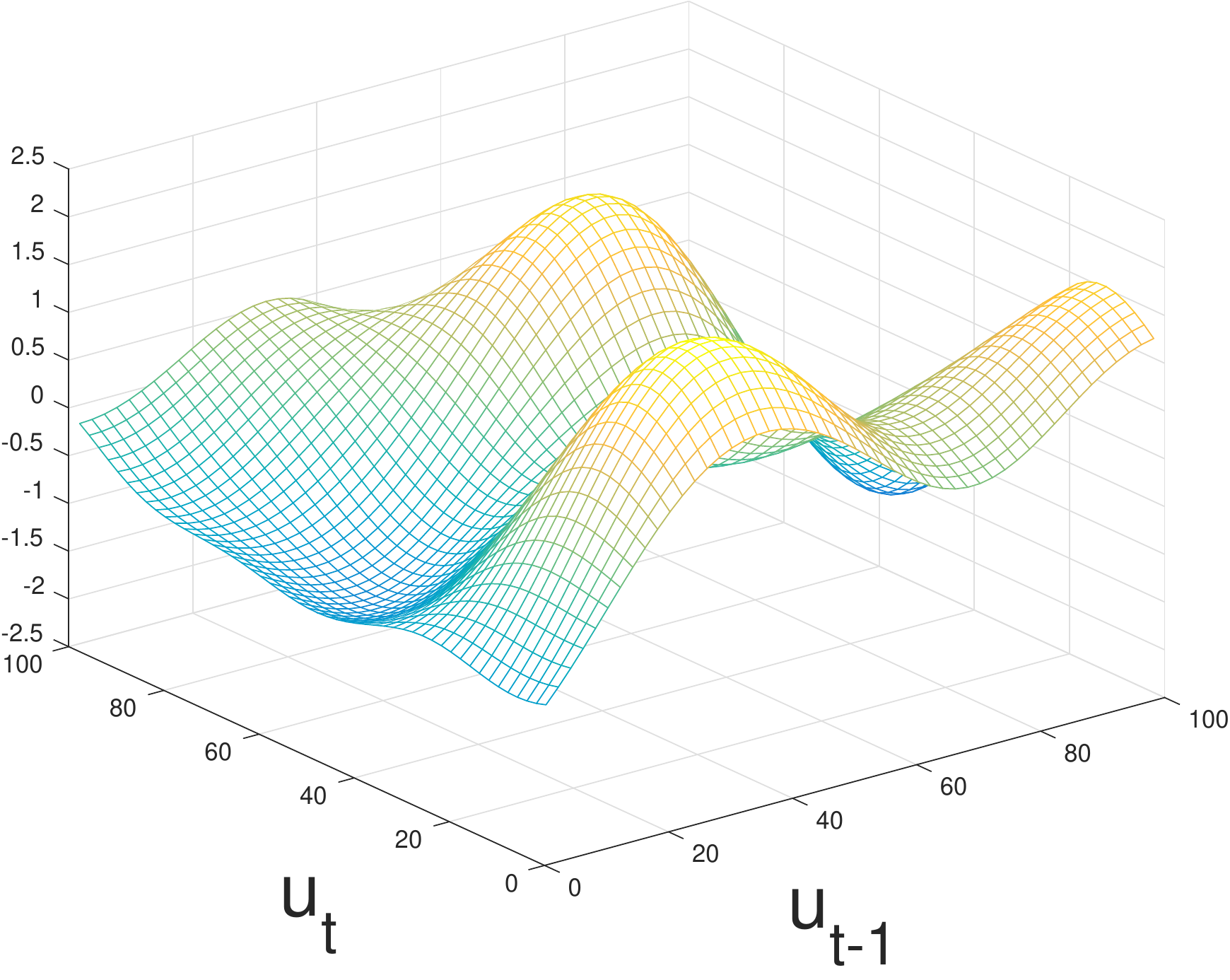}}  
			\end{tabular}
			\caption{\emph{Bayesian interpretation of regularization}. Linear setting (left): realizations of impulse responses modeled as zero-mean Gaussian vectors with covariance set to the stable-spline/TC kernel \eqref{SSKernel}. Nonlinear setting (right): realization of an NFIR $f(u(t),u(t-1)$ modeled as a zero-mean random Gaussian field with covariance equal to the Gaussian kernel \eqref{GaussKernel}.
			} \label{FigRandomFields}
		\end{center}
	\end{figure}

	A simple way to understand the model features encoded in a kernel is to exploit the Bayesian setting by drawing some realizations from a Gaussian process/random field of zero-mean and covariance proportional to $K$.
	In the linear setting, using \eqref{LinKernel} and \eqref{eq:FIRx}, 
	%any function is $f(X)=\theta^\top x$ and 
	the FIR coefficients are the components of the Gaussian vector $\theta$ with covariance $P$. Fig.~\ref{FigRandomFields} (left) plots  realizations from the stable spline/TC prior \eqref{SSKernel}. They all exhibit a smooth exponential decay, confirming what was discussed in a deterministic setting below \eqref{SSKernel}.
	
	In the nonlinear setting, the Gaussian kernel \eqref{GaussKernel} models the input-output relationship $f(x)$ as a smooth random surface drawn from a stationary Gaussian random field.   
	Fig.~\ref{FigRandomFields} (right) plots a realization %of the model 
	in the NFIR case $f(u(t),u(t-1))$. %with system memory $m=2$ (noiseless system output is $y(t)=f(u(t),u(t-1)$).
	
	\subsection{Deep learning using kernels over an intermediate latent space}\label{IntLatSpace}
	
\begin{figure*}
		\begin{center}
			\begin{tabular}{c}
				{\includegraphics[scale=0.35]{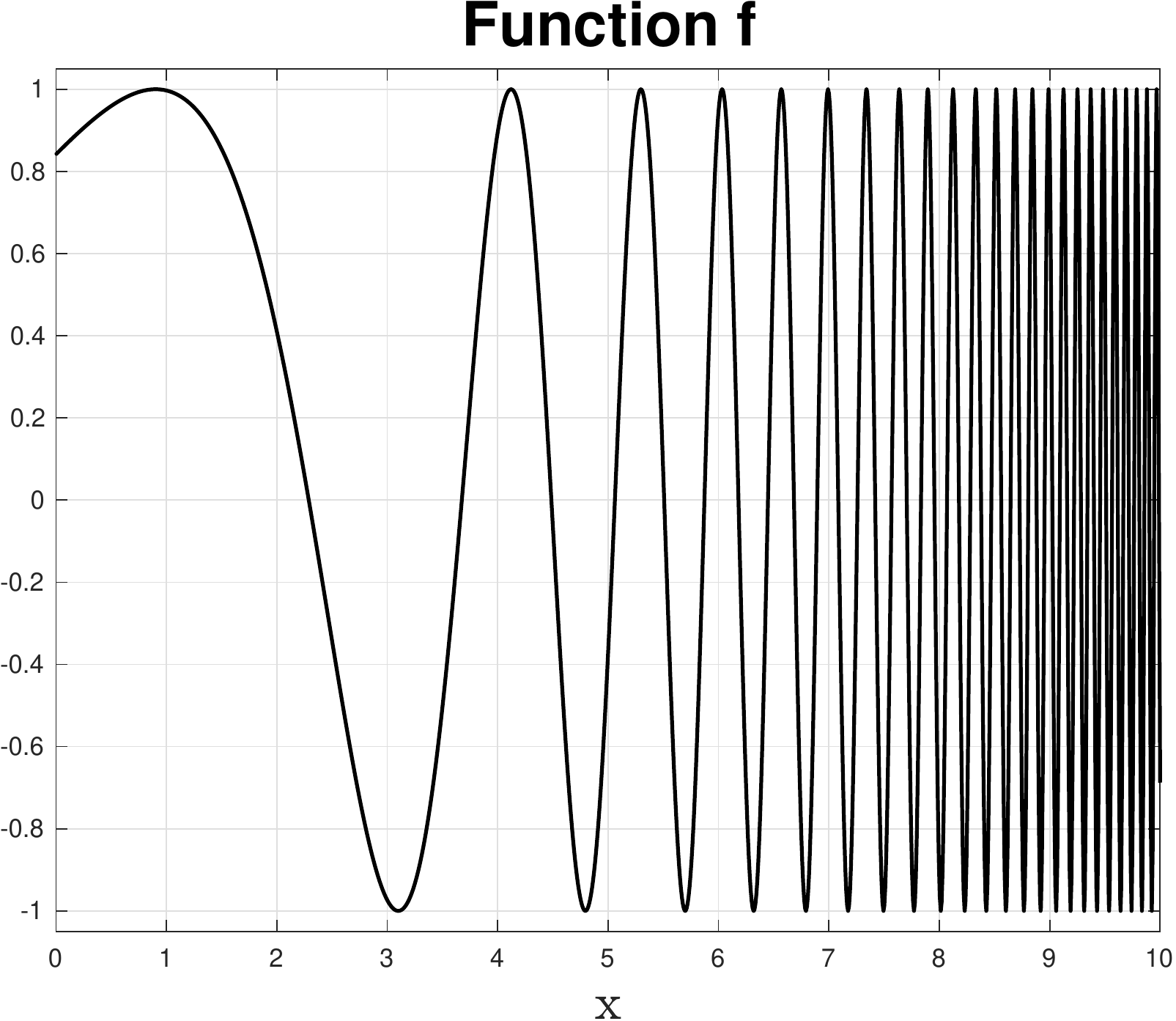}} \   {\includegraphics[scale=0.35]{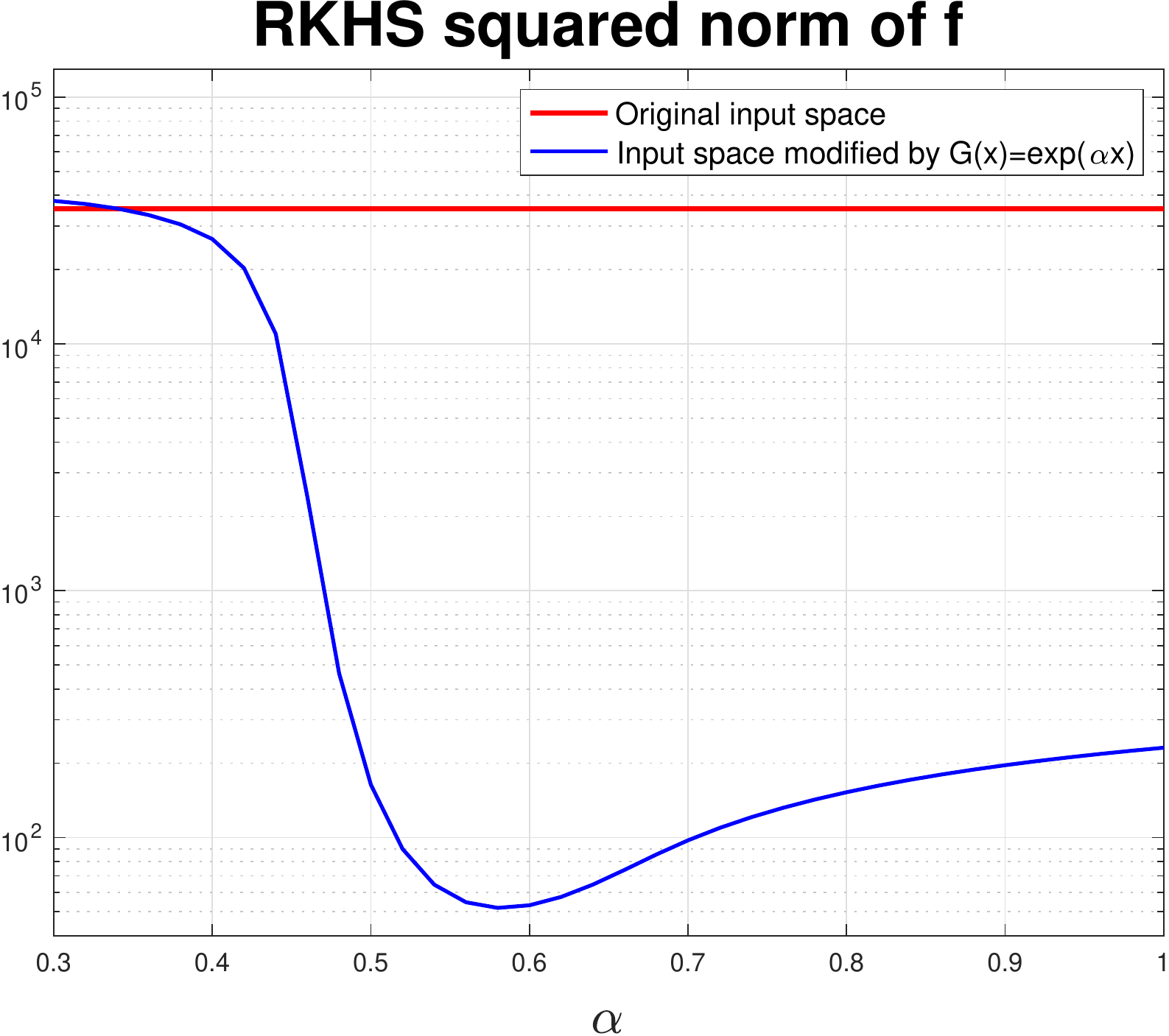}}  
			\end{tabular}
			\caption{\emph{Left}: the function $f=\sin(e^{x/2})$. Due to the presence of the exponential function
				the oscillations increase with~$x$. 
				\emph{Right}: the squared RKHS norm of this function, measured by the Gaussian kernel
				\eqref{GaussKernel} with $\rho=0.1$, is numerically computed by approximating $f$ with 1000 equally spaced kernel sections.
				It is very large, given by the vertical value of the red line.  Unless $\gamma$ is set to a very small value (with the risk of overfitting) the estimator \eqref{KernelRidge} 
				will be severely biased:
				the ranking induced by the Gaussian kernel privileges much smoother functions. The blue line shows the RKHS squared norm,
				obtained by mapping the input space through $G(x)=\exp(\alpha x)$, as a function of $\alpha$. Such transformation can modify the ranking, 
				%allowing \eqref{GaussKernel}  to describe $f$ 
				greatly reducing the penalty term value present in \eqref{KernelRidge} assigned to $f$ by \eqref{GaussKernel}. 
			} 
			\label{FigKernelG}
		\end{center}
	\end{figure*}

	%When using \eqref{RepTh}, the choice of the kernel has a 
	%major effect on the quality of the estimate.
	We start by describing an approach where only the last layer of 
	a deep architecture is defined by a kernel.\\
	Kernel models like % \eqref{LinKernel}  and 
	\eqref{GaussKernel} are good at encoding smoothness properties. % such as smoothness. % or impulse responses stability.
	At the same time they can struggle when it comes to describing functions 
	with complicated frequency contents. An example is provided 
	in the left panel of Fig.~\ref{FigKernelG}. This function has many oscillations and it is not reasonable to model it e.g. through the Gaussian kernel \eqref{GaussKernel}. In fact, the Bayesian interpretation of regularization introduced at the end of the last section says that \eqref{GaussKernel} describes stationary stochastic processes. Other examples where this kernel may fail include discontinuous maps as in hybrid dynamic systems  \cite{JPR:06,OhlssonL:13,Ohlsson2010,PillonettoHybrid}.
	Kernel flexibility can be enhanced
	using combinations of kernels  
	\cite{Duvenaud2013} e.g. in the setting of the %working e.g. in the framework of the
	so called multi-task learning \cite{Ling2019,Maurer2016,Xu2018,Zhang2019}. 
	Advantages of sums of stable spline/TC kernels in linear system identification 
	are described in \cite{CALCP14}.
	However, obtained models 
	cannot stray too far from the original ones, still mainly 
	encoding high-level assumptions such as smoothness.
	A different route is to think of the unknown map as the composition 
	of two functions, i.e. $f= \tilde{f} \circ G$. Here, $\tilde{f}$ is a regular function 
	over $\tilde{X}$ 
	which can be modeled well using traditional kernels. % like the Gaussian one.
	Instead, $G$ has to map the original function domain 
	into an intermediate latent/feature space, i.e.
	\begin{equation}\label{MapG}
	G: \mathcal{X} \rightarrow \tilde{\mathcal{X}},
	\end{equation}
	trying to find useful data representations (features) for the regression problem.
	The usefulness of $G$ to enhance the expressive power of the Gaussian kernel \eqref{GaussKernel}
	%to reconstruct e.g. a function with power at high-frequencies 
	is illustrated in Fig.~\ref{FigKernelG} through a very simple example.
	Here, considering the function $f=\sin(e^{x/2})$ in the left panel, an exponential transformation $G$
	of the domain $X$ may lead to a smooth function $\tilde{f}$ over the latent space $\tilde{X}$.
	For instance, with $G(x)=e^{x/2}$, we have that $f=\tilde{f} \circ G$ with $\tilde{f}(x)=\sin(x)$ 
	which does not contain any high-frequency oscillations. Hence, it can be conveniently estimated by kernel ridge regression e.g. with a Gaussian kernel.
	%This can also counteract the curse of dimensionality,
	%e.g. discovering that inputs lie on a certain manifold.
	%Usefulness of $G$ to enhance the expressive power of the Gaussian kernel \eqref{GaussKernel}
	%to reconstruct e.g. a function with power at high-frequencies 
	%is illustrated in Fig. \ref{FigKernelG} through a very simple example.\\

	One approach is to determine $G$ through unsupervised learning, where 
	only the input locations are used. As a very simple example, let $x_i=[a_i \ b_i]$ and assume that its two components are related via a deterministic relationship. This can be learnt from the input locations and then one can e.g. use $G([a_i \ b_i])=a_i$.  
	This faces the curse of dimensionality by reducing function estimation over $\mathbb{R}^2$ to a problem over $\mathbb{R}$. For these purposes, one can use dimensionality reduction techniques such as PCA, Kernel PCA and also the approaches discussed in Section~\ref{sec:latent-variable} like the auto-encoders~\cite{Hastie01,KernelPCA98,Vincent2008}. 
	See also \cite{OhlssonRGL:07} for manifold learning applied to nonlinear system identification.
	Another technique models $G$ 
	using deep networks %to model $f_1$  %deep networks overviewed in the previous sections
	estimated %jointly with kernel hyperparameters 
	from the input-output data \cite{MacKay98:gp}.
	This corresponds to using~\eqref{DeepStart}
	with $G= f^{L-1} \circ \ldots \circ f^2 \circ f^1$ described by a deep network
	whose outputs live in $\tilde{X}$
	and $\tilde f = f^L$ described by a kernel $\tilde{K}$ over $\tilde{X} \times \tilde{X}$.
	%Formally, the map $G$ now depends on the parameters of the neural network.
	%Formally, if $\tilde{K}$ is the chosen kernel over $\tilde{X} \times \tilde{X}$, % and $\tilde{x}_i=G(x_i)$,
	The overall kernel becomes
	\begin{equation}\label{OverallK}
	K(x_i,x_j) := \tilde{K}(\tilde{x}_i,\tilde{x}_j) =  \tilde{K}(G(x_i),G(x_j)).
	\end{equation}
	While traditional kernels originally adopted in machine learning depend on a limited number of
	parameters (e.g. just $\rho$ using the Gaussian kernel \eqref{GaussKernel}), 
	$K$ can now depend on a parameter vector, denoted by $\eta$, of possibly large dimension. In fact, it contains not only the 
	regularization parameters and the hyperparameters of $\tilde{K}$,
	but also the networks weights which define $G$.
	
	If $G$ is a deterministic map, \eqref{OverallK} can be seen as the covariance of a %zero-mean
	\emph{manifold Gaussian process} $f$ \cite{Calandra2016}.
	The stochastic interpretation is useful for tuning $\eta$   
	via the \emph{Empirical Bayes} (EB) method \cite{ABCP:14,Chiuso16,Efron1973}. It exploits 
	the \emph{marginal likelihood} defined by marginalization of the joint density 
	$\mathrm{p}(Y,f | \eta)$ with respect to $f$. 
	Under the same assumptions used in Section \ref{GaussReg}, we obtain
	$$
	Y \sim \mathcal{N}(0,Z(\eta)), 
	$$
	with
	$$
	Z(\eta) = \lambda \mathbf{K} + \sigma^2  I_{N}.
	$$
	The estimate of networks weights and hyperparameters is
	then given by 
	\begin{equation} \label{MLestimate}
	\eta^{\mathrm{ML}} = \arg\min_{\eta} \ Y^{\trnsp}  \big(Z(\eta))^{-1} Y +\log\det(Z(\eta)).
	\end{equation}
	Finally, the deep estimate of $f$ is available in closed form
	using \eqref{RepTh} with $\eta$ replaced by $\eta^{\mathrm{ML}}$.

	The use of the marginalized likelihood may control complexity.
	In fact, the likelihood $\mathrm{p}(Y| \eta)$ can be approximated
	as the product of the full likelihood and an Occam factor
	that penalizes unneccessarily complex systems \cite{MacKayNC92,PC16}
	and corresponds to $\log\det(Z(\eta))$ in \eqref{MLestimate}.
	However, the use of deep networks %One drawback related to the use of EB is that 
	makes \eqref{MLestimate} a (non-convex) problem 
	over a possibly very high-dimensional domain.
	Hence, in presence of many weights and hyperparameters, 
	the Occam factor of the marginal likelihood can be insufficient to avoid overfitting, as described in \cite{Ober2021}.
	A remedy can be the adoption of a \emph{fully Bayesian} method.
	
	In the full Bayesian setting here of interest, also $\eta$ is seen as a random vector while $f$ is Gaussian conditioned on $\eta$. The problem is that computing the posterior mean of~$f$ now requires the solution 
	of an analytically intractable integral since also marginalization over $\eta$ is required. These problems can be handled 
	by techniques like Markov chain Monte Carlo \cite{Gilks}, 
	where the posterior of~$f$ and~$\eta$ is reconstructed in sampled form. 
	The key advantage is to achieve deep estimates and Bayes intervals around them 
	that take into account all the weights and hyperparameters 
	uncertainty. A drawback is that stochastic simulation 
	can be computationally demanding and require the design of proposal densities 
	(to generate the Markov chains) whose tuning can be delicate.  
	See e.g. \cite{Hendriks2021,Ninness2010,Schon2015}  for
	%for some recent 
	contributions on this subject in a system identification framework.

	\subsection{Deep learning using vector-valued RKHSs}\label{Vector-valued}
	
	In this technique, all the layers of a deep architecture are defined by different kernels.
	The main idea is that maps like $G$
	used in \eqref{OverallK} to enhance kernel expressivity can be
	itself searched for in RKHSs.% defined by other kernels.\\ 
	
	To build deep kernel-based architectures, the RKHSs so far described can be too restrictive since they contain
	only real-valued functions. The extension of the theory described in the literature
	can appear tough since it considers RKHSs of functions $f: X \rightarrow \mathcal{Y}$ 
	where $\mathcal{Y}$ is a Hilbert space with an inner product $\langle \cdot, \cdot \rangle_{\mathcal{Y}}$ \cite{Micchelli:2005}.
	Such spaces are induced by particular operator-valued kernels $K$ which map $X \times X$
	into $\mathcal{B}(\mathcal{Y})$ which denotes the set of linear and bounded operators from $\mathcal{Y}$ into itself. % \dg{what is $\mathcal{B}$ here?}.
	Note that in the previous subsections we just considered %we just considered the case 
	%$\mathcal{Y}=\mathcal{B}(\mathcal{Y})=\mathbb{R}$. Now,
	$\mathcal{Y}=\mathbb{R}$. Now,
	the operator-valued RKHSs are in one-to-one correspondence with kernels which are
	\begin{itemize}
		\item \emph{symmetric}, i.e. $K^{*}(a,b)=K(b,a)$  
		where $K^{*}$ is the adjoint operator of $K$ \cite{Zeidler};
		\item \emph{positive semidefinite}, i.e.
		$$
		\sum_{i=1}^{p}\sum_{j=1}^{p} \langle   y_i,K(x_i,x_j)y_j \rangle_{\mathcal{Y}} \geq 0, %becomes the covariance between p Gaussian vectors of dimension d
		$$
		for any finite natural number $p$, $x_i \in X$, $y_j \in \mathcal{Y}$, with  
		$K(x_i,x_j)y_j$   
		that denotes the operator $K(x_i,x_j) \in \mathcal{B}(\mathcal{Y})$ applied to 
		$y_j$. % one obtains $K(x,x_i)y_i \in \mathcal{Y}$ 
	\end{itemize} 
	Still in analogy with the scalar case, the associated RKHS 
	is generated by kernel sections which are
	functions $f(\cdot)$ from $X$ into $\mathcal{Y}$ defined---for any couple
	$(x_i \in X, y_i \in \mathcal{Y})$---by $K(\cdot,x_i)y_i$.
	
	These generalized RKHSs can be used to define nonparametric 
	deep structures. If the output from one layer is $d$-dimensional, 
	spaces of vector-valued kernels are required.
	In this case, we have $\mathcal{Y}=\mathbb{R}^d$ which also implies
	that $\mathcal{B}$ is identified with $\mathbb{R}^{d \times d}$.  
	This greatly simplifies the analysis since 
	%we can now think that 
	the kernel $K$ now maps 
	$X \times X$ into the familiar space of $d \times d$ matrices
	and any $f$ encapsulates $d$ functions $\{f_i\}_{i=1}^d$.
	Furthermore, $y_i$ can be seen as a column vector,
	$K(\cdot,x_i)$ is a collection of $d \times d$ functions
	and the kernel sections $K(\cdot,x_i)y_i$ are vector-valued functions with $d$ components.
	We will later on see that they generate deep kernel-based estimates. Also for this reason, it is relevant to know that, if $f$ is sum of $N$ 
	kernel sections $K(\cdot,x_i)$ through coefficients $c_i$, its norm is
	\begin{equation}\label{NormVV}
	\| \sum_{i=1}^N K(\cdot,x_i)c_i \|^2_{\mathcal{H}} = c^\top \mathbf{K} c,
	\end{equation}
	where the column vector $c \in \mathbb{R}^{Nd}$ gathers all the $c_i \in \mathbb{R}^{d}$ and
	the $Nd \times Nd$ matrix $\mathbf{K}$ is composed by $N^2$ blocks, with
	the $(i,j)$-th given by $K(x_i,x_j)$.
	
	To understand the meaning of $K(x_i,x_j)$, and to obtain guidelines for 
	the kernel choice, it is useful to extend the Bayesian interpretation
	mentioned at the end of Section~\ref{GaussReg}.
	The components $f_i$ of $f$ can be seen as zero-mean jointly Gaussian random fields
	and $K(x_i,x_j)$ represents the cross-covariance matrix of the two Gaussian vectors 
	$f(x_i)$ and $f(x_j)$. 
	If any $K(x_i,x_j)$
	is diagonal, the functions are mutually independent. 
	Coming back to the deterministic setting, this means that any $f_i$ belongs to a scalar-valued RKHS $\mathcal{H}_i$
	and
	\begin{equation}\label{decoupled}
	\| f\|_{\mathcal{H}}^2 = \sum_{i=1}^d \| f_i\|_{\mathcal{H}_i}^2 \quad \forall f \in \mathcal{H}.
	\end{equation}
	Examples of matrix-valued kernels %that can be found in the literature 
	build also upon the relationship with the so-called multi-task learning 
	\cite{Caruana:1997,Thrun97}. They e.g. model spatio-temporal fields \cite{Sarkka2013,Todescato2020} or 
	maps which are sum of a common function (that has to capture similarities among the $f_i$) and of an independent shift specific for any component of $f$ 
	\cite{Bakker2003,Evgeniou:2005,PillPAMI}.

	%\subsection{Deep learning using vector-valued RKHSs}
	
	Flexible deep networks can now be built by specifying a collection of
	the generalized kernels described above. Each kernel is assigned to 
	a layer that becomes associated with a vector-valued function in the 
	RKHS $\mathcal{H}_\ell$ induced by $K_\ell$.
	
	Assume that the architecture contains $L$ layers 
	and that each $f^{\ell} \in \mathcal{H}_{\ell}$ is $d_{\ell}$-dimensional,
	where $\ell=L$ corresponds to the last layer.
	Overall, the deep network defines the following compositional map
	$$
	f(x) = f^L \circ f^{L-1} \circ \ldots \circ f^1(x). %f(x) = f^1 \circ f^2 \circ \ldots \circ f^L(x)
	$$
	%for any input location $x$  given e.g. by \eqref{InpLocNFIR}.
	The system can be estimated from the identification data by the following interesting
	extension of \eqref{KernelRidge}
	\begin{equation}
	\label{KernelRidgeVV}
	%\{ \hat f^\ell \}_{\ell=1}^L = \argmin_{f^\ell \in \mathcal{H}_\ell} %\sum_{i=1}^n (y_i -  f^1 \circ f^2 \circ \ldots \circ f^L(x_i))^2 + 
	\{ \hat f^\ell \}_{\ell=1}^L = \argmin_{f^\ell \in \mathcal{H}_\ell} \sum_{i=1}^N (y_i -  f^L \circ % f^{L-1} 
	\ldots \circ f^1(x_i))^2 +
	\sum_{\ell=1}^L \gamma_\ell \| f^{\ell} \|_{\mathcal{H}_\ell}^2, 
	\end{equation}
	%\dg{overfull equation, split line?}
	where the penalty term $\| f^{\ell} \|_{\mathcal{H}_\ell}^2$ controls the complexity of the $\ell$-th layer.
	Remarkably, %similarly to what happened for shallow kernel-based networks,
	each estimate $\hat f^\ell$ of the $\ell$-th layer 
	belongs to a finite-dimensional subspace of dimension $N d_{\ell}$.
	In fact, according to the so-called \emph{concatenated representer theorem} (CRT),
	each $\hat f^\ell$ is a linear combination of the $d_\ell$-dimensional 
	vector-valued functions %corresponding to the $d_\ell$ 
	defined by the columns of
	$$
	%K_\ell(f^{\ell+1} \circ f^{\ell+2} \circ \ldots \circ f^{L}(x_i),\cdot), \quad i=1,\ldots,n,
	K_\ell(f^{\ell-1} \circ f^{\ell-2} \circ \ldots \circ f^{1}(x_i),\cdot), \quad i=1,\ldots,N,
	$$ 
	see \cite{Bohn2019} where the result is also extended to a much wider class of kernel machines.
	%with losses possibly different from quadratic.
	Using this result in a recursive way, the (possibly infinite-dimensional)
	variational problem \eqref{KernelRidgeVV} is reduced to a finite-dimensional problem. 
	However, contrary to~\eqref{RepTh}, there is no closed-form solution available for the 
	kernel-based deep architecture. One has to face a non-convex optimization problem of dimension 
	%$\sum_{\ell=1}^L nd_{\ell}=n(1+\sum_{\ell=2}^L d_{\ell})$ 
	$\sum_{\ell=1}^L Nd_{\ell}=N(1+\sum_{\ell=1}^{L-1} d_{\ell})$ and, typically, %. Furthermore,
	the regularization parameters $\gamma_\ell$ have also to be determined from data.
	
	As an example, consider $L=2$, a structure with two layers originally studied in
	\cite{Dinuzzo2010} (also connecting it to multi-task learning). The kernels $K_2$ 
	and $K_1$ are real- and vector-valued, respectively. 
	According to CRT, the deep estimate $\hat{f}$ (composition of % the layers estimates 
	$\hat f^2$ and $\hat f^1$) satisfies
	\begin{equation}
	\label{KernelRidgeVV2}
	\hat{f}(\cdot) = \sum_{j=1}^N \alpha_j K_2\left( \sum_{i=1}^N \sum_{k=1}^{d_1} c_{i k} K_1(x_i,x_j) e_{k} , \cdot  \right), % \quad \forall x,
	\end{equation}
	where $e_{k}$ is the column vector of dimension $d_1$ with $k$-th entry equal to one and zero elsewhere.
	The scalars $\alpha_j$ and $c_{ik}$ are the expansion coefficients which can be obtained by optimizing  
	\eqref{KernelRidgeVV} after replacing $f^2 \circ f^1(x)$ with \eqref{KernelRidgeVV2} and using
	\eqref{NormVV} in a recursive way to evaluate the penalty terms.
	It can also be seen that \eqref{KernelRidgeVV2} corresponds to \eqref{KernelRidge} if $K$ is set to the compositional kernel
	$$
	K(x,y) = K_2\left(\sum_{i=1}^N \sum_{k=1}^{d_1} c_{i k} K_1(x_i,x) e_{k} ,\sum_{i=1}^N \sum_{k=1}^{d_2} c_{i k} K_1(x_i,y) e_{k} \right).
	$$
	The last formula also reveals the connection with \eqref{OverallK}: the parametric map $G$ has been replaced by a regularized estimate defined by the kernel  $K_1$.
	
	\subsection{Deep prediction networks}\label{DPredN}
	
	%\ahr{could it make sense to move the discussion below to section 4.4 together with the discussion about the Concatenated representer theorem?}
	
\begin{figure}
		\begin{center}
			\begin{tabular}{c}
				{ \includegraphics[scale=0.33]{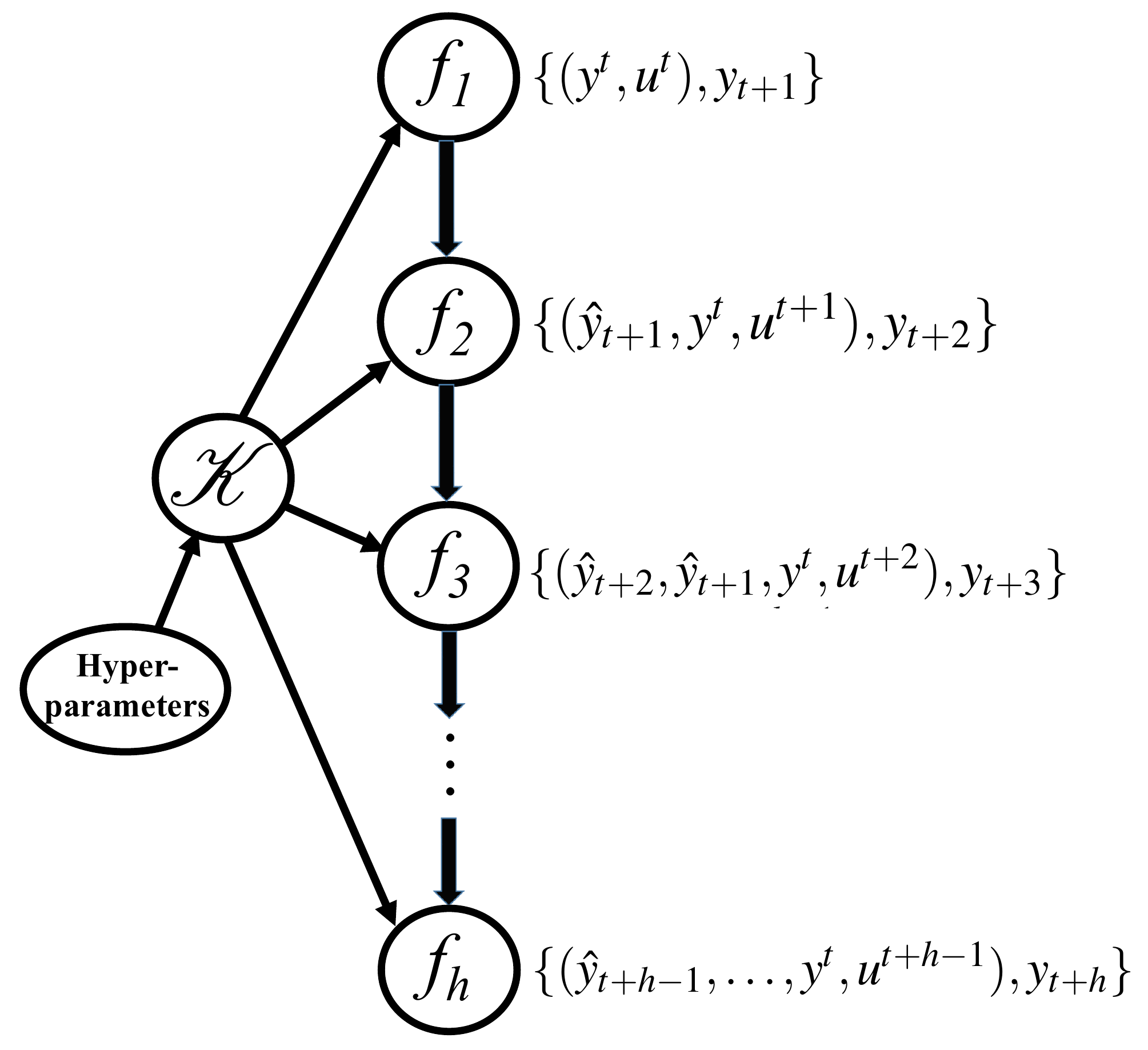}}   
			\end{tabular}
			\caption{Each layer of the network focuses on a certain prediction horizon using the data reported on
				the right of each node. Here, $u^t$ and $y^t$ are, respectively, inputs and outputs 
				collected up to instant $t$. $f_1$ denotes the one-step ahead predictor that is trained  with 
				$\{(y^t,u^t),y_{t+1}\}$ using the kernel-based estimator \eqref{KernelRidge}.
				One step-ahead predictions $\hat{y}_{t+1}$ are then propagated to the second layer and
				$f_2$ is trained exploiting $\{(\hat{y}_{t+1},y^{t},u^{t+1}),y_{t+2}\}$ and \eqref{KernelRidge}.
				This process where output data are progressively
				replaced by their predictions 
				%\textcolor{red}{form new datasets}\dg{can we reformulate the red part?} and to determine new predictors 
				continues up to the desired depth $h$. 
				The layers share a common kernel that provides information on the expected similarities among the $f_i$. Hyperparameters can be e.g. estimated via marginal likelihood optimization using \eqref{MLestimate}.
			} 
			\label{FigDPN1}
		\end{center}
	\end{figure}

	The estimator \eqref{RepTh} can be used in a recursive fashion in a deep structure where any layer 
	is defined by a common kernel and associated to a specific prediction horizon. %different prediction horizon and are defined by a common kernel.\\
	%In particular, the estimator \eqref{RepTh} can be used in a recursive fashion 
	%in a deep structure obtaining the 
	This leads to the so called \emph{deep prediction networks} \cite{DallaL:2021b} whose aim is to improve long-term predictions, 
	a problem already discussed in Section \ref{sec:fully_connected}.
	Any function in \eqref{LongTerm} is assigned to a layer and can be now different from each other, i.e. one now has
	\begin{eqnarray}\label{LongTerm2}
	\hat{y}(t+1|t) &=& \hat{f}_1(y(t),y(t-1),\ldots,u(t+1),u(t),\ldots), \\ 
	\nonumber
	\hat{y}(t+2|t) &=& \hat{f}_2(\hat{y}(t+1),y(t),\ldots,u(t+2),u(t+1),\ldots)\\ \nonumber
	&\vdots& %\\ \nonumber
	%\hat{y}(t+h) &=& \hat{f}_h(\hat{y}(t+h-1),\hat{y}(t+h-2)+h-3},\ldots,u_{t+h-1},u_{t+h-2},\ldots)
	\end{eqnarray}
	Any $f_i$ is also assigned the kernel $K$ and
	the $i$-th layer obtains the estimate $\hat{f}_i$ by kernel ridge regression using the output predictions computed and propagated by the previous layers. Hence, to fit the outputs $y(t)$, the first layer will use \eqref{KernelRidge} with 
	$[y(t-1),y(t-2)\ldots,y(t-m), u(t),\ldots,u(t-m)]^\top$ as input locations, the second one will instead exploit $x(t)=[ \hat{y}(t-1|t),y(t-2)\ldots,y(t-m), u(t),\ldots,u(t-m)]^\top$ and so on, 
	as explained 
	in Fig.~\ref{FigDPN1}. This can be seen as the kernelized version of the concepts exposed in Fig.~\ref{fig:narx-configurations}.
	
\begin{figure}
		\begin{center}
			\begin{tabular}{c}
				%{\includegraphics[scale=0.29]{Figures/FigDeep12}} \   
				{\includegraphics[scale=0.33]{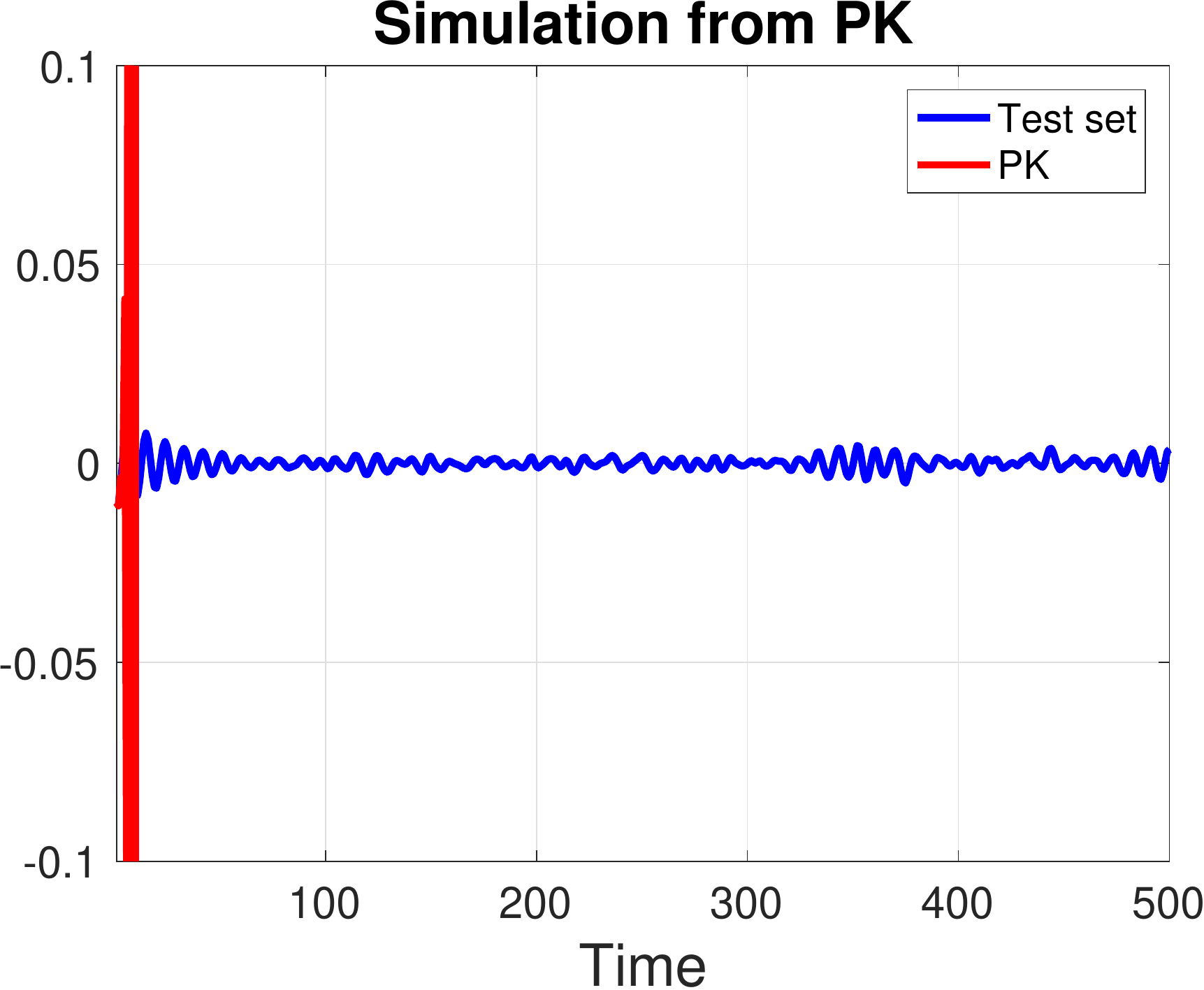}} \\  {\includegraphics[scale=0.36]{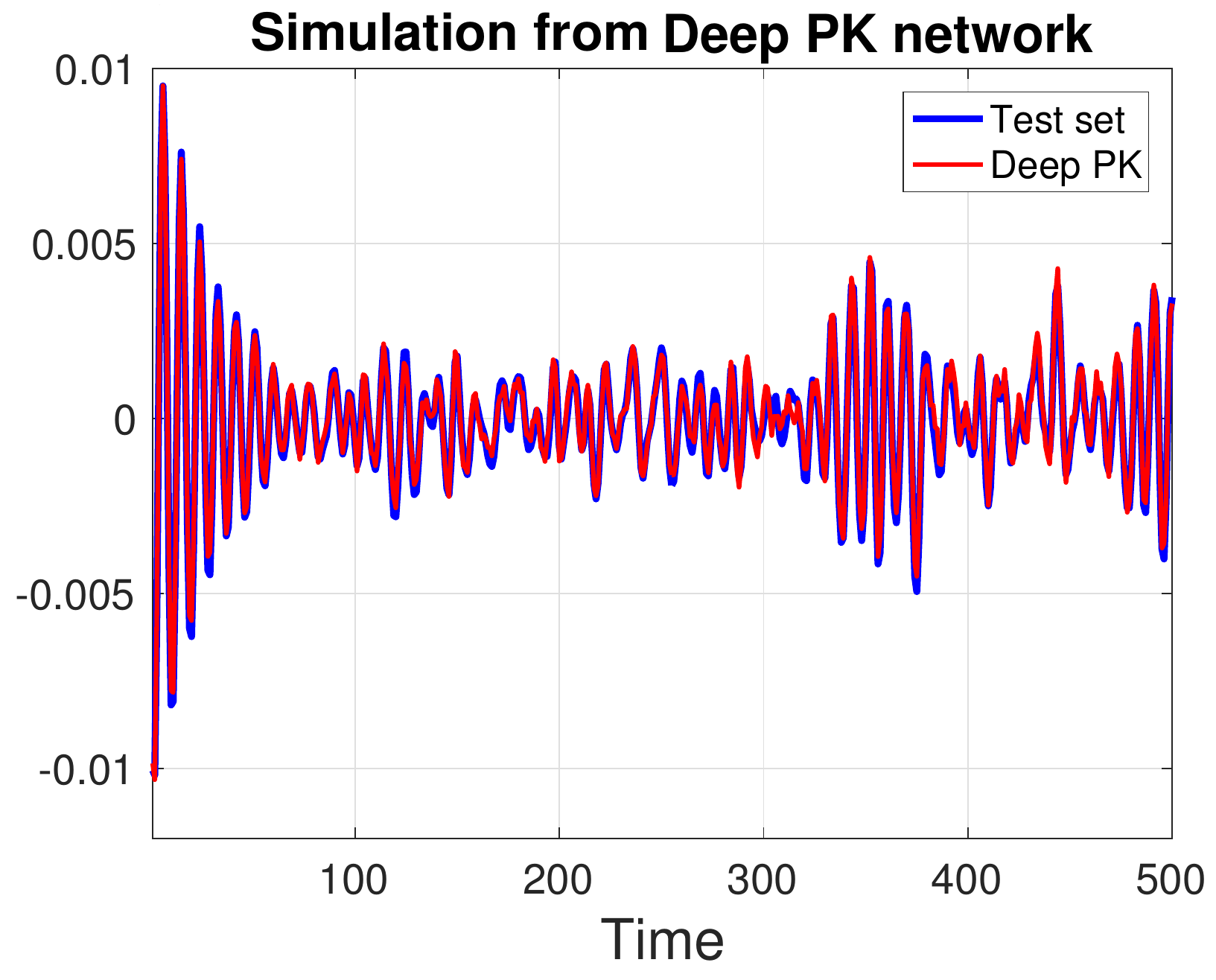}} 
			\end{tabular}
			%\caption{\emph{Left}: input-output 
			\caption{Identification and test data are taken from the Silverbox benchmark problem, see Section \ref{DuffingExample} for details.
				\emph{Top}: test set (blue line) and simulation (red) obtained by \eqref{KernelRidge}
				with the polynomial kernel (PK)  in \eqref{eq:GPvolt} of degree 3. The regularization parameter $\gamma$ is estimated using
				marginal likelihood optimization \eqref{MLestimate}. The estimated model is unstable and the prediction diverges after few time instants.
				\emph{Bottom}: system simulation obtained by plugging the same polynomial kernel of degree 3 
				in a deep prediction network with 500 layers. The regularization parameter $\gamma$ is the same used to obtain the results in the middle panel. 
				Stability problems are no more present and accurate long-term predictions are achieved.
			} \label{FigDPN2}
		\end{center}
	\end{figure}

	An application of a deep prediction network is illustrated in Fig. \ref{FigDPN2}. It shows that the performance of 
	the kernel-based estimator  \eqref{KernelRidge}, 
	used to obtain long-term predictions via \eqref{LongTerm}, can be much improved when plugged in the new computational architecture.
	
	%The performance of a kernel that predicts poorly when used in conjunction with \eqref{KernelRidge} and \eqref{LongTerm}, 
	%can be much enhanced when plugged in the new computational architecture.
	%This is illustrated in the next example using 
	%a benchmark problem proposed in the literature. 
	%Real data come from the Silverbox system, an electronic implementation of the Duffing oscillator 
	%\cite{Wigren2013a}. 
	%Beyond a second-order linear system 
	%it contains also a third degree polynomial static transformation around it in feedback
	%(often present in mechanical systems). 
	%Fig.  \ref{FigDPN2} compares results obtained by \eqref{KernelRidge}  (middle panel) and 
	%a deep prediction network (right) using the polynomial kernel \eqref{eq:GPvolt} of third degree.

	\subsection{Building compositional kernels using limits of deep neural networks}\label{CompKernel}
	%Other forms of compositional kernels 
	%are also discussed e.g. in \cite{KernelDEEP2009}.
	
	Kernels can be defined by a deep neural
	network by placing priors on its weights and letting its width (number of hidden units)
	grow to infinity. This construction also includes the connection
	with Gaussian regression obtained by Neal in \cite{neal_bayesian_1995} and mentioned at the beginning of the section. Recalling \eqref{eq:fully_connected}, any layer $\ell$ of a fully connected neural network is defined by an activation function $\sigma$, weight matrices $W_{\ell} \in \R^{n_{\ell+1} \times n_{\ell} }$ and  bias vectors $b_{\ell} \in \R^{n_{\ell + 1}}$. We model all the entries of
	$W_{\ell}$ and $b_{\ell}$ as independent and zero-mean random variables with variances
	$$
	\text{Var} \ W_{\ell}(i,j) = \frac{1}{n_{\ell}}, \quad \text{Var} \ b_{\ell}(i) =\sigma_b^2.
	$$
	where $n_{\ell}$ is the width of the $\ell$-th layer. Consider just a $L$-layer structure whose output vector is $\hat{y}(x) = \alpha^{L+1}(x)$ where we use $\alpha^{\ell}(x)$ to denote the pre-activation signals in the $\ell$-th layer, i.e., 
	\begin{subequations}
		\begin{align}
		\alpha^{1}(x) &= W_1 x + b_1\\
		\alpha^{\ell+1}(x) &= W_2  \sigma(\alpha^{\ell}(x))  + b_2 \\
		& \text{for }~\ell = 1, \cdots, L\nonumber
		\end{align}
	\end{subequations}
	Any component $\alpha^{\ell}$ is the sum of $n_\ell$ independent random variables. Hence, if the layer's width approaches infinity, all 
	$\alpha^{\ell}$  will converge to  
	Gaussian processes according to the multivariate central limit theorem. 
	They turn  also out mutually independent, allowing to iterate this argument making the hidden layer widths go to infinite in succession \cite{InfiniteDNNMatt,lee2018deep}. In particular, any component 
	of the output from the layer $\ell$ will be a zero-mean Gaussian process 
	$\alpha^{\ell} \sim \mathcal{N}(0, \Sigma^{(\ell)})$ 
	%with kernel denoted by $\Sigma^{\ell}$. If we denote with $\alpha^{\ell} \sim \mathcal{N}(0, \Sigma^{(\ell)})$ a GP with covariance function $\Sigma^{(\ell)}$, the 
	with covariance (kernel) functions given by:
	\begin{equation}
	\label{eq:nngp_kernel}
	\begin{aligned}
	\Sigma^{(1)}\left(x_i, x_j\right) &=\frac{1}{n_{0}} x_i^{T} x_j+\sigma_b^{2} \\
	\Sigma^{(\ell+1)}\left(x_i, x_j\right) &=\mathbb{E}_{\alpha^{l} \sim \mathcal{N}\left(0, \Sigma^{(\ell)}\right)}\left[\sigma\left(\alpha^{l}(x_i)\right) \sigma\left(\alpha^{l}\left(x_j\right)\right)\right]+\sigma_b^{2}
	\end{aligned}
	\end{equation}
	%The following recursive formula for kernel computation is also available $$K^{\ell}(x,z)=\sigma_b^2+\sigma_w^2 C_{\sigma}(K^{\ell-1}(x,z),K^{\ell-1}(x,x),K^{\ell-1}(z,z)), $$ where $C_{\sigma}$ depends on the activation function. Efficient numerical strategies for its computation are described in \cite{lee2018deep}.  
	The advantage of this construction is that the resulting kernel may inherit some features of deep networks which can be implemented through convex estimators like kernel ridge regression \eqref{KernelRidge}. For ReLU activations, a closed-form expression is available and the result is the arccosine kernel originally derived in
	\cite{KernelDEEP2009}:
	$$
	\Sigma^{\ell}(x,z)=\frac{1}{\pi} \|x\|^{\ell}\|z\|^{\ell}J^{\ell}(\beta), \quad \beta=\arccos\left(\frac{x^{\top} z}{\|x\|\|z\|}\right)
	$$
	with
	$$
	J^{\ell}(\beta)=(-1)^{\ell}(\sin(\beta))^{2\ell+1}
	\left(\frac{1}{\sin(\beta)}\frac{\partial}{\partial\beta}\right)^{\ell}\left(\frac{\pi-\beta}{\sin(\beta)}\right).
	$$
	Interestingly, this kernel has distinct features compared e.g. with the Gaussian and polynomial kernels described in \eqref{GaussKernel} and \eqref{eq:GPvolt}, respectively. In fact, it favors nonnegative and sparse data representations, see \cite{KernelDEEP2009} for more details.
	
	\begin{figure*}
		\centering
		\includegraphics[width=0.4\textwidth]{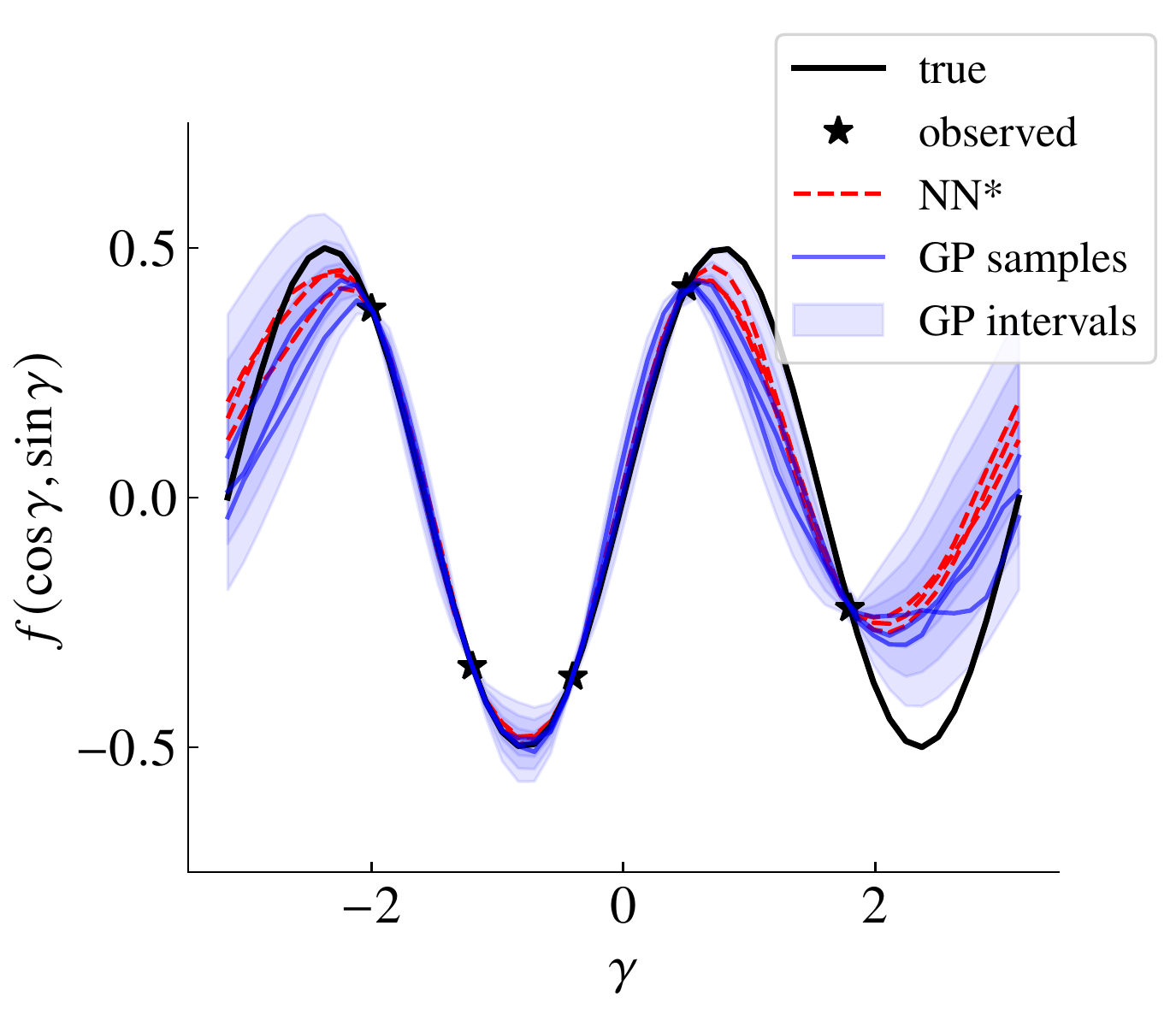}
		\includegraphics[width=0.4\textwidth]{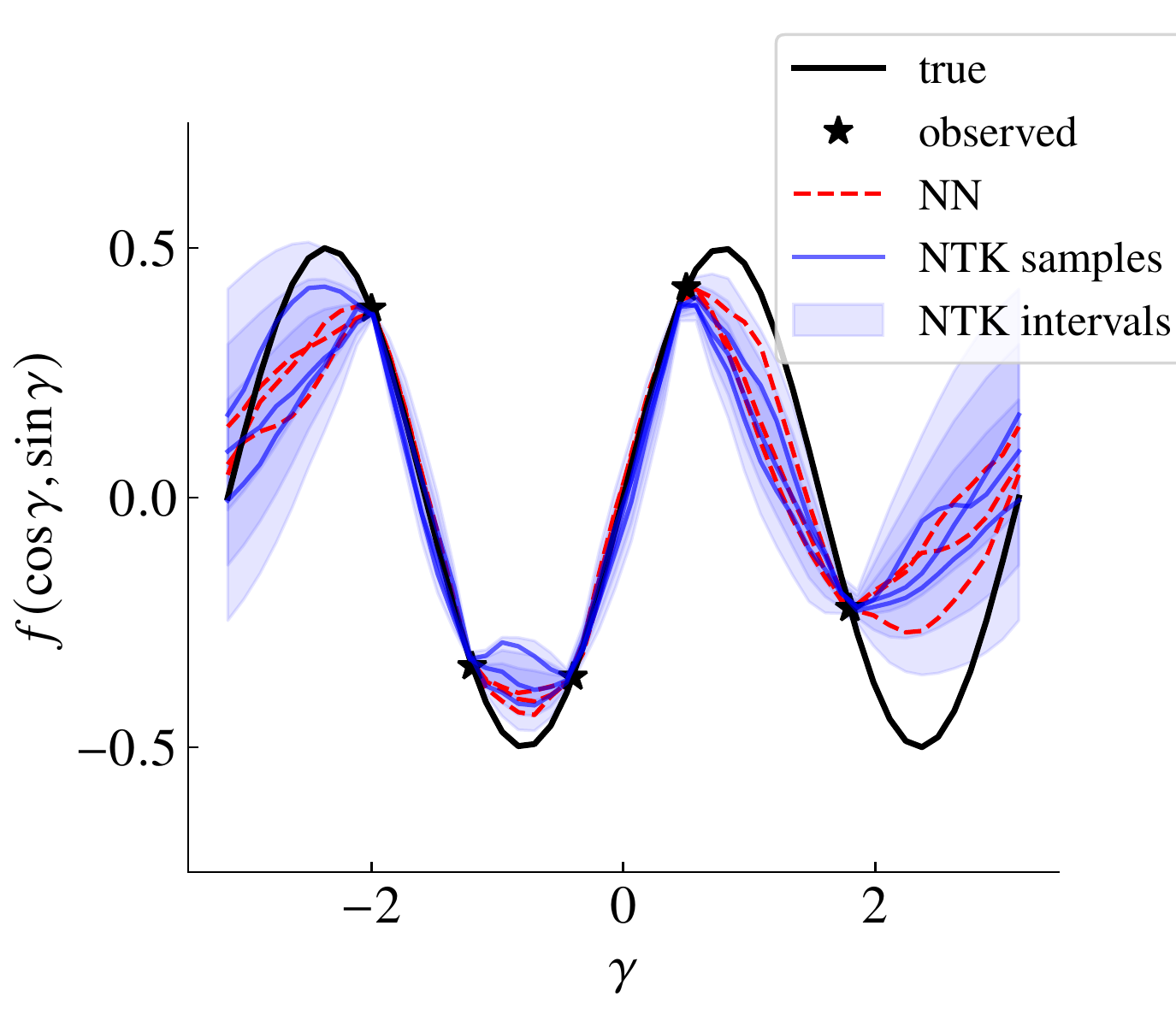}
		\caption{\emph{Left:} Neural network Gaussian process.  \emph{Right:} Neural tangent kernel. The example is inspired by~\cite[Example 6.1]{jacot_neural_2018}. The true function is displayed in \textbf{black} $f(x) = x_{1} x_{2}$. The plot show this function evaluated in the unit circle for $x = (\cos\gamma, \sin\gamma)$. Observed points used for training the model are indicated by $\star$. In both cases, we consider neural networks with $L=4$ layers.  On the left, we compare the neural network Gaussian process (i.e. Eq.~\eqref{eq:nn-gp}) and the predictions of a neural network with weights initialized at random for which only the last layer is trained (NN*). The shaded region shows the 1, 2 and 3 standard deviations of the Gaussian process (i.e. 68\%, 95\% and 99.7\% credible intervals). In \textcolor{blue}{blue}, we have 3 solutions sampled from the posterior of the  Gaussian process and in \textcolor{red}{red} 3 estimates from the neural network for 3 different initializations. On the right, we show an equivalent plot for the neural tangent kernel solution given by \eqref{eq:ntk}. The shaded regions show the 1, 2 and 3 standard deviations (credible intervals).  In \textcolor{blue}{blue}, we have 3 solutions sampled from the Gaussian distribution in Eq.~\eqref{eq:ntk} and in \textcolor{red}{red} 3 solutions of the neural network for different initializations (NN). We consider a neural network with width $n_\ell = 1000$ trained with gradient descent and learning rate $\eta = 1$.}
		\label{fig:nngp}
	\end{figure*}

	A numerical example that illustrates the convergence of a DNN to a
	Gaussian process is given in Fig.~\ref{fig:nngp} (left panel).
	In the infinite wide limit, we have seen that
	the function $\hat{y}(\cdot) = f(\cdot)$ is a Gaussian process with a kernel $\Sigma = \Sigma^{(L)}$ derived from \eqref{eq:nngp_kernel}. Let $X = [x_1, \cdots, x_N]^T$ denote the stacked observed inputs while $z$ is a test point. The function $f$, evaluated on these points, becomes the following Gaussian random vector:
	\begin{equation}
	f\left(\begin{bmatrix}
	z \\
	X
	\end{bmatrix}\right) \sim \mathcal{N}\left(\begin{bmatrix}
	0 \\
	0
	\end{bmatrix}, \begin{bmatrix}
	\Sigma(z, z) & \Sigma(z, X) \\
	\Sigma(X, z) & \boldsymbol{\Sigma}(X, X)
	\end{bmatrix}\right)
	\end{equation}
	where we used $\boldsymbol{\Sigma}(X, X)$ to denote an $N \times N$ matrix with $(i, j)-{\text{th}}$ entry $\boldsymbol{\Sigma}(x_i, x_j)$. Similarly,
	notations like $\boldsymbol{\Sigma}(z, X)$ indicate the cross-covariance matrix between $f(z)$ and $f(X)$ that corresponds to $f$ sampled on $X$.
	Using standard Gaussian conditioning formulas, we can compute the value at the test point conditioned on the observations $f(X)=Y$, i.e.
	$$f \left(
	z\right)
	| \left(f(X) = Y\right)
	\sim
	\mathcal N\left(\mu, C\right)
	$$
	where:
	\begin{subequations}
		\label{eq:nn-gp}
		\begin{align}
		\mu &= \Sigma(z, X) \Sigma(X, X)^{-1} Y \\ C &=  \Sigma(z, z) - \Sigma(z, X) \Sigma(X, X)^{-1} \Sigma(X, z)
		\end{align}
	\end{subequations}
	In Fig.~\ref{fig:nngp} (left) we compare estimates from a neural network
	of large width (and only with the last layer trained) 
	%coming from 3 different initializations
	and from Gaussian regression with covariance defined by the same network
	with the width growing to infinity.
	They turn out quite similar.\\
	
	As final remarks, the idea that infinitely wide neural networks can be seen as Gaussian Processes generalizes 
	%to deep neural networks~\cite{lee2018deep}, 
	to convolutional neural networks~\cite{novak_bayesian_2018} and to recurrent neural networks~\cite{yang_wide_2019}. There are also formulations that try to generalize to a combination of any of these architectures in systematic ways~\cite{novak_fast_2022,yang_wide_2019} and obtain the corresponding kernel from its specification. 
	Other kernels inspired by deep structures can be found in 
	\cite{Damianou2013,Duvenaud2013,Bui2016} in the setting of the so-called deep Gaussian processes.

	\section{Theoretical development}
	\label{sec:theoretical development}
	
	The empirical success of deep learning methods brings to light the question of what makes these models so successful. Deep neural networks make little explicit effort to control model complexity and often have millions or, even, billions of parameters \cite{tan_efficientnet_2019}. While the number of parameters is not always a good measure of the model capacity, deep learning models have been shown to have enough capacity to fit a training set labeled at random~\cite{zhang_understanding_2017} and are therefore interpolators. It has also been observed that these models in some cases seem to indefinitely display increased performance as the model size increases~\cite{tan_efficientnet_2019}. We adopt the terminology of describing models with more parameters than training data points as \textit{overparameterized}.
	
	Inspired by Bartlett et al.~\cite{bartlett_deep_2021}, we would like to highlight two major research developments that have received attention from a theoretical perspective after surprising empirical observations made by practitioners using deep learning. The first studies the role of overparameterization simplifying the non-convex optimization problem that arises when training neural network to a more tractable optimization problem. The second development concerns the interplay between overparametrization and generalization: traditional system identification (and statistical) ideas suggest that flexible models with too many parameters will fit training data too well (up to interpolating the training data). These flexible models should learn spurious components of the data such as noise and provide poor predictive accuracy in new datasets as a consequence. However, there are significant cases where deep learning appears to contradict this accepted statistical wisdom since it performs well on unseen data even when almost perfect fitting to training data is obtained. 
	
	This section is structured as follows: in Section~\ref{sec:infinitely-width-nn} we highlight the recent developments in the understanding of infinitely wide neural networks, explaining how non-convex optimization problems become more tractable as more parameters are used. We then illustrate the idea of implicit regularization (Section~\ref{sec:implicit-regularization}), showing how some optimization choices adopted in deep learning might be responsible for selecting solutions with desirable properties. In Section~\ref{sec:double-descent} we discuss the generalization of models that are overparameterized and perfectly interpolate training data. Section \ref{SecOverfitLinSysid} introduces a simple numerical experiment in the linear setting where overparameterized models perform worse, which highlights that there are examples where overparameterization can be harmful.
	Finally, Section \ref{sec:approxcapability} describes some recent results on approximation capabilities of deep neural networks
	anticipated at the end of Section \ref{sec:modeling-dynamical-sys}.

	\subsection{Infinitely-wide neural networks}
	\label{sec:infinitely-width-nn}
	
	In the limit, when a neural network grows infinitely wide   the optimization problem for training becomes convex in an infinite dimensional space~\cite{allen-zhu_convergence_2019,chizat_global_2018,du_gradient_2019,jacot_neural_2018}. -- Here wide infinitely many nodes in one hidden layer opposed to deep neural networks with many layers. The idea can be understood in simple terms: Let  the parametrized function $f(\cdot; \theta)$ denote the neural network and $\theta \in \mathbb{R}^m$ be the  vector of parameters. Assume that the number of parameters $m$ is very large and that training the neural network changes each parameter just by a small amount w.r.t. its initialization $\theta_0$ during every update step. A linearization of the model around $\theta_0$ yields 
	\begin{equation}
	\label{eq:nn_linearization}
	f(x; \theta) \approx  f(x; \theta_0) + \nabla_\theta f(x; \theta_0)^\top\tilde{\theta},
	\end{equation}
	where $\tilde\theta = \theta - \theta_0$. Hence, the problem can be approximated by an affine problem that could be solved using linear regression. Indeed, it can be established that as the neural network become infinitely wide, the training of the neural network actually amounts to solving a problem similar to that in \eqref{eq:nn_linearization}. 
	
	The notion can be formalized by considering the kernel that corresponds to the transformation $\phi_i: x \mapsto \nabla_\theta f(x; \theta_0)$, i.e.
	\begin{equation}
	\label{eq:neural_tangent_kernel}
	K(x, z; \theta_0) = \nabla_\theta f(x; \theta_0)^\top \nabla_\theta f(z; \theta_0).
	\end{equation}
	which for a fixed $\theta_0$ is a positive-semidefinite kernel as defined in Eq.~\eqref{eq:kernel_def}. 
	The term \textit{neural tangent kernel} was coined in~\cite{jacot_neural_2018} for this kernel since it describes the evolution of the neural networks during their training by gradient descent. Assume that the parameters are continuously updated in the negative direction of the gradient
	\begin{equation}
	\frac{d \theta_t}{d t}= - \eta \nabla_\theta V(\theta_t), \quad V(\theta) = \frac{1}{2N} \sum_{i = 1}^N (f(x_i; \theta) - y_i)^2.
	\end{equation}
	This type of update is sometimes called gradient flow and is a continuous version of the gradient descent method. Here $\eta$ can be seen as the learning rate of the algorithm. 
	A simple application of the chain rule reveals that
	\begin{equation}
	\label{eq:training_ode}
	\frac{d f(z, \theta_t)}{d t} = \frac{\eta}{N} \sum_{i = 1}^N  (f(x_i; \theta_t) - y_i) K(x_i, z; \theta_t).
	\end{equation}
	There are two challenges that make it difficult to apply traditional analysis to the above problem. First, for neural networks the parameters are usually initialized at random. Hence, we can think of $\theta_0$ as a random variable, and the kernel $K(\cdot, \cdot; \theta_0) $ is not deterministic. Second, the kernel depends on a parameter $\theta$ that is itself updated during training. Hence, $K(\cdot, \cdot; \theta)$ is a kernel that evolves with the training itself. 
	
	Jacot et al.~\cite{jacot_neural_2018} present a method of analysis that avoids these two problems: in Theorem 1 of~\cite{jacot_neural_2018} it is shown that in probability, $K(x, y; \theta_0) \rightarrow K^*(x, y)$ as the width of the neural network layers approach infinity, where $K^*$ is a deterministic kernel; Moreover, it shows that if $\theta$ is updated during training in the negative direction of the gradient, then uniformly on all the training paths, $K(x, y; \theta_t) \rightarrow K^*(x, y)$ holds. For a fixed kernel $K^*$, we have that~\eqref{eq:training_ode} is a  linear ordinary differential equation with no explicit dependence on $\theta$. Indeed, if we denote $f^{(t)}(z) = f(z, \theta_t)$,
	\begin{equation}
	\label{eq:training_ode2}
	\frac{d f^{(t)}(z)}{d t} = \frac{\eta}{N} \sum_{i = 1}^N  (f^{(t)}(x_i) - y_i) K^*(x_i, z),
	\end{equation}
	its solution is
	\begin{subequations}
		\label{eq:solution_gradflow}
		\begin{align}
		f^{(t)}(z) &=f^{(0)}(z)   + \sum_{i=1}^N \hat{c}_i^{(t)}  K_{x_i}^*(z), \\
		\hat{c}^{(t)} &= (\mathbf{K}^*)^{-1} (I - e^{-\eta \mathbf{K}^* t})(Y - f^{(0)}(X)),
		\end{align}
		where we denote (for compactness) the kernel matrix built with the input locations collected in $X$ as $\mathbf{K}^* = K^*(X, X)$.
		Moreover, as $t \rightarrow \infty$, we have that $e^{-\eta \mathbf{K}^* t} \rightarrow 0$. Hence, in the limit:
		\begin{equation}
		\hat{c}_i(\infty) = (\mathbf{K}^*)^{-1} (Y - f^{(0)}(X))
		\end{equation}
	\end{subequations}
	Comparing Eq.~\eqref{eq:solution_gradflow} with~\eqref{RepTh} we see that the solution obtained by gradient flow is similar to the solution of kernel ridge regression with parameter $\gamma \rightarrow 0^+$. It does have some additional offset terms $f^{(0)}(X)$ and $f^{(0)}(z)$ that represent the predictions of the neural network at initialization.
	
	As discussed in Section~\ref{CompKernel}, in the infinite wide limit,
	the function $f^{(0)}(\cdot)$ is a Gaussian process with a kernel $\Sigma$,
	\begin{equation}
	f^{(0)}\left(\begin{bmatrix}
	z \\
	X
	\end{bmatrix}\right) \sim \mathcal{N}\left(\begin{bmatrix}
	0 \\
	0
	\end{bmatrix}, \begin{bmatrix}
	\Sigma(z, z) & \Sigma(z, X) \\
	\Sigma(X, z)& \boldsymbol{\Sigma}(X, X)
	\end{bmatrix}\right)
	\end{equation}
	Equation~\eqref{eq:solution_gradflow} is thus an affine transformation of a multivariate normal variable. Some standard manipulation of normal multivariate variables (see~\cite[Section 9.A, Corollary 9.2]{lindholm_machine_2022})
	yields that $f^{(\infty)}(z)$ is normally distributed ${f^{(\infty)}(z)  \sim \mathcal{N}\left(\mu, C\right)}$ for mean and variance:
	\begin{subequations}
		\label{eq:ntk}
		\begin{align}
		\mu &= K^*(z, X) (\mathbf{K}^*)^{-1}Y,\\
		C &= \Sigma(z, z) + K^*(z, X) (\mathbf{K}^*)^{-1} \Sigma (X, X)(\mathbf{K}^*)^{-1}  K^*(X, z)\nonumber \\
		& + K^*(z, X)(\mathbf{K}^*)^{-1} \Sigma (X, z) + \Sigma (z, X) (\mathbf{K}^*)^{-1} K^*(X, z).
		\end{align}
	\end{subequations}
	In Fig.~\ref{fig:nngp} (right) we compare such predictions with  a neural network trained with gradient descent. Note that the mean prediction is the same as kernel ridge regression for the kernel $K^*$, but the presence of randomness at the initialization yield additional terms to the covariance matrix. Similarly to the kernel $\Sigma^{(l)}$ in Eq.~\eqref{eq:nngp_kernel}, the neural tangent kernel  $K^*(x_i, x_j) = \Theta^{(L)}(x_i, x_j)$ can be computed recursively as:
	\begin{equation}
	\begin{aligned}
	\Theta^{(1)}\left(x_i, x_j\right) &=\Sigma^{(1)}\left(x_i, x_j\right) \\
	\Theta^{(\ell+1)}\left(x_i, x_j\right) &=\Theta^{(\ell)}\left(x_i, x_j\right) \dot{\Sigma}^{(\ell+1)}\left(x_i, x_j\right)+\Sigma^{(\ell+1)}\left(x_i, x_j\right)
	\end{aligned}
	\end{equation}
	where,
	\begin{equation}
	\dot{\Sigma}^{(\ell+1)}\left(x_i, x_j\right) =\mathbb{E}_{\alpha^\ell \sim \mathcal{N}\left(0, \Sigma^{(\ell)}\right)}\left[\dot{\sigma}\left(\alpha^{\ell}(x_i)\right) \dot{\sigma}\left(\alpha^{\ell}\left(x_j\right)\right)\right]
	\end{equation}
	and  $\Sigma^{(\ell)}(\bullet, \bullet)$ is defined as in Eq.~\eqref{eq:nngp_kernel} and $\dot{\sigma}$ is the derivative of the activation function. Comparing Fig.~\ref{fig:nngp} panel thus illustrates the difference between neural-network GP and the neural tangent kernel solution. The neural tangent kernel can be generalized to convolutional neural networks~\cite{arora_exact_2019}, recurrent neural networks~\cite{alemohammad_recurrent_2021} and has been formalized in a systematic way through a programming language description~\cite{yang_tensor_2020}.

	The idea that infinitely-wide limit neural network optimization can be convex illustrates how overparametrization can be a way to obtain a more tractable optimization problem. That said, there are studies showing that only the infinitely-wide regime is not enough to justify the good properties of neural networks~\cite{chizat_lazy_2019}. Finite-dimensional neural networks are the solution to certain classes of Banach spaces other than RKHS: in~\cite{parhi_role_2020,parhi_banach_2021,parhi_what_2021}, representer theorems are derived for Neural Networks with ReLU activation functions, showing they are solutions to data fitting in bounded variation spaces. 
	
	These works suggest that the analysis of neural networks in the infinite width limit might not be enough to understand its properties, it also points to limitations of restricting the function space to RKHS. In~\cite{parhi_nearminimax_2021}, they draw an analogy between smoothing-\textit{vs}-adaptive splines and finite-\textit{vs}-infinite ReLU neural network  comparisons. Smoothing spline (that are kernel methods) have a hard time learning functions that are not spatially homogeneous, and adaptive splines are better at fitting functions with changes in the variation over spaces. The authors argue that similar differences hold for the neural network in the infinite-\textit{vs}-finite case.

	\begin{figure*}
		\centering
		\includegraphics[width=0.33\textwidth]{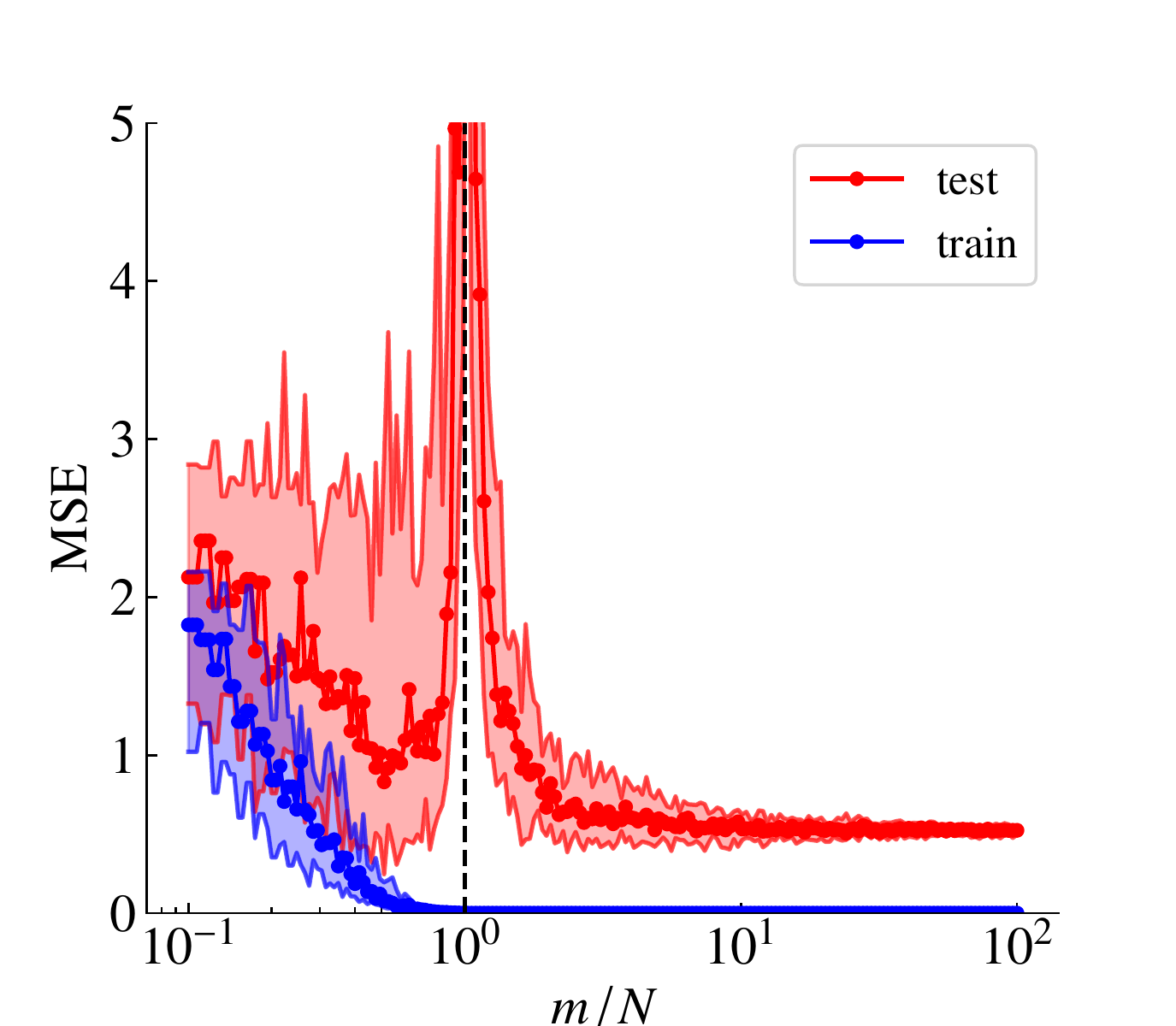}
		\includegraphics[width=0.33\textwidth]{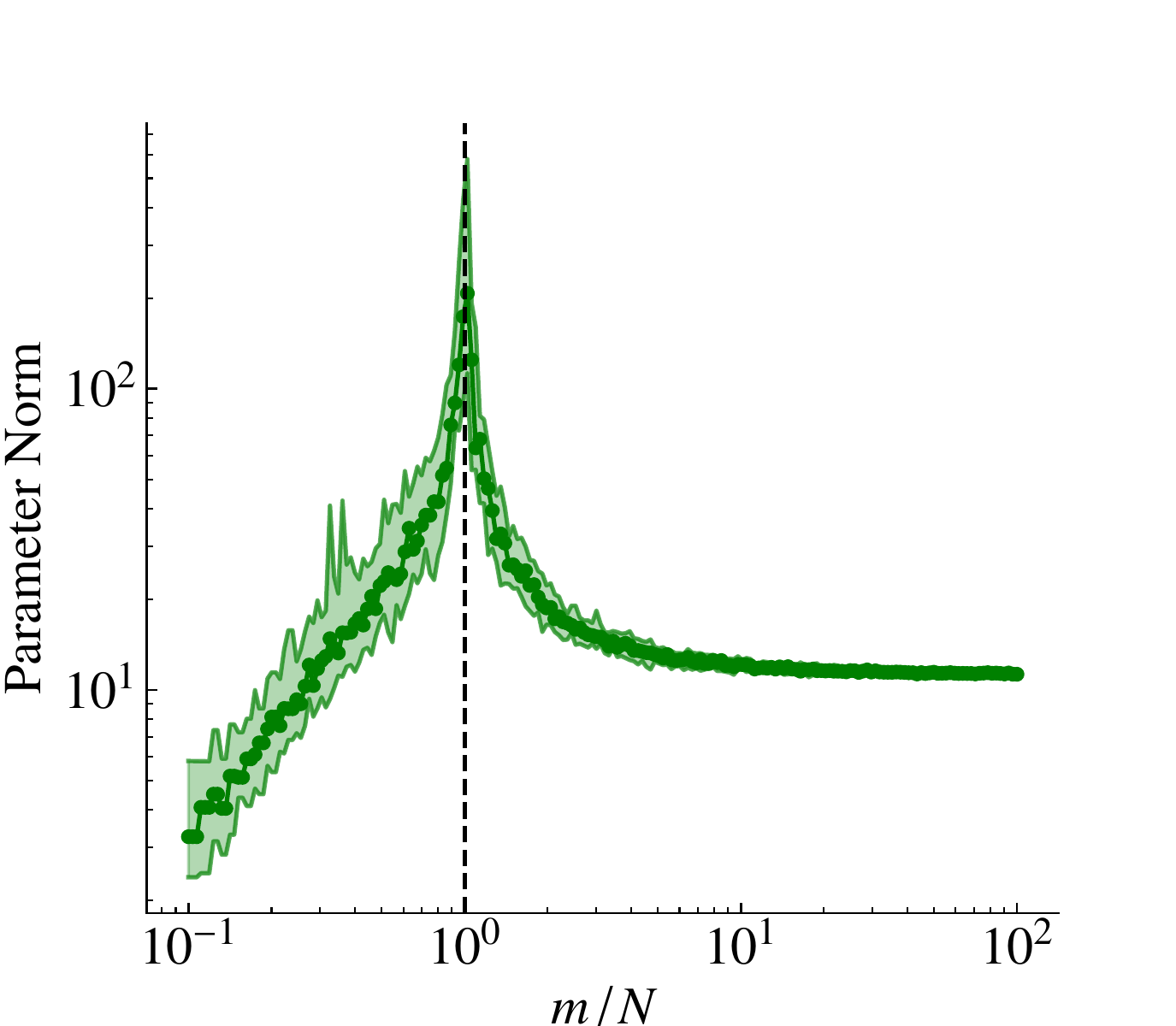}
		\includegraphics[width=0.33\textwidth]{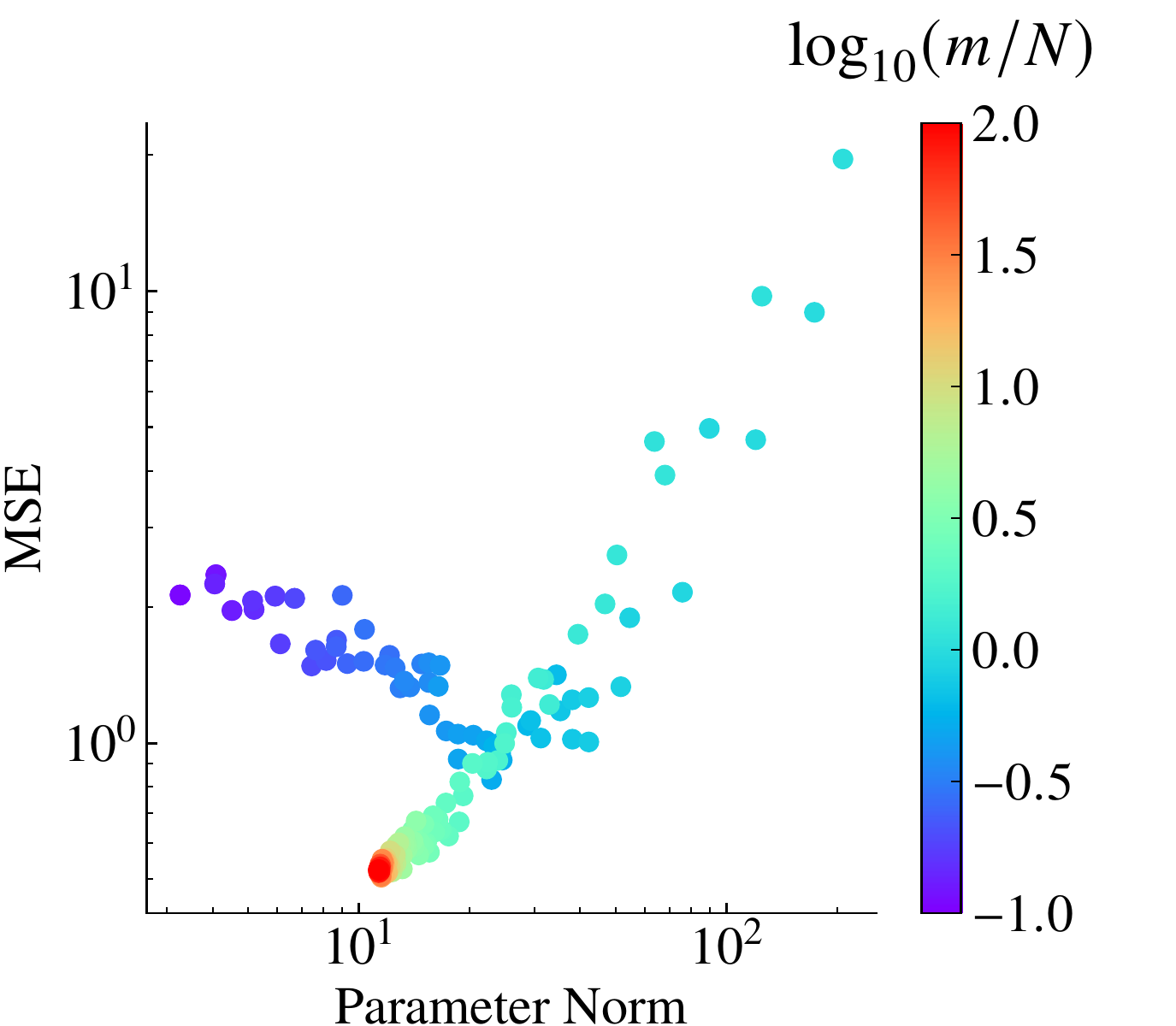}
		\caption{\textit{Double descent in nonlinear ARX models.} \textit{Left:} mean square error of one-step ahead prediction of training data (blue) and test data (red) as a function of the rate between features and number of training data points $m/N$. \textit{Middle:} parameter norm as a function of the rate between features and number of training data points $m/N$. \textit{Right:} mean square error of one-step-ahead prediction of test data as a function of the parameter norm (the proportion is given in the colorbar).  Data are generated using the nonlinear function $f$ presented in~\cite{Chen1990}: $y(t) = f(x(t)) + v(t)$ where $v(t) \sim N(0, 1)$ and $x(t) = [y(t-1), y(t-2), u(t-1), u(t-2)]$.  Identification data contain $N = 100$ samples and a hold-out test set of $100$ samples is then used. Training is performed through a nonlinear ARX model linear-in-the-parameters $\hat{y}(t) = \sum_{i=1}^m \theta_i \phi_i(x(t))$ where the $\phi_i$ map the input into a feature vector. In particular, Random Fourier Features are used \cite{rahimi_random_2008} in order to scale the input dimension w.r.t the number of training samples (this is why $m$ here indicates the number of features in place of the number of  model parameters). When $m > N$, in presence of multiple solutions the one with minimum parameter norm is selected.}
		\label{fig:dd_narx}
	\end{figure*}
	
	\subsection{Implicit regularization}
	\label{sec:implicit-regularization}
	
	In the discussion that followed from Fig.~\ref{Fig1}, we have categorized neural networks as belonging to the ``alternative approach'' to system identification: that is, the model is searched for in a very rich space and the model complexity is controlled through regularization. The classification into this category comes with some nuance though. While regularization methods that include dropout, weight decay and similar are usually included in order to get the best performance, they are not strictly needed. Even without explicit regularization it is still possible to obtain reasonably well performing models. To study this behavior, effort has been put into studying the implicit regularization mechanisms that act in a neural network during its training.
	
	% implicit factor of problem formulation
	A prime example of the implicit regularization through the problem formulation is the matrix completion or reconstruction problem studied in \cite{gunasekar2017implicit}. Here, we have a low-rank true matrix $X^\ast$ from which we only observe some entries yielding the partially observed matrix $X\in\mathbb{R}^{n\times n}$ and we try to regress the full matrix including the non-observed entries
	\begin{equation}
	\min_{X\succeq 0} || \mathcal{A}(X) - y||_2^2
	\end{equation}
	with the linear operator $\mathcal{A}: \mathbb{R}^{n\times n}\rightarrow \mathbb{R}^m$ and $y\in\mathbb{R}^m$ following $y=\mathcal{A}(X^\ast)$. Since $m\ll n^2$ the problem is overparameterized and has many global minima. Instead of directly optimizing this problem, we factorize $X=UU^\top$ with $U\in\mathbb{R}^{n\times d}$, which imposes a rank constraint and makes the problem non-convex. Here, we focus on the case of $d=n$ without rank constraint. We cannot expect the obtained solution to recover the true values or to generalise to unseen test data as we can set the unobserved values to any arbitrary number. When optimizing using gradient descent with small step size, we obtain in both formulations (with and without factorization) zero training error, which indicates that the model memorized the entries. The relative reconstruction error $||X-X^\ast||_F/||X^\ast||_F$ for unseen test data in the non-factorized formulation $X$ is equal to the fraction of non-observed entries. However, for the factorized formulation $X=UU^\top$ the reconstruction test error is surprisingly much lower. Hence, the gradient descent optimization found a specific solution, which generalizes better without any constraint on the problem but only by rewriting the problem formulation. It turns out that gradient descent will in this scenario implicitly induce a low nuclear norm. Hence, the optimization algorithm itself has a form of implicit regularization. 
	
	Extending this observation, \cite{neyshabur2015implicit} use neural networks with one hidden layer trained with SGD. The authors observe that despite increasing the capacity, i.e. the number of hidden units, to and beyond the point where the model fits training data perfectly, generalization to unseen data is not harmed. This indicates that the capacity of the model is not controlled by the number of model parameters when trained with SGD and hints further to a regularizing behavior of gradient descent converging to a minimum norm solution. 
	
	% minimum norm solution
	The solution that is found in the matrix completion problem is for small step sizes in fact the minimum norm solution. Similarly, for overparameterized linear regression let us assume the cost function $V(\theta) = \frac{1}{2}\|X \theta - y\|^2$ and the parameter update $\theta^{i+1} = \theta^i - \gamma \nabla_{\theta} V (\theta^i)$ with gradient descent. Then if the initial parameter $\theta_0$ lies in the row space of $X$ and for a sufficiently small $\gamma$, we have that the solution converges to the minimum-norm solution, i.e. the solution to the following optimization problem
	\begin{equation}
	\label{eq:min-norm-solution}
	\min_\theta \|\theta\|_2 \quad \text{subject to}\quad X\theta = y.
	\end{equation}
	
	% quantify implicit regularization
	Barrett and Dherin tries to quantify the implicit regularization for gradient descent \cite{barrett2021implicit} which is extended in \cite{smith2021on} for stochastic gradient descent with shuffling of the mini-batches. Due to the step-based nature of the algorithm, the optimization does not follow the continuous path of the gradient flow at each step exactly but follows a different trajectory of a modified cost including an implicit regularizer. Assume the same cost function and parameter update as for linear regression above with $\theta\in\mathbb{R}^m$. Then gradient descent follows more closely $\Dot{\theta}=-\nabla_\theta\widetilde{V}(\theta)$ with 
	\begin{align}
	\widetilde{V}(\theta) &= V(\theta) + \lambda R_{IG}(\theta)\\
	\text{with}\quad&\lambda=\frac{\gamma m}{4}, \quad R_{IG}(\theta) = \frac{1}{m}\sum_{j=1}^{m} (\nabla_{\theta_j} V(\theta))^2 \notag
	\end{align}
	where $\nabla_{\theta_j}$ is the gradient in the $j$th direction. Hence, this formulation shows that gradient descent penalize directions with large cost. This penalization implies that gradient descent is biased towards flat minima, i.e. minima where the magnitude of the eigenvalues in $\nabla^2V(\theta)$ are small, which the authors observe empirically.
	% sharpness of solution
	It is observed that that optimization solutions in sharp minima, i.e. with large Hessian, generalize less well. To make this more intuitive, assume that the test cost function is shifted w.r.t. the model parameters compared to the train cost function e.g. due to a change within the datasets. Hence, the obtained solution in the train minimum does not lie in the test minimum. When we found a solution in a sharp minimum this yields a large increase in test cost while a solution in a flat minimum does not yield such a large increase in test cost.
	
	% batch size and sharp minima
	Keskar et al. provide evidence that optimizing with smaller batch sizes yields flat minima solutions~\cite{keskar2017on}. The origin of this can be the stochasticity of smaller batch sizes.
	% stochasticity
	Pesme et al.~\cite{pesme2021implicit} analyse the effect of stochasticity on the dynamics of SGD in the squared parameterization linear regression model \cite{woodworth20a}, also known as diagonal linear network (meaning linear activation functions and diagonal weight matrices). The authors show that training with SGD always generalizes better than with gradient descent. The role of noise in the outputs or labels $y$ during training is analysed by \cite{blanc20a,haochen21a} which show that this stochasticity enforces sparse solution with minimum norm. The effect of the step size (or learning rate) is similar with larger step sizes leading to sharper minima \cite{nacson22a}.

	\subsection{Interpolation and double-descent curves}
	\label{sec:double-descent}

	The generalization properties of deep neural networks motivate the study of models that can (almost) perfectly fit the training dataset. This is also noted as interpolation of training data. Indeed, Zhang et al.~\cite{zhang_understanding_2017} showed that state-of-the-art deep architectures are sufficiently complex that they can be trained to interpolate noisy data. In a series of insightful papers Belkin et al.~\cite{belkin_reconciling_2019,belkin_two_2020,Belkin2018} illustrate that interpolation does not contradict good generalization. In~\cite{belkin_reconciling_2019} the authors coined the term double-descent curve for describing the curve of the model performance vs the number of parameters. The idea is that a ``double-descent'' curve subsumes the textbook U-shaped bias–variance trade-off curve in some scenarios with a second decrease in the error occurring beyond the point where the model has reached the capacity of interpolating the training data.
	
	\paragraph*{Double descent curve}
	Fig.~\ref{fig:dd_narx} describes the double-descent phenomenon in an example adapted from~\cite{Ribeiro2021}. The leftmost plot illustrates the one-step-ahead mean square error in training and test as a function of the number of parameters. A double descent curve can be observed in the test error. When $m < N$, i.e. when the number of parameters is smaller than the dataset size, we can observe a U-shape error curve in the hold-out test set as we vary $m$. However, as the number of parameters keeps increasing, it is possible to reach a point where the model starts to interpolate (usually at $m=N$) and the training error reaches zero. At this point, the model does not generalise well and will perform poorly on the unseen test dataset. However, if we continue to increase the number of parameters to $m>N$, the model is overparameterized and will eventually start to generalise well again on test data.
	
	One important element of the example is that in the overparameterized or interpolation region there are infinitely many solutions and we select the minimum-norm solution i.e. \eqref{eq:min-norm-solution}. In Section \ref{sec:modeling-dynamical-sys}, we described cases where the number of data-points $N$ is larger than the dimension of the features vector $m$ and explained how in this setup sometimes the number of parameters $m$ can be used as a proxy for model complexity, e.g. as in \eqref{eq:625}. This does not apply for the case $m > N$ and indeed increasing the number of features can have the reverse effect.  The middle plot in Fig.~\ref{fig:dd_narx} illustrates that as the number of parameters increases it yields solutions with smaller parameter norm. The rightmost plot in Fig.~\ref{fig:dd_narx} illustrates the one-step-ahead mean square error in the test set as a function of the parameter norm $\hat{\theta}$. This plot does not show a double descent but more a classical U-Shaped bias-variance trade-off. It highlights that the lowest test risk is not obtained in a second descent but rather at a sweet-spot in this U-Shaped curve defined by an intermediate value of the parameter norm.
	
	The double-descent performance curve has been experimentally observed in diverse machine learning settings: Belkin et al.~\cite{belkin_reconciling_2019} show the phenomenon for random Fourier features, random forest and shallow networks, \cite{hastie2022surprises} for ridgeless regression, \cite{nakkiran_deep_2020} for transformers and convolutional network models. Setups closed to the system identification are discussed in~\cite{Ribeiro2021}.

	For completeness, we point out that in some cases it is possible to observe more than two descents in the generalization curve. To cite a few examples where this is demonstrated: Ascoli et al.~\cite{dascoli_triple_2020} empirically showed the triple-descent phenomena to random Fourier model (same model used to produce Fig.~\ref{fig:dd_narx} under different conditions) and~\cite{adlam_neural_2020} show it for the neural tangent kernel. Moreover,~\cite{chen_multiple_2021} illustrates a case with Gaussian mixtures, for which it is possible to obtain arbitrary many descents when the number of parameters increases and the number of training samples is kept constant.

	\paragraph*{Theoretical toolboxes} 
	Many of the traditional asymptotic results in system identification assume that the number of parameters $m$ is fixed and that the number of data points goes to infinity $N \rightarrow \infty$. As we become interested in models where the number of parameters scales with data, different types of results are needed. We mention two interesting sets of tools that can be used to derive theoretical models of the above phenomenon. In describing them, recall that as a proxy for model capacity the number of data features, i.e. covariates in the data, is often considered instead of the number of model parameters. In a single layer model these two measures coincide. Hence, to be compliant with literature which often considers the number of features, we abuse notation and denote both by $m$.
	
\begin{figure*}
		\centering
		\includegraphics[width=0.3\textwidth]{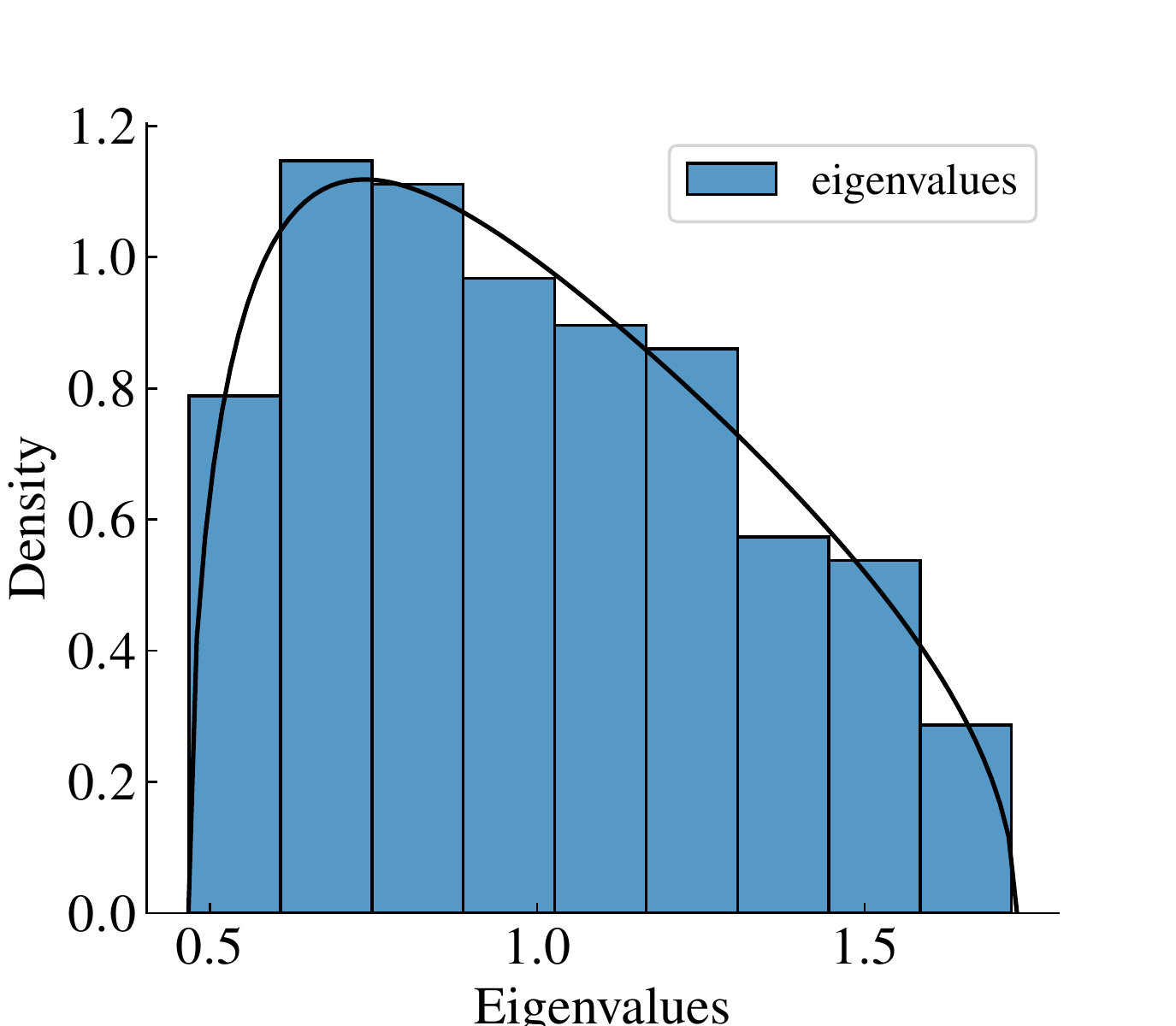}
		\includegraphics[width=0.3\textwidth]{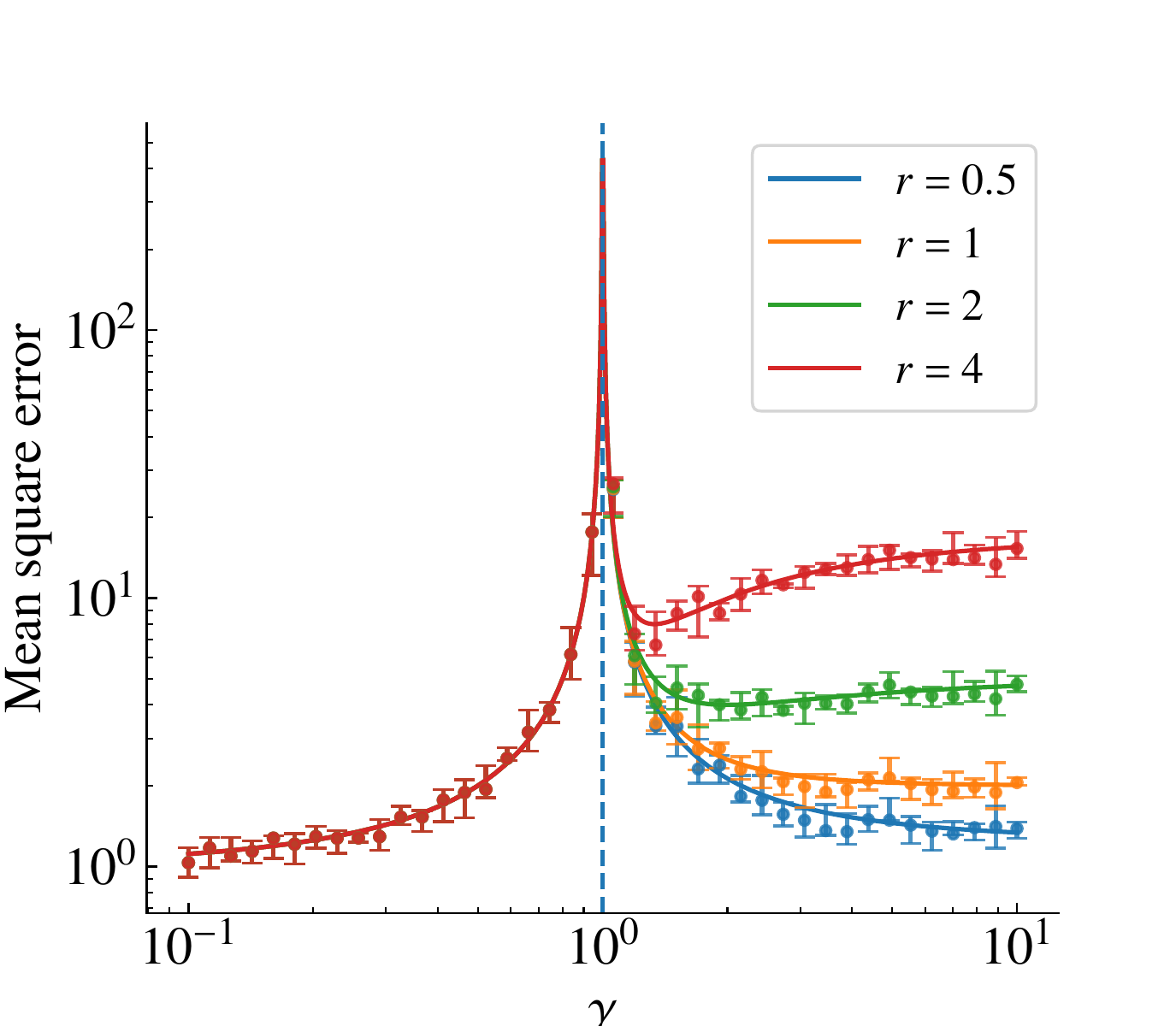}
		\includegraphics[width=0.3\textwidth]{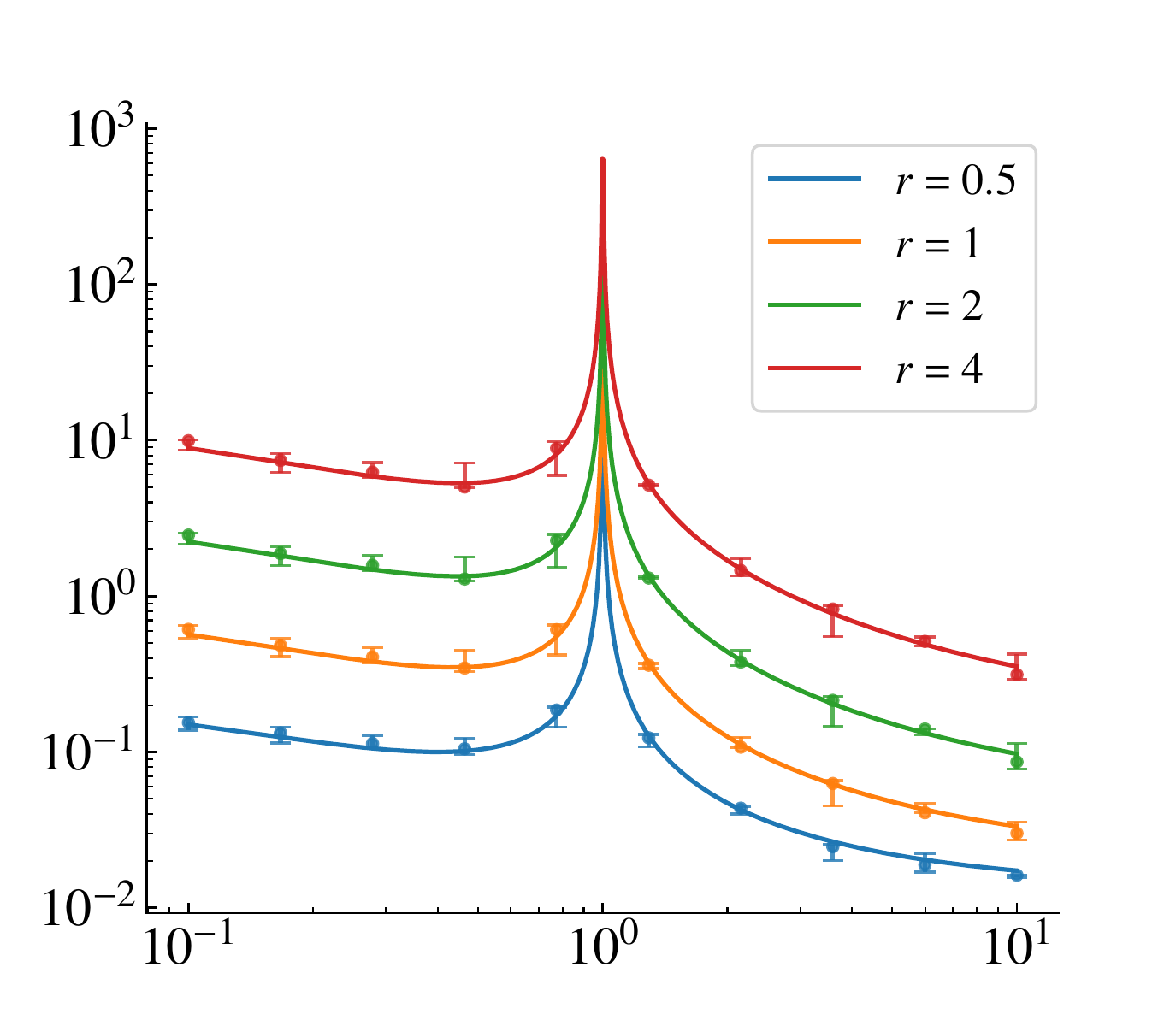}
		\caption{\textit{Asymptotic models of the double descent}. \textit{Left:} Marchenko–Pastur distribution in black for $\gamma = 0.1$. Together with the distribution we also show the histogram of the (nonzero) eigenvalues of the matrix $\frac{1}{N} X^T X$, where $X \in \mathbb{R}^{N x m}$ with $\gamma = m / N= 200 / 2000 = 0.1$. The similarities are apparent. \textit{Middle:} Mean square test error for a model with isotropic features, i.e. $\text{Cov}[x_i] = I$. In full line we illustrate the asymptotics provided in \eqref{eq:asymptotic_l2_isotropic}  and the error bars show the median and the interquartile range for 5 experiments. The experiment is for $\sigma = 1$ and for varying values of signal amplitude $r$. \textit{Right:} Example with spiked covariance. Now the first $d=10$ eigenvalues are have larger amplitude and the remaining one have smaller values. This example is described in Hastie et al.~\cite[Section 6.2]{hastie2022surprises}.}
		\label{fig:dd_asymptotics}
	\end{figure*}

	The first are \emph{high-dimensional asymptotic} results. They consider the number of features $m$ and the number of data points $N$ and make them jointly grow to infinity, i.e. $N, m \rightarrow \infty$ in such way that $\frac{N}{m} \rightarrow \lambda$ where $\lambda$ is a fixed fraction. For example, random matrix theory has asymptotic results for the eigenvalues, eigenvectors and other properties of matrices with random entries~\cite{anderson_introduction_2009,bai_spectral_2010,tao_topics_2012}. 
	%Such theory indeed deals with the cases where the number of features $m$ and the number of data $n$ tend to infinity, 
	The framework and potential of such theory, for explaining and studying neural networks when $m$ and $N$ tend to infinity, have been the focus of recent work, see e.g. \cite{pastur_random_2020,pennington_emergence_2018,pennington_nonlinear_2017} and also Fig.~\ref{fig:dd_asymptotics} where one result is illustrated. 
	Random matrix theory has also been a powerful tool in generating theoretical models for the double-descent phenomenon~\cite{advani_highdimensional_2020,adlam_neural_2020,belkin_two_2020,hastie2022surprises,mei_generalization_2022}. 
	
	Another powerful set of tools are \emph{non-asymptotic bounds}. Indeed, the benign overfitting phenomenon (described later on) is derived in~\cite{bartlett_benign_2020} using such bounds.
	In this case, the number of parameters $m$ and data points $N$ are considered as fixed to finite values. High-probability statements are then derived as a function of them. In the last decade the study of concentration inequalities has gone through significant developments due to its application in high-dimensional statistics and signal processing, see the textbooks~\cite{vershynin_highdimensional_2018,wainwright_highdimensional_2019} for a comprehensive cover this topic. We also mention that these areas are all closely related: tail bounds and concentration inequalities typically underlie the proofs of classical asymptotic theorems.

	\paragraph*{Asymptotic models of double-descent}
	
	As one simple example of asymptotic analysis, we highlight results from~\cite{hastie2022surprises}. Consider a simple linear regression problem where training and test data have been generated linearly with additive noise:
	\begin{align}
	\label{eq:linear-data-model}
	y_i = x_i^T \beta + \epsilon_i, \qquad (x_i, \epsilon_i) \sim P_x\times P_\epsilon.
	\end{align}
	In \eqref{eq:linear-data-model}, $P_\epsilon$ is a distribution in $\R$ such that $\mathcal{E} \epsilon_i = 0$ and  $\mathcal{E} \epsilon^2_i = \sigma^2$. $P_x$ is a distribution in $\R^m$ which implies $\beta\in\R^m$. Moreover, $\mathcal{E} x_i = 0$ and $\text{Cov}[x_i] = \Sigma$. The $\epsilon_i$ and the $x_i$ are all mutually independent. The norm of the data generation parameter is denoted by $\|\beta\|^2_2 = r^2$.
	
	Now, assume that $\beta$ has been estimated using the minimum-norm solution: $\widehat{\beta} = (X^TX)^\dagger X^T y$, where $\dagger$ denotes the pseudoinverse, we obtain that:
	\begin{align}
	\widehat{\beta} = (X^TX)^\dagger X^TX \beta + (X^T X)^\dagger X^T \epsilon.
	\end{align}
	The first term can be understood as the projection of the true parameter $\beta$ that generated the data in the row space of $X$. The second term is related to how much the model learns from the noise.
	
	When $\Sigma = I$, the eigenvalues of $\frac{1}{N} X^T X$ converge to the Marchenko–Pastur distribution, see the leftmost plot in Fig.~\ref{fig:dd_asymptotics}. This fact is used in~\cite{hastie2022surprises} to obtain asymptotics concerning the error in new data points as well as the parameter norm of the estimated parameters. Specifically, as $m, N\rightarrow \infty$ with $m/N \rightarrow \gamma$, almost surely 
	\begin{eqnarray}
	\E{\|y - x^T \widehat{\beta}\|} &\rightarrow&
	\begin{cases}
	\sigma^2 \frac{\gamma}{1 - \gamma} + \sigma^2& \gamma < 1, \\
	r^2 (1 - \frac{1}{\gamma}) + \sigma^2 \frac{1}{\gamma - 1}+ \sigma^2 & \gamma > 1,\end{cases}\label{eq:asymptotic_l2_isotropic}\\
	\E{\|\widehat{\beta}\|_2^2} &\rightarrow&
	\begin{cases}
	r^2 + \sigma^2 \frac{\gamma}{1 - \gamma} & \gamma < 1, \\
	r^2 \frac{1}{\gamma} + \sigma^2 \frac{1}{\gamma - 1} &  \gamma > 1.
	\end{cases}
	\end{eqnarray}
	This example is illustrated in the middle plot of Fig.~\ref{fig:dd_asymptotics}. While simple and displaying a double-descent behavior, this is a theoretical model for which the performance in the underparametrized region is still better than the one in the overparameterized region.

	The right panel of Fig. \ref{fig:dd_asymptotics} shows one example also from~\cite{hastie2022surprises} where the covariance matrix of $x$ has a spiked structure: the first $d$ eigenvalues have high value while the remaining ones are small and correspond mostly to error in variables components. As discussed in the next subsection, the profile of the eigenvalues of the data covariance matrix plays an important role for obtaining good performance in the overparameterized region. Usually, having some more important components and a sufficiently fast decay of the other ones is related to it.
	
	Asymptotic results for more realistic scenarios have also been derived. For instance, Mei and Montanari~\cite{mei_generalization_2022} derive asymptotics for the random features model. It
	is actually described as:
	\begin{align}
	\label{eq:random_fourier_features}
	\hat{y}_i = \beta^T  \sigma(W x_i) 
	\end{align}
	where the matrix $W$ is initialized from a random distribution and only the parameter $\beta$ is trained. This is the model used for estimating the nonlinear ARX model in Fig.~\ref{fig:dd_narx} and can be understood as a neural network where only the last layer is trained.
	
	\paragraph*{Benign overfitting}
	
	Bartlett et al.~\cite{bartlett_benign_2020} coined the term ``benign overfitting''. Overfitting describes phenomena where models fit noisy data too well. Most classical texts on statistics and machine learning associate to these situations poor generalization. The viewpoint has become so prevalent that the word ‘overfitting’ is often taken to mean both fitting data better than should be expected and also giving poor predictive accuracy as a consequence. In~\cite{bartlett_benign_2020}, the term is used in the literal meaning of the word ‘overfitting’. Next, conditions under which this overfitting is not necessarily harmful for model performance are obtained.
	
	Indeed Bartlett et al.~\cite{bartlett_benign_2020} provide a finite sample characterization of the error in linear regression problems and establish conditions for benign overfitting to occur. They show that the eigenvalues of the data covariance play a crucial role in the characterization of benign overfitting and need to decay with certain rates. The covariates must lie in some dominant directions corresponding to large eigenvalues while still have small components (and hence unimportant) along directions in parameter space where the label noise can be hidden. Mallinar et al.~\cite{mallinar2022benign} extend the taxonomy of overfitting from benign and catastrophic to ‘tempered’. The authors argue that many interpolators such as DNNs do neither overfit benignly nor catastrophically.

	% Robustness and adversarial attacks
	% Uniform convergence explanations to the generalization of neural networks? (Maybe in the open problems section, since there is a lot of debate about it...)
	% ---------- End AHR text ----------- %
	
	\subsection{The risk of overfitting in linear system identification}
	\label{SecOverfitLinSysid}
	
	Studies on overparametrization like \cite{bartlett_deep_2021,Liang2020} include conditions such that the ``simplest" (minimum norm) function that interpolates the identification data may generalize well on new outputs. However, results mainly concern predictions of future data which follow the same distributions of the training data. This can be a drawback in control where one often wants robustness e.g. with respect of significant variations in the system input and of disturbances. In this regard, carefully-tuned regularization can largely outperform the implicit regularization obtained by the pseudo-inverse since it can encode useful information about the system. This is now illustrated in a linear setting using the kernel-based regularization described in Section \ref{SectionDeepKernel}.

	%where impulse responses have to be estimated from convolutions with a 
	%\dg{LL: The rss systems are continuous time. theta i contains 1000 impulse coefficients,”time scaled so they are practically 0 after t=1”. What
	%exactly does that mean?}
	Different discrete-time stable impulse responses are obtained by converting in discrete-time random rational transfer functions of order 30 obtained by the MATLAB command $\texttt{rss}$. In particular, the continuous-time function returned by this routine is truncated over the finite interval $[0,T]$, with $T$ such as to capture $99\%$ of its two-norm. The impulse response is then uniformly sampled to obtain a vector $\theta=[\theta_1 \ \theta_2 \ \ldots \theta_{1000}]$ containing impulse response coefficients. Our measurements model is
	\begin{equation}\label{MeasOverfitting}
	y_i = \sum_{i=1}^{t_i}  \theta_i + e_i \quad i=1,\ldots,200 %\ \int_0^{t_i} f(t)dt+ e_i=:L_i[f]+e_i,  \quad i=1,\ldots,200
	\end{equation}
	where $t_i$ are realizations of independent uniform random variables over $\{1,2,\ldots,1000\}$. The $e_i$ form a white Gaussian noise, independent of $t_i$, and the signal-to-noise ratio is 100. Estimation of $\theta$ from $Y=[y_1, y_2, \ldots, y_{200}]^{\top}$ is ill-posed and ill-conditioned since the measurements are obtained by discrete-time integrators. This is a situation frequently encountered in real applications where a dynamic system can be subject to a low-pass input.
	
	The impulse response estimate $\hat \theta$ is computed using regularized least squares (ReLS) \eqref{RegFIR} with regularization matrix $P$ defined by the stable spline/TC kernel \eqref{SSKernel} which provides information on smooth exponential decay. The strength of the regularizer  is thus established by the regularization parameter $\gamma$ and the decay rate $\alpha$.
	
	Consider two versions of this estimator which differ in the adopted hyperparameters values. The first one, denoted by \text{ReLS-TC}, determines  $\gamma$ and $\alpha$ via marginal likelihood (ML) optimization \eqref{MLestimate}.  The second estimator uses the ML estimate of $\alpha$ but then computes the impulse response of minimum stable spline norm %(measured by $\theta^{\top} P^{-1} \theta$) 
	which interpolates the outputs, i.e. it computes
	\begin{align}
	\argmin_{\theta} \ \theta^{\top} P^{-1} \theta \ \ \text{s.t.} \ \ y_i = \sum_{i=1}^{t_i}  \theta_i, \ i=1,\ldots,200. 
	\end{align}
	%$$
	%\lim_{\gamma \rightarrow 0} \left(\Phi^{\top} \Phi + \gamma P^{-1}\right)^{-1}\Phi^{\top}Y.
	%$$
	The resulting estimator is denoted by \text{PI-TC} where PI stays for Pseudo Inverse.
	
	Given validation outputs $y_{\text{v}}$, the prediction fit returned by a predictor $\hat y$ is
	\begin{equation}
	\mathcal{F} = 100 \left(1-  \frac{\|y_{\text{v}} -\hat y \| }{\| y_{\text{v}} - \bar y_{\text{v}} \| } \right),
	\label{eq:fit nrmse}
	\end{equation}
	where $\bar y^{\text{v}}$ is the average value of the validation outputs. In our case, we have two predictors at stake, associated to \text{ReLS-TC} and \text{PI-TC}. In addition, we will consider two different validation sets of size 1000. The first one contains outputs generated as in \eqref{MeasOverfitting} but without adding noise and generating other $1000$ independent realizations of $t_i$. So, the first validation set is still connected with a low-pass (LP) input. The second one contains 1000 noiseless outputs from the FIR with coefficients in $\theta$ with white Gaussian noise (WN) as input.\\
	
\begin{figure}
		\begin{center}
			\begin{tabular}{c}
				{\includegraphics[scale=0.43]{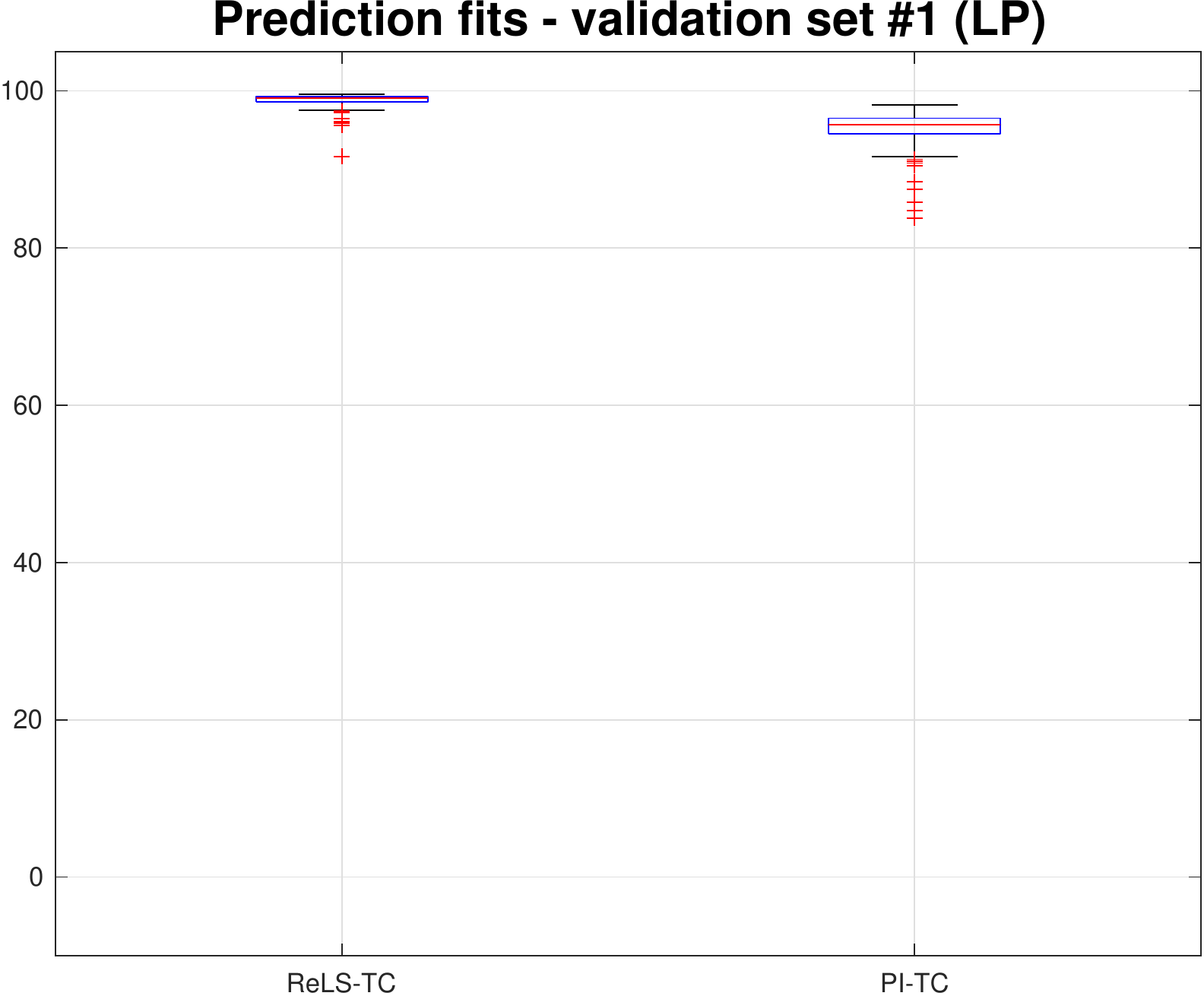}} \\
				{\includegraphics[scale=0.43]{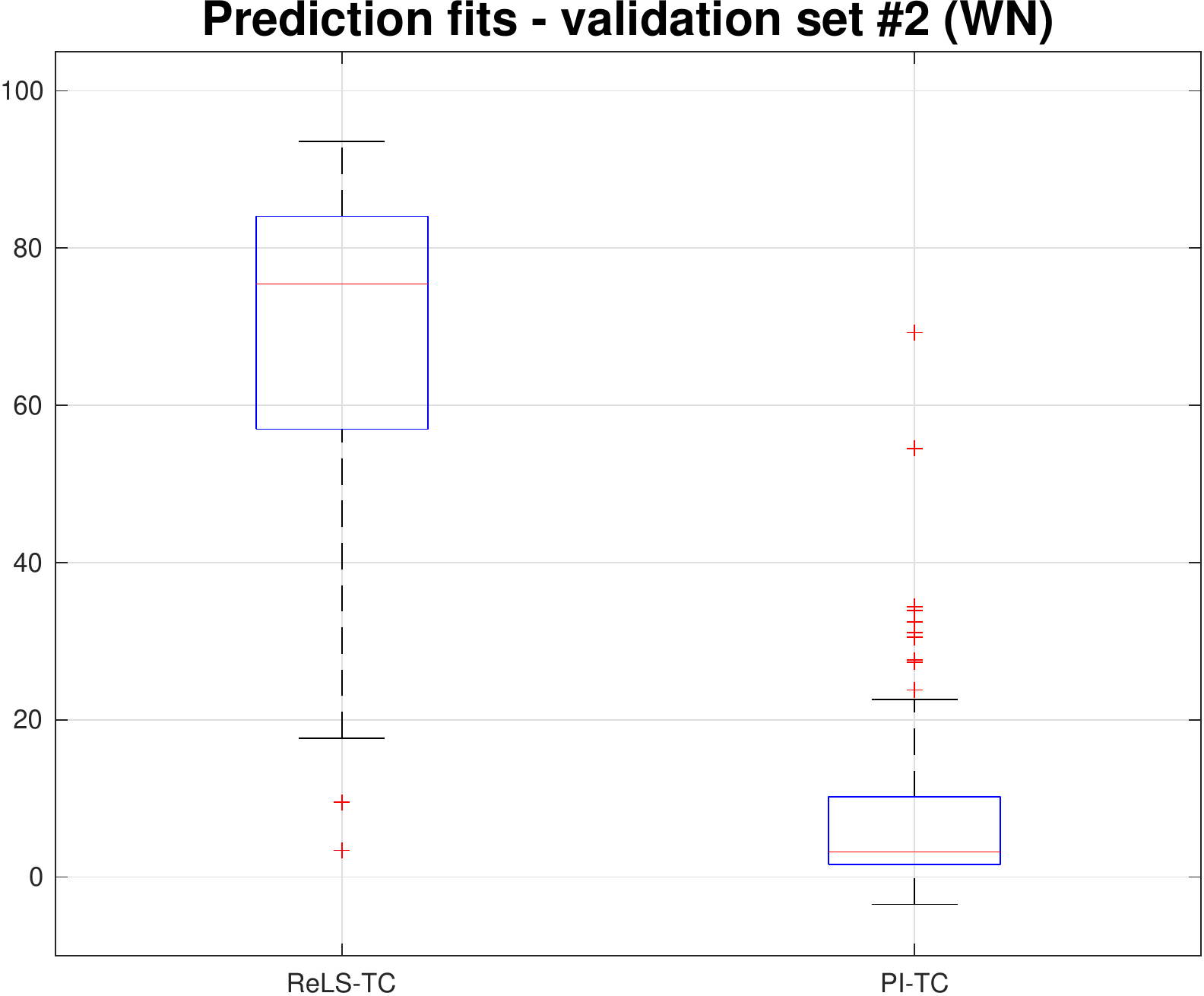}}  
			\end{tabular}
			\caption{Boxplots containing prediction fits returned by \text{ReLS-TC} and \text{PI-TC} when validation data follow the same statistics of the training data defined by low-pass inputs (top) and when they are generated using white Gaussian noise as input (bottom). In this latter case, carefully-tuned regularization (where e.g. information on smooth exponential decay of impulse responses is encoded) may thus lead to predictions more robust w.r.t. input variations in comparison with minimum-norm solutions (which we have seen to play a special role in the study of overparameterized models).}
			%\dg{LL: Estimating models with regularized kernels and computing  performance for data with different characteristics is something I am quite familiar with. But what does fig 12 tell us about the double descent issue?} 
			\label{FigDDP}
		\end{center}
	\end{figure}

	Results from a Monte Carlo experiment of 100 runs are in Fig. \ref{FigDDP} where Matlab boxplots of the 100 prediction fits returned by the two estimators are displayed. Using the first validation set (top), \text{ReLS-TC} provides the best results but the prediction performance of \text{PI-TC} is comparable. This means that the stable spline pseudo-inverse may well generalize on new data which are still integral versions of the impulse responses. The picture changes completely by considering validation data generated using $WN$ as input. Using the second test set (bottom) one can see that carefully tuned regularization in many cases generalize well also when future data follow statistics much different from those underlying the training set.

	\subsection{Approximation properties of deep neural networks}
	\label{sec:approxcapability}
	
	Some recent studies on deep network properties in approximating input-output maps are now briefly discussed. They support the use of these networks to face the curse of dimensionality in regression problems.
	
	Consider the fully connected neural network with one hidden layer given by \eqref{eq:fully_connected} with null $b_2$. It is composed of $n$ units (basis functions) that receive the vector $x \in \mathbb{R}^m$ as input and return as output the scalar
	\begin{equation}\label{ShallowNet}
	\hat{y}(x) = \sum_{k=1}^{n} \alpha_k \sigma(w_k^\top x + \beta_k).
	\end{equation}
	%Weights vectors $w_k$ and scalars $\alpha_k,\beta_k$ 
	%have typically to be tuned from data. 
	Consider now the following approximation problem. A smooth map $g$ over a compact set of $\mathbb{R}^m$ and a resolution $\epsilon>0$ are given.  %continuous partial derivatives of order $m$ 
	Then, the network has to adjust weight vectors $w_k \in \mathbb{R}^m$ and scalars $\alpha_k,\beta_k$ to synthesize $f$ such that
	\begin{align}
	\sup_x |f(x)-g(x)| \leq \epsilon.
	\end{align}
	It turns out e.g. from \cite{Barron93,Mhaskar96,PoggioPNAS2020,IJAC2017} that this goal can always be achieved by the shallow network \eqref{ShallowNet} which is therefore universal. However, the number of required trainable parameters is $O(\epsilon^{-m})$, hence the network's complexity scales exponentially with the required accuracy\footnote{More specifically, this and the next result here discussed hold by choosing $\sigma$ infinitely differentiable and different from a polynomial. Furthermore, the sup-norm of $g$ and of its first-order partial derivatives are assumed bounded over its compact domain.}. This is the best possible result \cite{Micchelli1989} unless additional information on $g$ is provided.
	
	As the intuition suggests, a situation where a deep network can significantly improve this outcome is when $g$ has a compositional structure. As an example, $g$ could be defined by local functions of only two variables, e.g. when $x$ has only 8 components $x_i$, we have
	\begin{align}
	\begin{split}
	g(x) =h_3(&h_{21}(h_{11}(x_1,x_2),h_{12}(x_3,x_4)), \\
	&h_{22}(h_{13}(x_5,x_6),h_{14}(x_7,x_8))),
	\label{BinaryTree} 
	\end{split}
	\end{align}
	
	where each function $h_3,h_{21},\ldots$ is implemented via a node of the network through a certain number of units. It turns out that a deep network, given by a simple binary tree with three layers if \eqref{BinaryTree} holds, achieves the same resolution $\epsilon$ as the shallow network but with a different number of parameters, $O((m-1)\epsilon^{-2})$. Such results extend to other hierarchical functions and the complexity depends on the (possibly different) number of edges entering the nodes \cite[Section 4]{IJAC2017}. Hence, shallow and deep networks are both universal approximators (they can approximate any reasonable continuous function arbitrarily well). However, by knowing the compositional structure the challenging curse of dimensionality can sometimes be handled by reducing the number of model parameters from exponential to linear in the dimension~$m$ of the function domain.
	
	Other studies that support deep networks for regression are found in \cite{Daubechies2021,Telgarsky2015}. Consider the function
	$$%$$\begin{equation}\label{DeepStart}
	f = f^L \circ \ldots \circ f^2 \circ f^1,
	$$%\end{equation}
	induced by a deep network which adopts 
	the ReLU $\sigma(z)=z_{+}=\max(z,0)$. Each unit thus returns the piecewise
	linear function %hence re
	%$\alpha_k$
	$\sigma(w_k^\top x + \beta_k)$ 
	with breakpoint established by $\beta_k$. 
	This makes also the overall function $f$ 
	%makes any term in \eqref{ShallowNet} 
	piece-wise linear being formed by linear operations mixed with the max
	function. So, the flexibility of this network
	can be measured by how many breakpoints can be introduced in the
	curve $f$.
	Such concept has some limitations since there can be hidden correlations between the breakpoints induced by the model. However, focusing mostly on the case of scalar inputs ($m=1$) and outputs, %interesting studies 
	results in \cite{Daubechies2021,Telgarsky2015} and simulations based on random networks in \cite[Section 6]{LjungDeep2020} suggest that, while the number of breakpoints grows quickly with the depth of the network, the number of parameters does not increase with the same rate. This means that increasing the number of layers is more effective than increasing the number of units per layer. %Focusing on the univariate case, 
	%\cite{Daubechies2021} also shows that,
	For very special selections of weights and biases, the maximum number of breakpoints scales as $d^L$ if all the $L$ hidden layers have the same  
	width (number of units) $d$. It is also proved that deep ReLU networks reach the same approximation power of free knot linear splines with a comparable number of parameters, but can also generate other functions distant from them. Advantages of deep learning for other classes, like power and Lipschitz functions, are discussed in \cite{YarotskyA2017,YarotskyB2017}.

	\section{Applications}\label{SectionDeepApp}
	
	\subsection{Software issues}
	Deep neural networks have been used extensively and a substantial amount of software  has been developed in different contexts and for different platforms. Python implementations, like \textsc{PyTorch}\footnote{https://pytorch.org/} \cite{paszke2017automatic}, \textsc{TensorFlow}\footnote{https://www.tensorflow.org/}  \cite{abadi2016tensorflow} and  \textsc{JaX}\footnote{https://github.com/google/jax} \cite{jax2018github} are quite popular and are open source libraries. In this survey we focus on \textsc{Matlab} and \textsc{PyTorch}. 
	
	MATLAB is a proprietary programming and numeric computing platform. \textsc{Matlab} has a \emph{Deep Learning Toolbox} (DLT) which implements most of the deep networks for various applications, described in this paper. Of particular interest for system identification is how the DLT cooperates with the \emph{System Identification Toolbox} (SITB) \cite{manualeLjung} to identify systems using deep learning. SITB has a model object \texttt{idnlarx} that implements nonlinear models of ARX-character
	\begin{equation}
	\label{eq:1}
	\hat{y}(t|\theta) = g(\theta,\varphi(t)),
	\end{equation}
	%(cf Section{\ref{}}) 
	where $\varphi(t)$ is a vector of a finite number of past inputs and outputs while $g$ is a nonlinear function that can be defined in many different ways. If \texttt{NET} is a shallow or deep neural network defined in DLT, a \texttt{nlarx} model can be defined by
	\begin{verbatim}
	mn = idFeedforwardNetwork(NET)
	\end{verbatim}
	The object \texttt{mn} is now a regular SITB model that can be estimated and evaluated by any SITB command, e.g.
	\begin{verbatim}
	mhat = nlarx(Data,[2 2 1],mn)
	compare(Data,mhat);
	ys = sim(mhat,newdata);
	resid(Data,mhat)
	\end{verbatim}
	
	\textsc{PyTorch} is an open-source framework implemented in Python.  The framework provides the basic components for the implementation and training of neural networks and it is widely used in the machine learning community and is the basis of many state-of-the-art applications. There is no specific system identification package, but the building blocks provided allow for the facilitated implementation of all the models described in this paper. For instance, a fully connected network with one hidden layer would be implemented as:
	\begin{verbatim}
	class FCNetwork(nn.Module):
	def __init__(self, n_inputs, n_hidden):
	super(FCNetwork, self).__init__()
	self.net = nn.Sequential(
	[
	nn.Linear(n_inputs, n_hidden),
	nn.ReLU(), 
	nn.Linear(n_hidden, 1)
	])
	
	def forward(self, x):
	return self.net(x)
	\end{verbatim}
	
	The package provides:
	\begin{enumerate}
		\item  GPU acceleration capabilities. The acceleration is obtained with few lines of code and can make the neural network implementation and training significantly faster, e.g.,
		\begin{verbatim}
		model = FCNetwork()
		device = torch.device("cuda")
		model.to(device=device)
		\end{verbatim}
		\item Automatic differentiation capabilities, that allow the derivatives of the neural network to be easily computed and gradient-descent-based optimization to be used to update the model. For instance, one step of an stochastic gradient descent could be computed as:
		\begin{verbatim}
		model_output = model(x)
		loss = loss_function(model_output, y)
		loss.backward()
		optimizer.step()
		\end{verbatim}
		
	\end{enumerate}
	
	An example implemented in \textsc{MatLab} is given in Section~\ref{DuffingExample} and an example with \textsc{PyTorch} is given in Section~\ref{ce-example}.

	\subsection{An example with real data: forced duffing oscillator}
	\label{DuffingExample}
	%\emph{\textcolor{red}{must be modified due to copyright issues}}
	The system is an electronic circuit that mimics a nonlinear mechanical system with a cubic hardening spring. This class of nonlinear systems has a very rich behavior, including regular and chaotic motions, and the generation of sub-harmonics. The dataset\footnote{Data are available at \url{www.nonlinearbenchmark.org/\#Silverbox.} The selection of estimation and validation data is the same as in Listing S1 in \cite{SchoukensLjung:2019}.}has often been used in nonlinear identification benchmarks, also to test deep networks, see e.g. \cite{LjungDeep2020,SchoukensLjung:2019}. 
	%A dataset that has been often used  as a nonlinear benchmark case is the so-called ``Silverbox Example''. This is a  Forced Duffing Oscillator which is an electronic circuit that mimics a mechanical system with a cubic hardening spring. The experiment is described in \cite{SchoukensVP:16} and the data is available on www.nonlinearbenchmark.org/\#Silverbox. 
	
\begin{figure}
		\centering \includegraphics[scale=0.4]%{Figures/edata_enc.pdf} \\ \hspace{-0.0cm}\includegraphics[scale=0.40]
		{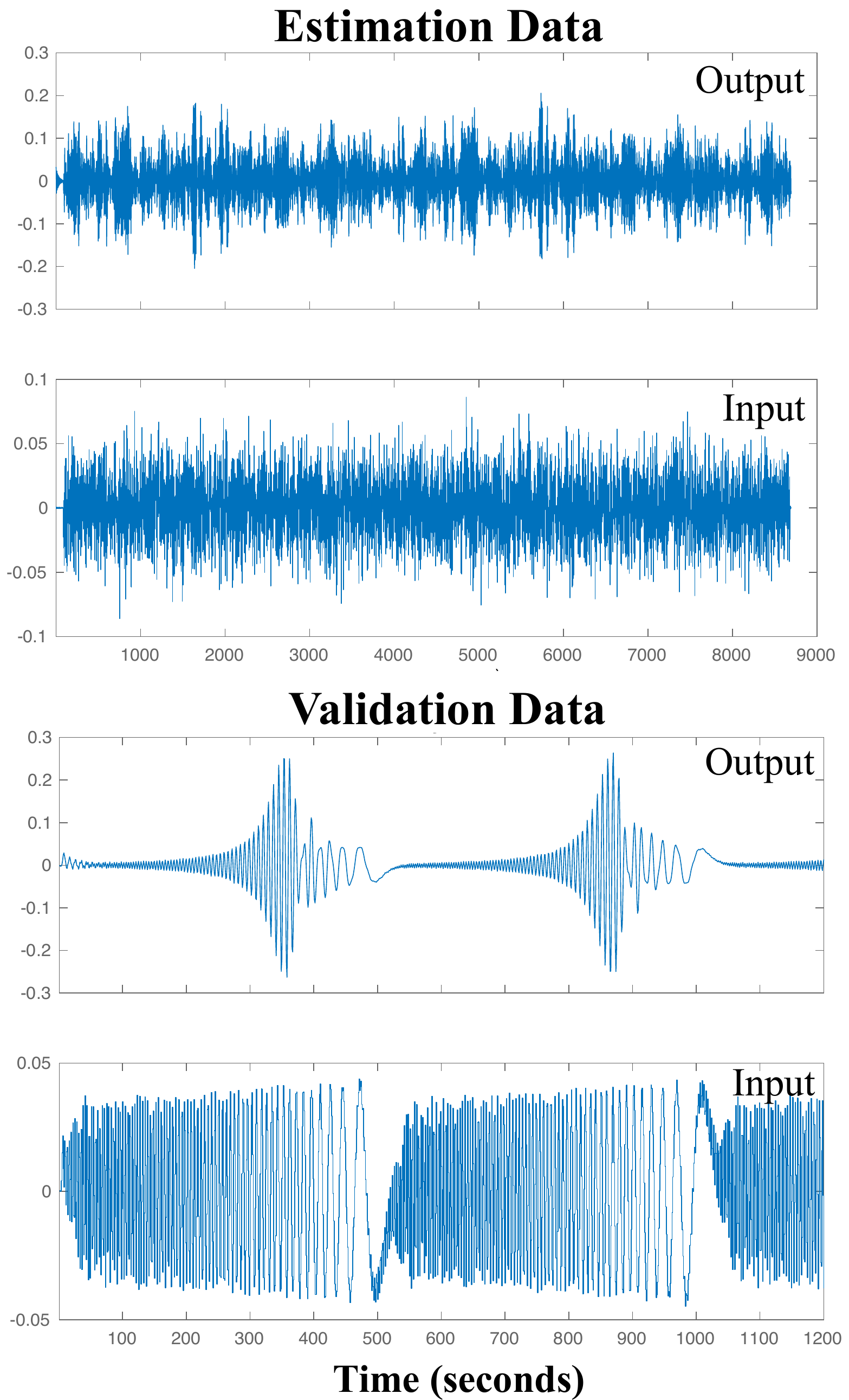}%{Figures/vdata_enc.pdf}}
		\caption{The Silverbox data. From top: estimation and validation data.}
		%\caption{The Silverbox data. Left: estimation  output data, estimation input data, validation output data, validation input data.}
		\label{fig:SB}
	\end{figure}

	Two sets of data were collected with different excitations, depicted in Fig.~\ref{fig:SB}. 
	The estimation data are used to fit a model. Then, following what was illustrated in Section~\ref{sec:modeling-dynamical-sys}, the model's validity is checked by how well it reproduces the validation data according to \eqref{ValidationFit}. That is, the model is simulated using the validation data input and the discrepancy between model output and measured validation data output is determined. %This is cross-validation, a standard system identification task, as described in eq (\ref{ValidationFit}).
	
\begin{figure}
		\centering
		\includegraphics[scale=0.4]{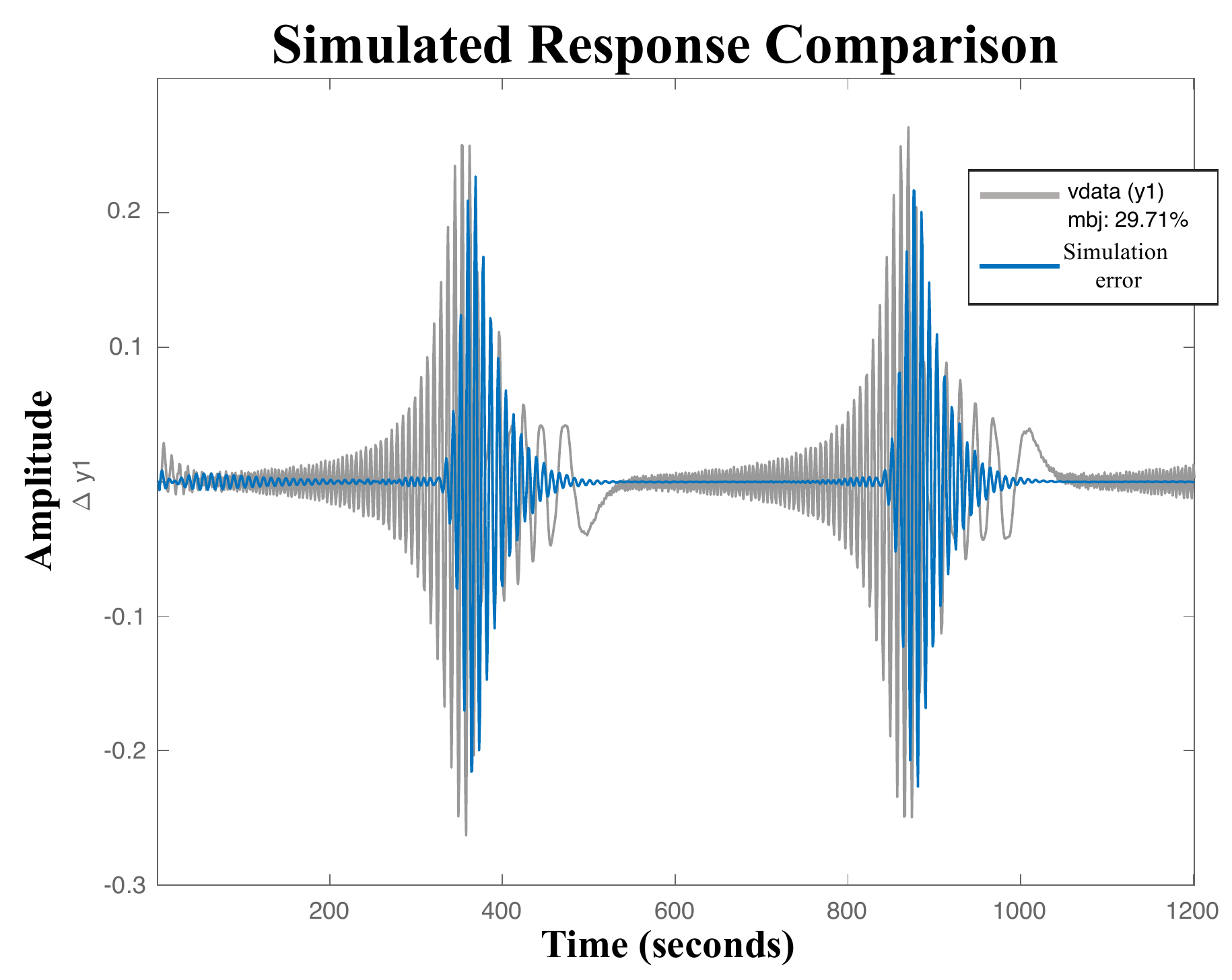}
		\caption{The validation output (grey) and the error between the validation output and the model
			simulated output (blue) for the Box-Jenkins model. The fit is
			\textbf{29.7 \%}}
		\label{fig:bj_comp}
	\end{figure}

\begin{figure}
		\centering
		\includegraphics[scale=0.4]{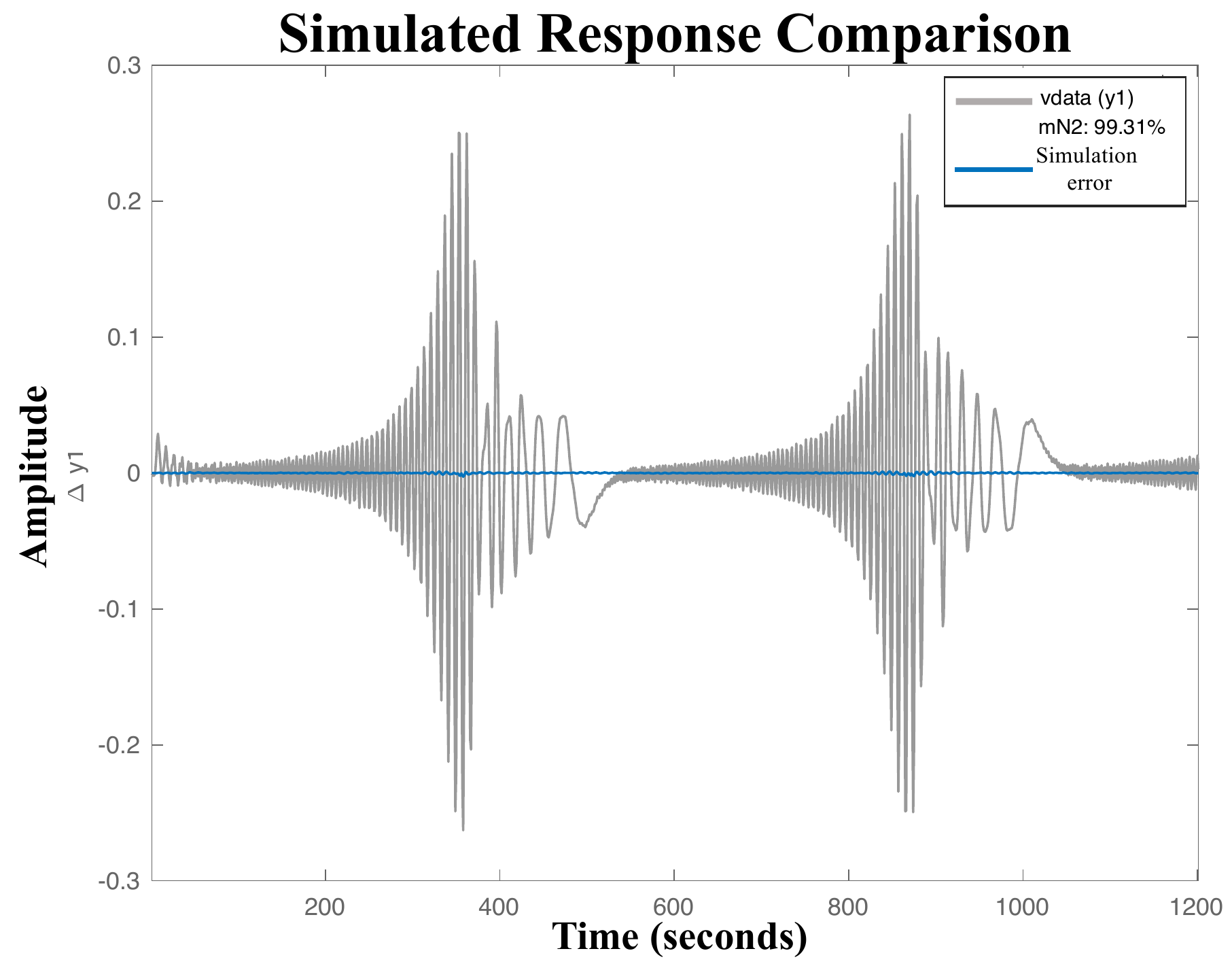}
		\caption{The validation output (grey) and the error between validation output and the model
			simulated output (blue) for the deep cascadeforwardnet
			model. The fit is \textbf{99.2 \%}}
		\label{fig:nn6_comp}
	\end{figure}

	%To load the estimation and validation data the following MATLAB sequence can be used
	%\begin{verbatim}
	%load SNLS80mV.mat
	%fs=1e7/2^14;
	%edat=iddata((V2(30550:38500)',...
	%       V1(30550:38500)',1/fs);
	%load Schroeder80mV
	%ySchroeder=V2(10585:10585+1023);
	%uSchroeder=V1(10585:10585+1023);
	%vdat=iddata(ySchroeder',uSchroeder',1/fs);
	%end{verbatim}
	\paragraph*{Linear Model}
	A common linear model of the Box-Jenkins type (see e.g.~Chapter 4 in \cite{Ljung:99})  with 4 poles, 4 zeros and a second order noise transfer function, is estimated by
	\begin{verbatim}
	mbj=bj(edat,[4 4 2 2 0]);
	\end{verbatim}
	The accuracy of this model is computed by
	\begin{verbatim}
	[ys,Fit]=compare(vdat,mbj)
	Fit=29.71 % (percent of the output variation ...
	reproduced by the model)
	\end{verbatim}
	%Note that "Fit" is defined by (\ref{Fith}).
	%\dg{Can we mathematically explain what the 'fit' means?}
	with the Fit defined as in \eqref{eq:fit nrmse}.
	%\end{document}
	The validation output and the error between this and the simulated model output is
	shown in Fig.~\ref{fig:bj_comp}. It can be seen that the simulation is not very good. The error is at times as big as the signal itself.
	%\end{document}
	Residual analysis as mentioned in Section \ref{par_mv} can also be performed by the command 
	\texttt{resid(vdata,mbj)}. 
	It shows that there is quite significant correlation between the residuals and the input. The model should thus be rejected.%left in the residuals.
	%\begin{figure}
	% \centering
	%  \includegraphics[scale=0.4]{Figures/bj_resid}
	%  \caption{Residual analysis for the validation data and the
	%    Box-Jenkins model. Left: the autocorrelations for the
	%    residuals. Right: Cross correlation between input and residuals}
	%  \label{fig:bj_resid}
	%\end{figure}
	\paragraph*{Deep Cascaded Network}
	A deep learning model %5(\ref{eq:dnL}) 
	with 6 layers of feedforward nets (cf Sec. \ref{sec:cascade}) with 6 nodes each is estimated and validated by
	\begin{verbatim}
	net=cascadeforwardnet([6,6,6,6,6,6]);
	N2=idFeedforwardNetwork(net);
	mN2=nlarx(edat,[4 4 0],N2);
	[Ys,Fit]=compare(vdat,mN2)
	Fit=99.18
	\end{verbatim}
	The validation output and the error between this and the simulated model output is
	shown in Fig.~\ref{fig:nn6_comp}. The simulation is remarkably good: 
	the error is hardly visible. Recall that the simulation is done 
	without any access to the measured validation output, and note that the
	character of the validation data is quite different from that of the estimation data.
	
\begin{figure}
		\centering
		\includegraphics[scale=0.4]{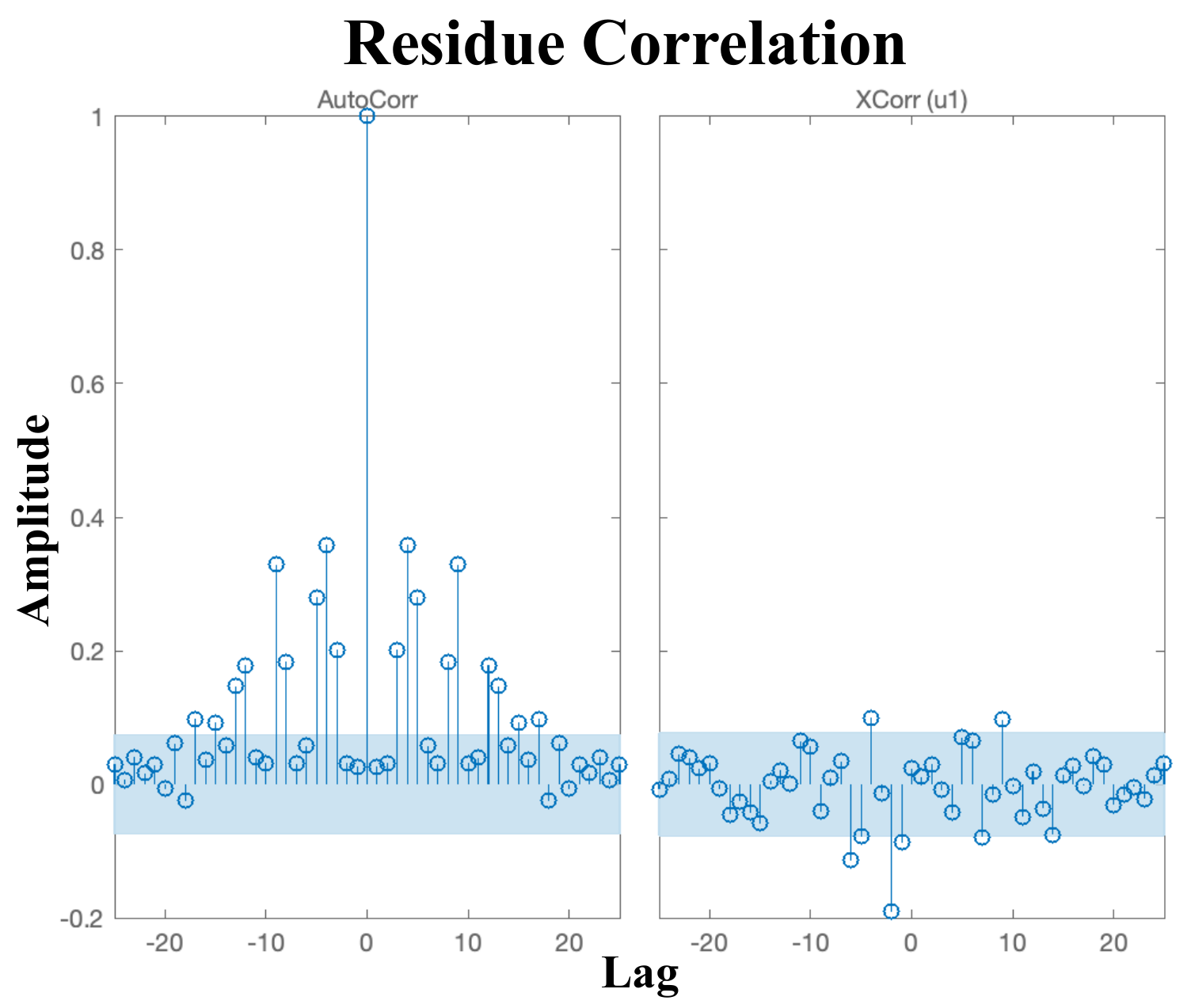}
		\caption{Residual analysis for the validation data and the
			deep net model. Left: the autocorrelation for the
			residuals. Right: Cross correlation between input and residuals. Shaded area: confidence interval}
		\label{fig:nn6_resid}
	\end{figure}

	The residual analysis (\texttt{resid(vdata,mN2)}) for this model is
	shown in Fig.~\ref{fig:nn6_resid}. The correlation between the
	residuals and past inputs (lag larger than 0)  in the right panel is quite insignificant. There is
	some correlation left among the (very small) residuals themselves (left panel) but no attempt
	has been made to build a model for the color of the additive
	disturbances. %Anyway the size of the residuals is very small \dg{informal sentence}.
	
	\subsection{\textsc{PyTorch} example: coupled electronic drives}
	\label{ce-example}
	In this example we will make use of \textsc{PyTorch} code to implement deep neural networks to model an open nonlinear system identification benchmark. The underlying physical system consists of two electrical motors which are connected to a pulley through a flexible belt \cite{wigren_coupled_}. Two different input signals are available\footnote{The complete dataset and dataset descriptions are available at \url{https://www.nonlinearbenchmark.org/benchmarks/coupled-electric-drives}.}: a pseudo-random binary sequence and a uniformly distributed input signal. We use the uniformly distributed input signal to learn the dynamics using 60\% of the data and validate it on the remaining 40\%. This nonlinear dataset was used in previous studies \cite{HendriksGRWS:2021,khandelwal_datadriven} and can be modeled by a Wiener-Hammerstein model~\cite{wigren_coupled_}.
	
	We compare four models; 1) a fully connected (FC) neural network, see Section~\ref{sec:fully_connected}, with three layers and 300 hidden units; 2) an LSTM, see Section~\ref{sec:recurrent-nn} with one layer and 200 hidden units; 3) a deep state-space model (SSM), see Section~\ref{sec:latent-variable}, specifically a VAE-RNN from \cite{fraccaro2018deep,Gedon2021} with one layer for all representations, a hidden size of 100 for the deterministic latent variable and 50 for the stochastic one; and 4) a baseline linear ARX model for comparison. 
	We train all the deep models using stochastic gradient descent with momentum for 100 (FC) or 150 (LSTM and Deep SSM) epochs. The baseline ARX model was trained using the least squares solution. For all models we use an input and output memory of 5, see \eqref{eq:nnarx}, which was the one that gave the best results by grid search for the feedforward neural network model. As a comparison, the reference model provided in \cite{wigren_coupled_} is a continuous system of third order.
	
\begin{figure}
		\centering
		\includegraphics[width=0.8\columnwidth]{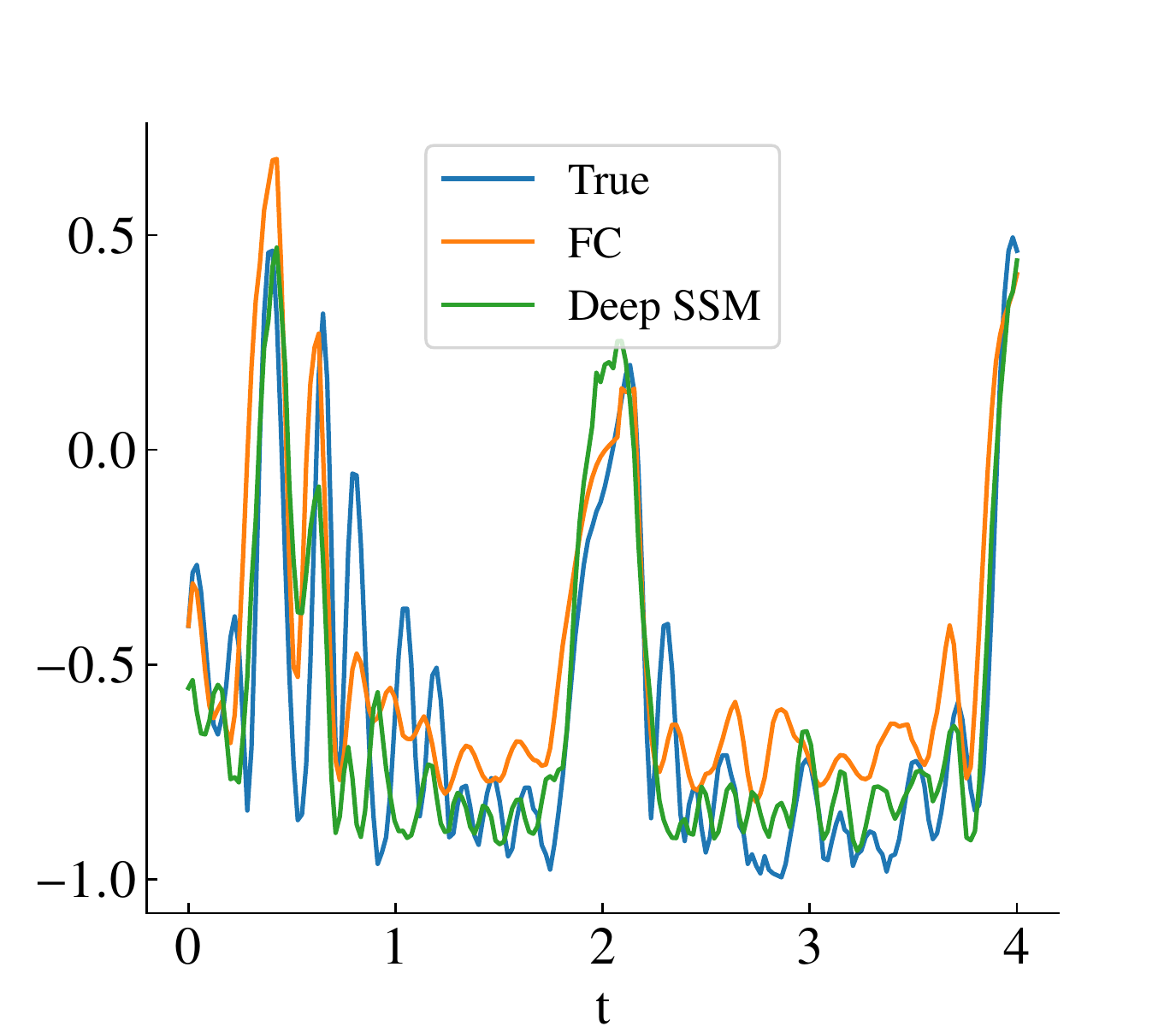}
		\caption{The validation data (blue) together with the simulated response from the fully-connected (FC) network and the deep state-space model (SSM).}
		\label{fig:ce-results-simulated}
	\end{figure}

\begin{figure}
		\centering
		\includegraphics[width=0.7\columnwidth]{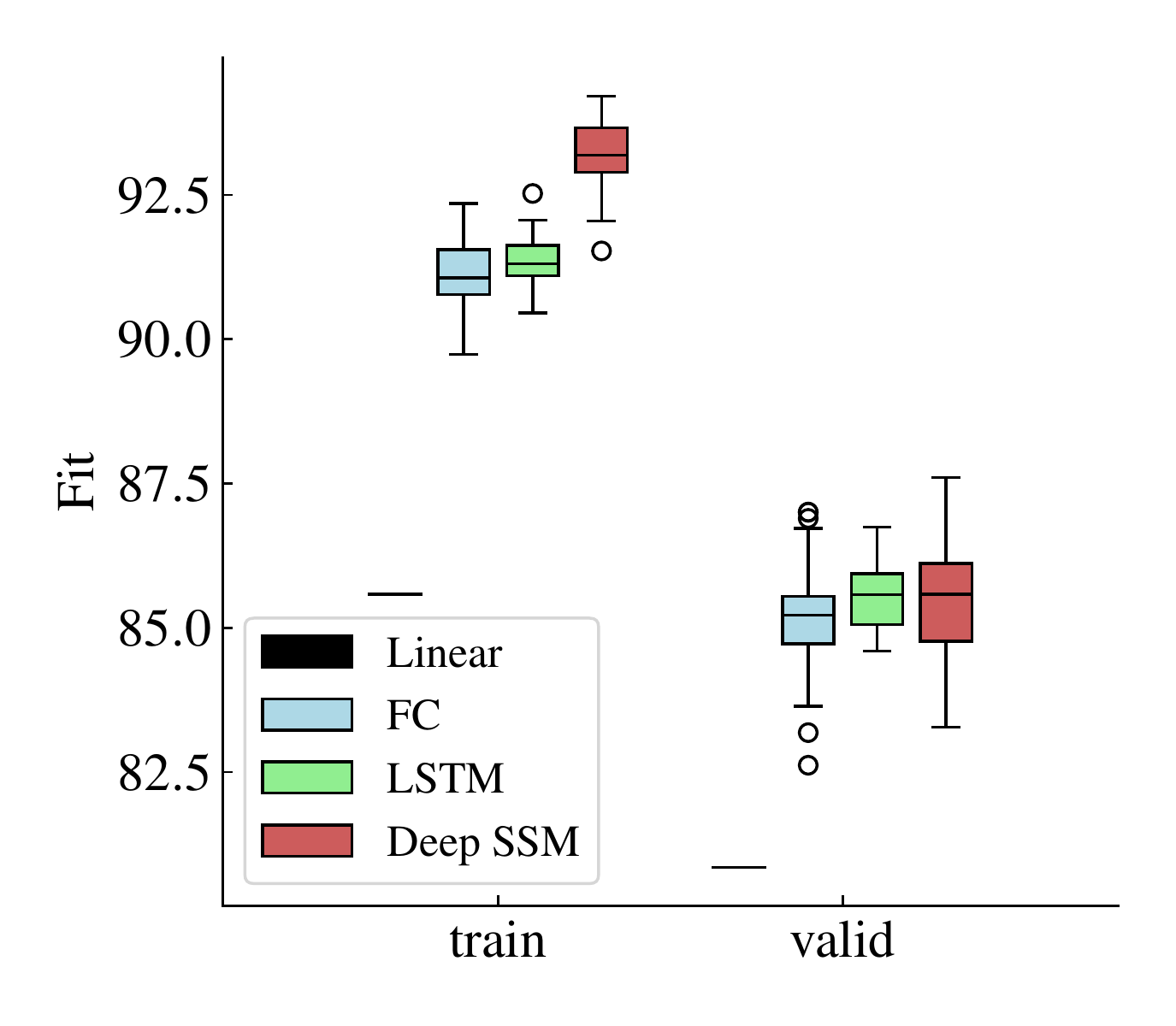}
		\caption{The boxplots show the fit of the simulated response to the training data (left) and validation data (right) for 25 different seeds initialization of the neural network.}
		\label{fig:ce-boxplot}
	\end{figure}

	We evaluate the models using the 40\% hold-out validation dataset. In Fig.~\ref{fig:ce-results-simulated}, we compared the measured validation output with the simulated response from some of the models. We compute the fit as defined in \eqref{eq:fit nrmse} to the validation data as well as to the training data.
	%We compute the Normalized Root Mean Square Error (NRMSE), see \eqref{eq:fit nrmse}.
	%$$NRMSE= 1 - \frac{\|y -\hat{y}\|_2}{\|y -\bar{y}\|_2},$$
	%here $y\in \R^N$ is the measured value, $\hat{y}\in \R^N$ is the vector of simulated values, and $\bar{y}\in \R$ is the mean of the measured values. 
	The same metric is used internally in the function \texttt{compare} in \textsc{MatLab} to provide the \texttt{fit}. In Fig.~\ref{fig:ce-boxplot} we show the fit of the simulated response, the boxplot is displayed for 25 different initializations of each architecture.\footnote{For the linear model there is no stochasticity because of the global least square solution.}
	
	% some discussion
	The results indicate that the deep models show a good performance in comparison to the linear baseline model. This is especially noteworthy as we use 300 time samples for training while our models have $184210$ (FC), $169801$ (LSTM) and $111902$ (deep SSM) parameters. Hence, despite the fact that the models are overparameterized, they generalize quite well on validation data, which is in accordance with the theoretical discussions in Sections~\ref{sec:implicit-regularization} and \ref{sec:double-descent}.

	\section{Conclusions}
	Inferring models of dynamical systems from observed input-output data is the aim of \emph{System Identification}. We have demonstrated in this survey that sophisticated modern  \emph{deep learning} techniques also share the essential features of this problem. Deep (neural) networks, as e.g. described in Section \ref{sec:DNN architectures}, can in fact serve as model structures producing predictions of future observations that can be fitted to observed data, just as in the conventional prediction error approach.
	Analogously, concatenated kernels can define new regularized methods, 
	as the estimator \eqref{KernelRidgeVV} described in Section \ref{SectionDeepKernel}.
	
	What is essential in these deep estimation techniques is (a) that they employ \emph{(very) many parameters}
	and (b) that the parameterizations can
	be made in various compositions,
	exploiting repeated structures (layers) and other concatenations, 
	as in the simple illustrative example \eqref{BinaryTree}. 
	This motivates the concept of \emph{deep} structures.
	
	The large size of the parameter vector calls for special attention, methods,  and ``tricks" in the numerical and algorithmic treatments, as detailed in Section \ref{SectionDeepOpt}. The overparameterized case (more parameters than  observations) also gives rise to model behavior that apparently defies statistical truths: "generalization behavior too good to be true", as illustrated in Section \ref{sec:double-descent}.
	
	In this way, deep learning is not just ``a special case of system identification" and the survey has shown a rich set of issues and techniques that could be useful for expanding the understanding of this fundamental control area. This can be pursued by focusing on several  interesting {\bf{open problems}}. \emph{}
	For instance, while Section~\ref{sec:DNN architectures} presented common deep architectures and how they have been adapted for system identification, there is also a range of modern architectures which have been successful in other tasks but still not extensively used for nonlinear system identification.
	They include: 
	\begin{itemize}
		\item \emph{Attention models} \cite{bahdanau2014neural,luong2015effective} and \emph{transformers} \cite{vaswani2017attention} which are designed for time series and weigh inputs in a non-uniform way to obtain predictions.
		\item \emph{Flow based models} which are generative models that explicitly describe a probability density via normalizing flows \cite{rezende2015variational}. Extensions for dealing with time series are described in \cite{rasul2020multivariate}.
		\item \emph{Generative Adversarial Networks} (GANs) \cite{goodfellow_generative_2014} which consist of two networks trained in a game. One networks generates realistic input data from noise while the other tries to distinguish artificial from real data. Extensions for time series can be found in \cite{yoon2019time}.
		\item \emph{Graph Neural Networks} \cite{bronstein2021geometric,zhou2020graph} which are a natural choice for modelling anything that can be represented by a graph. They can be used as physics simulators \cite{sanchez2020learning} but also for time series forecasting \cite{cao2020spectral}.
		%\item ...
	\end{itemize}
	Regarding overparametrization, Section~\ref{sec:double-descent} described both experimental and theoretical results for \emph{double-descent}. But studies have so far focused on static systems: the challenge is to derive  
	(asymptotic or finite-sample) theoretical models for
	data generated by dynamic equations. Finally, robustness issues, as the building of \emph{uncertainty bounds} around 
	deep models of dynamic systems, can also play an important role in many future control applications.
	\bibliographystyle{plain}
	\bibliography{Biblio}

\end{document}